%% file: manuscript.tex
\documentclass[11pt,a4paper]{article}

\usepackage{iftex}
\ifPDFTeX
  \usepackage[T1]{fontenc}
  \usepackage[utf8]{inputenc}
  \usepackage{lmodern}
  \usepackage[scale=0.92]{tgheros}
\else
  \usepackage{fontspec}
\fi
\usepackage{microtype}

\usepackage[a4paper,
            left=2.8cm,
            right=2.8cm,
            top=2.8cm,
            bottom=2.8cm,
            headheight=36pt]{geometry}

\usepackage{xcolor}
\definecolor{swNavy}{HTML}{0E2A3F}   
\definecolor{swBlue}{HTML}{1B4F72}   
\definecolor{swSteel}{HTML}{3D6E91}  
\definecolor{swSky}{HTML}{8AAFC9}    
\definecolor{swMist}{HTML}{E1EAF1}   
\definecolor{swCloud}{HTML}{F0EEE9}  
\definecolor{swPaper}{HTML}{F8F7F4}  
\definecolor{swStone}{HTML}{5F5B55}  
\definecolor{swLine}{HTML}{CFCBC3}   
\definecolor{swSage}{HTML}{3C7A68}   
\definecolor{swBrick}{HTML}{9B3A3A}  

\colorlet{arpaSteel}{swBlue}
\colorlet{arpaDark}{swNavy}
\colorlet{arpaGreen}{swSage}
\colorlet{arpaCrimson}{swBrick}
\colorlet{arpaAmber}{swSteel}
\colorlet{arpaPurple}{swSteel}
\colorlet{arpaSlate}{swStone}
\colorlet{arpaBorder}{swLine}
\colorlet{arpaBg}{swCloud}
\colorlet{arpaCodeBg}{swPaper}
\colorlet{lightrule}{swLine}
\colorlet{graytext}{swStone}
\colorlet{accent}{swBlue}

\usepackage{amsmath, amssymb, amsthm, bm}
\theoremstyle{definition}
\newtheorem{definition}{Definition}

\usepackage{graphicx}
\usepackage{booktabs}
\usepackage{array}
\usepackage{tabularx}
\usepackage{colortbl}
\usepackage{enumitem}
\setlist{noitemsep, topsep=2pt, parsep=1pt, partopsep=1pt}
\arrayrulecolor{swNavy}
\newcolumntype{L}{>{\raggedright\arraybackslash}X}
\newcolumntype{P}[1]{>{\raggedright\arraybackslash}p{#1}}
\newcolumntype{R}[1]{>{\raggedleft\arraybackslash}p{#1}}
\newcolumntype{C}[1]{>{\centering\arraybackslash}p{#1}}


\usepackage{tikz}
\usetikzlibrary{arrows.meta, positioning, calc, fit, backgrounds,
                shapes.geometric, shapes.misc, patterns, patterns.meta}
\usepackage{pgfplots}
\pgfplotsset{compat=1.18}

\tikzset{
  sw/.style={
    font=\sffamily\footnotesize,
    >={Stealth[length=4.6pt, width=3.8pt]},
    line cap=round, line join=round,
    every node/.append style={inner sep=3pt}
  },
  swbox/.style={
    draw=#1, line width=0.7pt, fill=white, rounded corners=2pt,
    align=center
  },
  swbox/.default=swBlue,
  swtint/.style={
    draw=#1, line width=0.7pt, fill=#1!8, rounded corners=2pt,
    align=center
  },
  swtint/.default=swBlue,
  swbar/.style={
    fill=#1, text=white, font=\sffamily\bfseries\footnotesize,
    rounded corners=2pt, align=center
  },
  swbar/.default=swBlue,
  swzone/.style={fill=#1, draw=none, rounded corners=3pt},
  swzone/.default=swMist,
  swpill/.style={
    fill=white, draw=swSky, line width=0.5pt, rounded corners=1.5pt,
    font=\ttfamily\scriptsize, text=swBlue, inner sep=1.6pt
  },
  swnote/.style={font=\sffamily\scriptsize, text=swStone, align=center},
  swtag/.style={
    font=\sffamily\bfseries\scriptsize, text=#1, fill=white, inner sep=1.6pt
  },
  swtag/.default=swBlue,
  swflow/.style={draw=#1, line width=0.9pt, ->},
  swflow/.default=swBlue,
  swdash/.style={draw=#1, line width=0.8pt, densely dashed, ->},
  swdash/.default=swStone,
}

\newlist{figok}{itemize}{1}
\setlist[figok]{label=\textcolor{swSage}{\cmark}, leftmargin=1.05em,
  labelsep=0.35em, nosep, topsep=1pt}
\newlist{figbad}{itemize}{1}
\setlist[figbad]{label=\textcolor{swBrick}{\xmark}, leftmargin=1.05em,
  labelsep=0.35em, nosep, topsep=1pt}
\newlist{figdot}{itemize}{1}
\setlist[figdot]{label=\textcolor{swSky}{\textbullet}, leftmargin=0.8em,
  labelsep=0.3em, nosep, topsep=1pt}

\DeclareRobustCommand{\cmark}{\tikz[baseline=-0.1ex, x=1ex, y=1ex]
  \draw[line width=0.14ex, line cap=round, line join=round]
  (0.08,0.48) -- (0.38,0.14) -- (0.96,0.88);}
\DeclareRobustCommand{\xmark}{\tikz[baseline=-0.1ex, x=1ex, y=1ex]
  \draw[line width=0.14ex, line cap=round]
  (0.15,0.12) -- (0.85,0.82) (0.15,0.82) -- (0.85,0.12);}

\usepackage[
  font={small, color=swStone},
  labelfont={bf, color=swNavy},
  labelsep=period,
  justification=justified,
  singlelinecheck=false,
  skip=6pt
]{caption}
\usepackage{titlesec}
\titleformat{\section}
  {\normalfont\Large\bfseries\color{swNavy}}{\thesection}{0.8em}{}
\titleformat{\subsection}
  {\normalfont\large\bfseries\color{swBlue}}{\thesubsection}{0.8em}{}
\titleformat{\subsubsection}
  {\normalfont\normalsize\bfseries\color{swNavy}}{\thesubsubsection}{0.8em}{}
\titlespacing{\section}{0pt}{16pt plus 3pt minus 2pt}{6pt plus 1pt}
\titlespacing{\subsection}{0pt}{11pt plus 2pt minus 1pt}{4pt plus 1pt}
\titlespacing{\subsubsection}{0pt}{8pt plus 1pt minus 1pt}{3pt plus 1pt}

\usepackage{fancyhdr}
\fancypagestyle{firstpage}{%
  \fancyhf{}%
  \fancyhead[L]{\includegraphics[height=1.1cm]{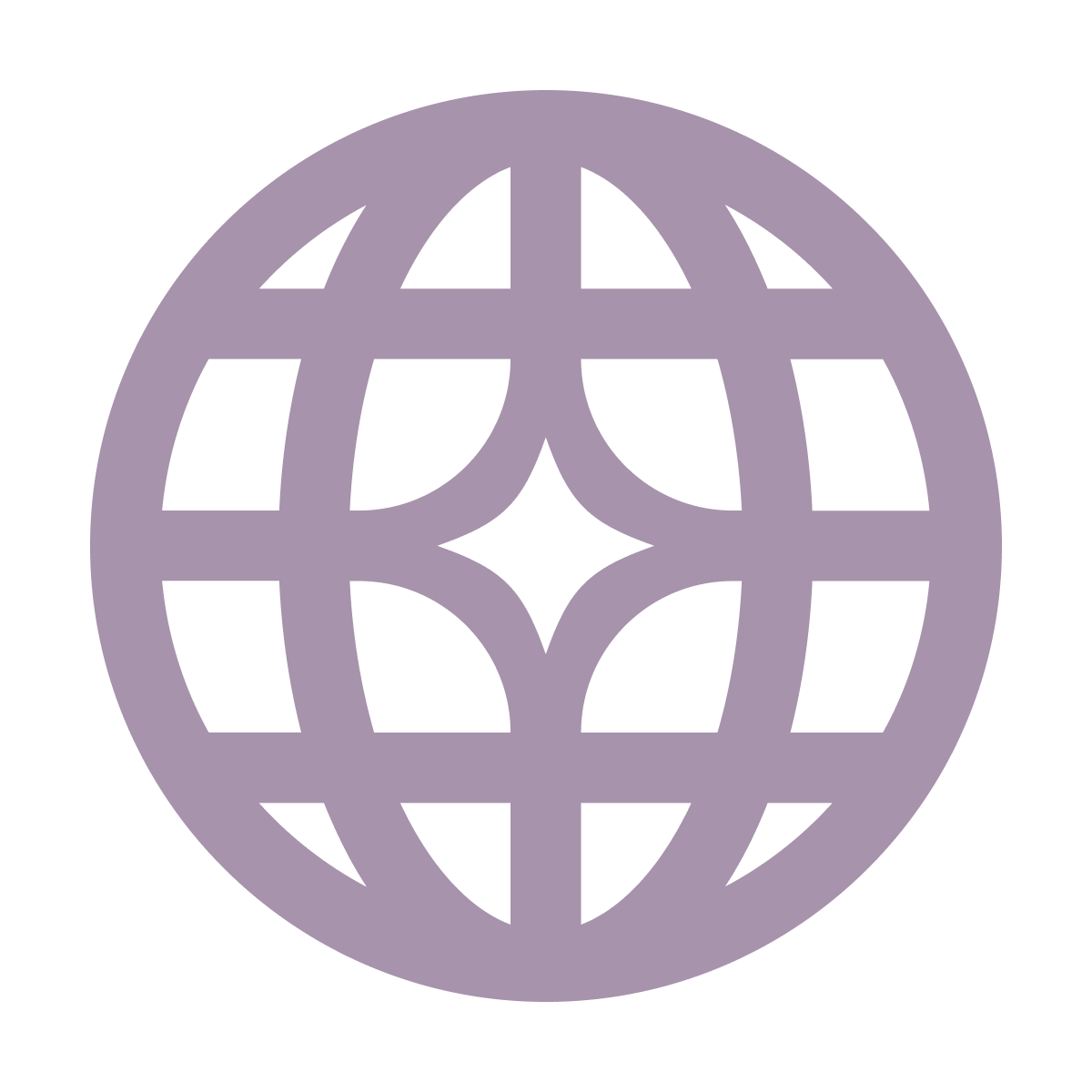}}%
  \fancyhead[R]{\small\color{swStone}ARPA Hellenic Logical Systems}%
  \fancyfoot[C]{\small\color{swStone}\thepage}%
}

\usepackage{tcolorbox}
\tcbuselibrary{skins, breakable}
\tcbset{
  swcallout/.style={
    enhanced, breakable,
    colback=#1!4!white, colframe=#1,
    boxrule=0.4pt, leftrule=2.4pt,
    arc=1.5pt, outer arc=1.5pt,
    left=9pt, right=9pt, top=7pt, bottom=7pt,
    before skip=10pt, after skip=10pt,
    fonttitle=\sffamily\bfseries\small,
    coltitle=#1,
    attach boxed title to top left={yshift=-2.2mm, xshift=7pt},
    boxed title style={colback=white, colframe=white, boxrule=0pt,
                       left=3pt, right=3pt, top=0pt, bottom=0pt}
  },
  contribbox/.style={swcallout=swBlue, title={Summary of Contributions}},
  protocolbox/.style={swcallout=swSteel},
  doctrinebox/.style={swcallout=swNavy, colback=swPaper},
  threatbox/.style={swcallout=swBrick},
}

\usepackage{upquote}
\usepackage{listings}
\lstdefinelanguage{yaml}{
  morekeywords={true,false,null},
  sensitive=true,
  comment=[l]{\#},
  morestring=[b]",
  morestring=[b]',
}
\lstdefinestyle{swcode}{
  basicstyle=\footnotesize\ttfamily\color{swNavy},
  keywordstyle=\color{swBlue}\bfseries,
  stringstyle=\color{swSage},
  commentstyle=\color{swStone}\itshape,
  backgroundcolor=\color{swPaper},
  frame=leftline,
  framerule=2pt,
  rulecolor=\color{swBlue},
  framexleftmargin=6pt,
  xleftmargin=26pt,
  xrightmargin=2pt,
  numbers=left,
  numberstyle=\scriptsize\color{swSky},
  numbersep=14pt,
  breaklines=true,
  breakatwhitespace=true,
  columns=fullflexible,
  keepspaces=true,
  showstringspaces=false,
  tabsize=2,
  aboveskip=8pt, belowskip=6pt,
  literate={~}{{\textasciitilde}}1,
}
\usepackage[colorlinks=true,
            citecolor=swBlue,
            linkcolor=swBlue,
            urlcolor=swBlue,
            breaklinks=true,
            bookmarksnumbered=true]{hyperref}

\usepackage[style=numeric-comp, backend=biber,
            sortcites=true, giveninits=true,
            maxcitenames=2, maxbibnames=99,
            doi=true, url=false, isbn=false]{biblatex}
\title{%
  \vspace{-0.6cm}%
  {\LARGE\bfseries\color{swNavy} Don't Prompt Your Agents. Equip Them.}\\[0.35em]
  {\Large\normalfont\color{swStone}\itshape
   Skillware as a Modular Standard for Sovereign, Auditable,\\
   and Deterministic Capability Execution for Logical Systems}
}

\author{%
  \textbf{Vladimiros Peilivanidis}\\[0.25em]
  {\small ARPA Hellenic Logical Systems, Thessaloniki, Greece}\\
  {\small \href{mailto:r1@arpacorp.net}{\texttt{r1@arpacorp.net}}}
  \quad{\small ORCID: \href{https://orcid.org/0009-0003-0121-465X}{0009-0003-0121-465X}}
}

\date{%
  \small November 2026\quad\textbar\quad
  {\normalfont\color{swStone}%
   \href{https://github.com/arpahls/skillware}{\texttt{github.com/arpahls/skillware}}}%
}

\begin{document}

\thispagestyle{firstpage}
\maketitle
\thispagestyle{firstpage}

\input{sections/00_abstract.tex}

\newpage

\input{sections/01_introduction.tex}
\input{sections/02_problem_layers.tex}
\input{sections/03_skillware_architecture.tex}
\input{sections/04_runtime_chaining_config.tex}

\input{sections/05_slm_vs_llm_token_economy.tex}
\input{sections/06_comparative_analysis.tex}

\input{sections/07_tripartite_ecosystem_aura_mnemolink.tex}
\input{sections/08_security_privacy_sovereignty.tex}
\input{sections/09_embodied_robotics_bmi_horizon.tex}
\input{sections/10_conclusion.tex}

\printbibliography

\end{document}

%% file: sections/00_abstract.tex

\begin{abstract}
\noindent Current autonomous logical systems rely on a fragile assumption: that language models should dynamically write code, guess API payloads, and memorize operational constraints through prompt engineering. In practice, this burns massive context windows, inflates API bills, and produces fragile agents that fail unpredictably in production.

\vspace{0.25em}
\noindent In this work, we present \textbf{Skillware}, an open architectural framework and capability registry that formally separates what an agent wants to do from how that task is executed. Instead of prompting models with loose text instructions or markdown files baptized as ``skills'', Skillware equips them with modular, versioned software capability bundles. Much like skill hotkeys in video games, which execute deterministically under fixed world rules rather than conversational negotiation, each bundle combines a typed JSON input contract, deterministic Python code, task directives, bundled datasets, and automated unit tests. In-process adapters translate these capabilities into native tool declarations for any model, from frontier cloud APIs to open-weight, local, and edge runtimes with zero network latency.

\vspace{0.25em}
\noindent Evaluating 23 frontier and open-weight models across 368 benchmark conditions, we show that small local models, even 3B to 8B parameters, equipped with Skillware consistently outperform unequipped frontier models on complex multi-step workflows. They cut token usage by up to $84.8\%$, slash inference costs by up to $\mathbf{97.3\%}$, and eliminate parameter errors entirely. True agent reliability does not come from larger models or longer prompts, but from equipping logical systems with verified, deterministic software.

\vspace{0.25em}
\noindent Finally, we articulate the trajectory of capability bundles across an invariant continuum of information processing systems. Today, Skillware serves as a modular, installable capability library for digital logical systems. Tomorrow, the exact same framework powers robust capabilities for IoT devices, smart appliances, vehicles, drones, and embodied robotics. Down the line, as brain-computer interfaces advance, this modular architecture provides the foundation to install verified capabilities directly into organic information processors (brains), turning the instant skill acquisition of science fiction into a concrete, auditable engineering discipline.
\end{abstract}

\vspace{0.35em}

%% file: sections/01_introduction.tex

\section{Introduction: The Stochastic Impedance Mismatch}
\label{sec:introduction}

Computer systems research has historically progressed by establishing formal boundaries between distinct abstraction layers: the separation of physical memory from virtual addressing, the isolation of user space from kernel execution, and the decoupling of declarative query specification from relational execution engines~\cite{saltzer1975protection, lampson1974protection, parnas1972criteria, hoare1969axiomatic}. Each transition replaced stochastic, ad-hoc programming conventions with deterministic, verifiable contracts.

The contemporary paradigm of autonomous artificial intelligence agents represents a sharp departure from this historical trajectory. As foundation models have scaled, practitioners have attempted to bridge the gap between linguistic reasoning and real-world execution by coercing auto-regressive language models into serving simultaneously as planners, compilers, runtime engines, and policy arbiters~\cite{wang2024survey, xi2023rise, yao2023react, schick2023toolformer}. In this prevailing architectural approach---which we term the \emph{prompt-driven execution paradigm}---an agent is provided with unstructured natural-language descriptions of external APIs, instructed to synthesize execution code or structured payloads on the fly, and expected to self-regulate its operational boundaries through in-context prompting.

This approach suffers from a foundational \textbf{stochastic impedance mismatch}. Deep transformer architectures compute conditional probability distributions over discrete token sequences:
\begin{equation}
\label{eq:autoregressive}
p(y_1, \dots, y_T \mid x) = \prod_{t=1}^T p(y_t \mid y_{<t}, x)
\end{equation}
While auto-regressive token generation is exceptionally well-suited for semantic interpretation, open-domain summarization, and human-machine dialogue, it possesses no formal mechanisms for enforcing computational invariants. When task complexity scales compositionally, transformer architectures experience severe combinatorial degradation, non-deterministic branching, and state hallucination~\cite{dziri2023faith, valmeekam2023planning, ji2023survey, huang2023survey}.

\subsection{The Crisis of Prompt-Driven Capability}
\label{subsec:five_flaws}

Treating auto-regressive language models as direct computation runtimes introduces five critical systemic bottlenecks:
\begin{enumerate}[leftmargin=*]
  \item \textbf{Stochastic Non-Determinism:} An identical operational intent executed across multiple turns yields variable argument casing, hallucinated parameters, and dropped structural invariants.
  \item \textbf{Quadratic Context Dilution:} Injecting thousands of tokens of API manuals and directives into system prompts permanently taxes transformer attention ($\mathcal{O}(L^2)$), causing models to ignore critical safety predicates~\cite{liu2024lost}.
  \item \textbf{Egress Insecurity and the Confused Deputy:} Natural-language prompts cannot enforce security perimeters; untrusted inbound text easily repurposes ambient model authority into unauthorized side-effects~\cite{greshake2023not, liu2023promptinjection}.
  \item \textbf{Schema Balkanization:} Fragile, incompatible tool formats across major cloud providers (e.g., Protobuf, XML, JSON) force developers to maintain brittle prompt shims that break across model releases.
  \item \textbf{Cyber-Physical Incompatibility:} Multi-second cloud latency and stochastic parameter drift are fundamentally incompatible with real-time physical control and millisecond reflex envelopes~\cite{ahn2022can, brohan2023rt2, willett2023speech}.
\end{enumerate}

These failure modes are not cosmetic developer inconveniences; they form a systemic crisis where critical software workflows become trapped inside a centralized, rented loop. Section~\ref{sec:problem_layers} provides a comprehensive technical deconstruction of these failures across enterprise, developer, macroeconomic, and cyber-physical layers.

\subsection{The Core Principle: Don't Prompt. Equip.}
\label{subsec:thesis}

In 1975, Saltzer and Schroeder articulated the foundational tenets of secure and reliable systems: \emph{Economy of Mechanism}, \emph{Fail-Safe Defaults}, \emph{Least Privilege}, and \emph{Complete Mediation}~\cite{saltzer1975protection}. In classical computing, operating systems do not generate utility binaries at runtime; they package pre-compiled, verifiable, and isolated software libraries behind typed system calls.

A compelling systems analogy emerges from role-playing video games (RPGs), where players and non-player companions personalize and customize characters through modular skill systems. In a game engine, activating an ability is never an ambiguous negotiation, a conversational plea, or a probabilistic discussion with the character model. It is a hotkey bound to deterministic engine logic: provided environmental preconditions and background policies hold true---the entity is in combat outside a sanctuary town, targeting a hostile creature rather than an immune town guard, and possessing sufficient resource reserves---the skill executes with a single trigger and produces the exact same state transition every time. The character never invents the physics of the ability at runtime, nor does it negotiate its operational boundaries; it is strictly in the position to trigger the action. Skillware introduces this exact computational discipline to autonomous agents: the language model evaluates semantic context and decides when to trigger a capability, but has zero authority over how the capability executes, how its parameters are validated, or how its constitutional policies are enforced.

\begin{tcolorbox}[doctrinebox, title={THE PRINCIPLE OF SEPARATION OF COGNITION AND CAPABILITY}]
\textbf{Formal Boundary Invariant:}\\[0.2em]
Let an autonomous system be partitioned into a \textbf{Cognitive Planner} $\mathcal{M}$ and a set of \textbf{Executable Capabilities} $\mathcal{K}$. The planner $\mathcal{M}$ is restricted strictly to semantic routing and parameter instantiation. The capability $\kappa \in \mathcal{K}$ must be a self-contained, pre-compiled, deterministic software bundle enforcing an immutable contract, local dependency isolation, and offline verification. \textbf{The model decides when to act; the capability bundle governs how the action is executed. The assistant is a guest; the function does the work.}
\end{tcolorbox}

In this paper, we present \textbf{Skillware}, an open architectural framework and registry that formalizes this separation. While the term \emph{skillware} has historical currency in corporate compliance software and speculative fiction, its coinage and introduction into artificial intelligence and autonomous systems was originated by ARPA Hellenic Logical Systems~\cite{skillware2026framework}. Recent academic literature, notably Fan and Lan~\cite{fan2026skillware}, has acknowledged this origin while exploring theoretical software ontologies for persistent agent behaviors. In this work, we present the production-validated execution standard: self-contained, versioned capability bundles. Each bundle unifies a typed contract (\texttt{manifest.yaml}), a deterministic execution class (\texttt{skill.py}), a cognitive directive (\texttt{instructions.md}), offline assurance verifiers (\texttt{test\_skill.py}), and metadata presentation (\texttt{card.json}). 

Through in-process dynamic adapters, Skillware translates declared capabilities into native host formats across all major frontier and local models without IPC or network overhead. We demonstrate that this architecture not only restores determinism to the execution layer of autonomous systems, but enables Small Language Models (SLMs) to achieve superior operational reliability at a fraction of the token and financial cost of unequipped frontier models.

\begin{tcolorbox}[contribbox]
\begin{itemize}[leftmargin=2.2em, itemsep=2.5pt, labelsep=0.5em]
  \item[\textbf{\textcolor{swBlue}{C1.}}] \textbf{The Separation Invariant} (\S\ref{subsec:thesis}): We formalize the separation of cognitive planning from deterministic execution and show why runtime code generation and prompt-only tools fail production safety and economics.
  \item[\textbf{\textcolor{swBlue}{C2.}}] \textbf{The 5-Role Capability Bundle} (\S\ref{sec:architecture}): We specify the anatomy of an Agent Skill (Contract, Effect, Directive, Assurance, Presentation), supported by static lookup corpora, enabling offline verification and type safety.
  \item[\textbf{\textcolor{swBlue}{C3.}}] \textbf{Universal In-Process Adapters} (\S\ref{subsec:interface_morphism}): We provide zero-IPC interface morphisms across Gemini, Claude, OpenAI, DeepSeek, Bedrock, and Ollama, eliminating vendor lock-in and client-server daemon overhead.
  \item[\textbf{\textcolor{swBlue}{C4.}}] \textbf{Deterministic Chaining Engine} (\S\ref{sec:runtime_operations}): We introduce declarative DAG execution with dot-notation state bindings and zero-token conditional skip gates, accompanied by segregated local operator stores.
  \item[\textbf{\textcolor{swBlue}{C5.}}] \textbf{The SLM Parameter Paradox} (\S\ref{sec:slm_token_economy}): We show that 3B--8B parameter models equipped with Skillware outpace unequipped frontier models in execution reliability, latency, cost ($97.3\%$ reduction), and air-gapped sovereignty.
  \item[\textbf{\textcolor{swBlue}{C6.}}] \textbf{Generalization Across Physical Scales} (\S\ref{sec:embodied_bmi_horizon}): We formulate the scaling of capability bundles from digital software runtimes to industrial control, embodied robotics, and neural interfaces.
\end{itemize}
\end{tcolorbox}

The remainder of the paper is organized as follows. Section~\ref{sec:problem_layers} dissects the capability crisis across technical layers and constituencies. Section~\ref{sec:architecture} formalizes the Skillware bundle and its interface morphism; Section~\ref{sec:runtime_operations} describes runtime discovery, chaining, and operator stores. Section~\ref{sec:slm_token_economy} analyzes complexity, empirical scaling, and token economics, and Section~\ref{sec:comparative_analysis} positions Skillware against adjacent ecosystems alongside multi-model empirical evaluation under \textsc{Skillware-Bench}. Sections~\ref{sec:tripartite_ecosystem} and~\ref{sec:security_privacy} cover ecosystem integration and adversarial security verification, Section~\ref{sec:embodied_bmi_horizon} extends the abstraction to cyber-physical scales, and Section~\ref{sec:conclusion} concludes.

%% file: sections/02_problem_layers.tex

\section{Deconstructing the Capability Crisis Across Systemic Layers and Constituencies}
\label{sec:problem_layers}

The failure of prompt-driven capabilities is not an isolated developer inconvenience; it is an architectural breakdown that propagates across the entire computing stack. To rigorously motivate the design of Skillware, we first analyze the failure modes across four technical layers, examine three insidious non-outage operational crises, and then demonstrate how Skillware resolves these challenges for four vital computing constituencies: software developers, autonomous systems, enterprise operators, and sovereign defense entities.

\subsection{The Four Technical Failure Layers of Prompted Agents}
\label{subsec:technical_layers}

\subsubsection{The Enterprise and Regulatory Layer: The Illusion of Compliance}
In regulated industries---including banking, defense logistics, healthcare, and telecommunications---\allowbreak autonomous operations are bound by statutory compliance regimes such as SOX, HIPAA, GDPR, and OFAC sanctions enforcement. 

When an agent's capability is enforced solely through natural-language system instructions (e.g., \emph{``You are a compliance officer. Check if this Ethereum address is on the OFAC sanctions list before approving transaction''}), the system lacks any defensible guarantee of execution. An auto-regressive model may hallucinate that an address is benign, misparse hexadecimal character casing, or fail to access real-time sanctions lists due to network timeouts. Furthermore, irreversible side-effects (such as submitting an on-chain transaction or dispatching an email to 10,000 corporate customers) are executed through direct, unverified model outputs.

Enterprise sovereignty demands:
\begin{itemize}[leftmargin=*]
  \item \textbf{Inviolable Confirmation Gates:} Destructive mutations must require explicit, typed boolean gates (\texttt{confirmed: true}) and preview dry-runs (\texttt{dry\_run: true}) enforced at the code level, completely independent of the model's textual disposition.
  \item \textbf{Deterministic Data Provenance:} Critical reference datasets (such as sanctions lists or regulatory taxonomies) must ship bundled beside the execution logic as static, offline assets, eliminating runtime scraping and dynamic tampering.
\end{itemize}

\subsubsection{The Developer Layer: Schema Fragmentation and Prompt Rot}
For software engineers, maintaining prompt-based tool definitions is an operational nightmare known as \emph{prompt rot}. 

First, foundation model providers lack standardization. Google Gemini requires uppercase Protobuf type declarations (\texttt{STRING}, \texttt{INTEGER}, \texttt{OBJECT}); Anthropic Claude requires its own tool envelope with a JSON Schema \texttt{input\_schema}; OpenAI requires its Chat Completions function signatures~\cite{anthropic2024mcp}. When an engineering team updates an API or switches from Claude to Gemini, hundreds of lines of fragile prompt shims must be rewritten and re-evaluated.

Second, prompt-driven tools violate standard software engineering lifecycle guarantees. There is no concept of a unit test for an instruction string. In contrast, standard software undergoes deterministic continuous integration: unit tests, regression suites, type validation, and dependency pinning. Prompt-based agents deploy completely unverified code directly into execution paths.

Finally, package management in agent stacks is broken. Equipping an agent with complex capabilities (e.g., cryptographic hashing, EVM RPC handling, or PDF parsing) often pollutes the global application environment, triggering severe dependency conflicts.

\subsubsection{The Macroeconomic Layer: The Token Burn of Disposable Logic}
From an economic perspective, the current industry approach is profoundly unsustainable. When an agent is prompted to solve a task through dynamic code generation or instruction following:
\begin{itemize}[leftmargin=*]
  \item \textbf{Frontier Token Rates for Static Logic:} Operators pay upwards of \$3.00 to \$15.00 per million output tokens for an LLM to re-synthesize common Python logic that has been solved for decades (e.g., regex sanitization, IMAP parsing, or JSON transformations).
  \item \textbf{Quadratic Attention Tax:} Stuffing multiple tool schemas and voluminous OpenAPI manifests into the context window consumes thousands of prompt tokens per turn. Over a 50-turn conversational session, an agent can easily consume \$20.00 in input tokens simply re-reading the identical tool documentation on every message.
\end{itemize}

Pre-compiled capabilities eliminate this overhead entirely. The model emits a compact JSON function call (e.g., 20 tokens), the host executes the deterministic Python code locally with zero token cost, and the structured response is returned for semantic synthesis.

\subsubsection{The Cyber-Physical Layer: Determinism in Real Space}
The final frontier of autonomy lies outside data centers: in autonomous quadcopters navigating contested airspace, industrial robotic arms assembling microelectronics, and neural prosthetic interfaces restoring motor function to paralyzed patients~\cite{brohan2022rt1, willett2023speech}.

In cyber-physical systems, failure is not a low-confidence text response; it is physical collision, equipment destruction, or physiological trauma. Such systems cannot tolerate 2.5-second cloud round-trip latencies, stochastic API argument hallucination, or network dependency. Physical agency requires \textbf{instantaneous, deterministic, and sandboxed execution} that can run entirely offline on low-power edge compute (such as an NVIDIA Jetson or ARM Cortex SoC).

\subsection{The Chronic Crises of Rented Intelligence: Infrastructure Cascades and Hidden Penalties}
\label{subsec:quieter_crises}

While catastrophic infrastructure outages make headlines and populate vendor status pages, four chronic operational penalties undermine production autonomy behind closed doors:

\begin{enumerate}[leftmargin=*]
  \item \textbf{The Rented Loop and Infrastructure Cascades:} Historically, software development was architecturally decentralized: Git repositories lived locally on disk, and remote servers were merely optional publishing targets. Today, that relationship has inverted into a \emph{rented loop}: editor completions, pull request reviews, continuous integration, and runtime capabilities depend on centralized model endpoints sharing a common failure domain. 
  
  When an upstream cloud infrastructure degrades---\allowbreak exemplified by the August 2026 cascading incident where sidecar proxy exhaustion, auth collapse, and client-side retry storms paralyzed repositories, CI pipelines, and Copilot on the identical incident clock---\allowbreak the entire cognitive workflow halts. Competing chat interfaces and cloud APIs frequently share identical geographical corridors and cooling zones. When the pipe darkens, an enterprise does not merely experience latency; it loses the digital hands it hired. Real operational autonomy requires that capability execution reside on local silicon, independent of external network availability.
  \item \textbf{Subscribed Cognition vs. Amortized Silicon:} Operators do not control the cost of automated thinking; they subscribe to it. Token pricing shifts at the whim of foundation model providers seeking margin expansion or sunsetting promotional tiers. A model executing on local silicon incurs predictable electricity consumption; a cloud-rented model sends an altered monthly invoice because a line changed in a vendor spreadsheet.
  \item \textbf{Geopolitical and Corporate Blockades:} Frontier model access is increasingly politicized and weaponized through geographic blacklists and corporate parentage audits. Developers and enterprises in disfavored jurisdictions can be severed from mission-critical tools overnight without an announcement or a dependency lockfile bump---exemplified by silent platform lockouts affecting millions of global developers under the guise of infrastructure rollouts. When capability is bound to a remote cloud identity, sovereignty is an illusion.
  \item \textbf{Unpinned Drift and Closed-Weight Amnesia:} In production software, dependencies are pinned to immutable commit hashes. In cloud AI, foundation model endpoints are continuously rerouted, compressed, and safety-tuned behind static API aliases without public version control. Developers across the industry repeatedly document sudden degradation: models become taciturn, omit procedural steps, violate negative constraints, or hallucinate task completion. Because proprietary cloud weights cannot be diffed, operators cannot inspect, reproduce, or rollback behavioral regressions.
\end{enumerate}

\subsection{Why Skillware is Better for the Four Core Constituencies}
\label{subsec:four_constituencies}

Having examined the structural layers of failure and the chronic crises of rented cognition, we synthesize how Skillware transforms operational stability across the four core computing constituencies:

\subsubsection{Software Developers: Eliminating Prompt Rot and Restoring Offline Verification}
\begin{itemize}[leftmargin=*]
  \item \textbf{Elimination of ``Prompt Rot'':} In prompt-based architectures, minor adjustments to a system prompt frequently break tool invocation schemas across unrelated domains. Skillware encapsulates each capability behind a typed JSON Schema in \texttt{manifest.yaml}. The tool signature is decoupled from prompt phrasing and validated programmatically at compile- and run-time.
  \item \textbf{Offline Testability} (\texttt{test\_skill.py}): Standard agent frameworks cannot be unit-tested without calling third-party LLM APIs, incurring latency, monetary costs, and stochastic results. Every Skillware bundle ships with a comprehensive offline test suite (\texttt{test\_skill.py}) that verifies that \texttt{execute()} honors \texttt{manifest.yaml} independently of any model invocation.
  \item \textbf{Isolated Package Management:} Complex capabilities require distinct external libraries (e.g., \texttt{web3} for blockchain validation, \texttt{requests} for REST calls, \texttt{fsspec} for cloud storage). Installing a monolithic agent framework frequently results in dependency hell. Skillware enforces granular dependency declarations per skill, mapped automatically to pip extras (e.g., \texttt{pip install "skillware[defi\_evm\_tx\_handler]"}). Missing packages are detected pre-flight, guiding the developer with precise installation hints.
\end{itemize}

\subsubsection{Autonomous Logical Systems: Bounded State and Safe Asynchronous Execution}
\begin{itemize}[leftmargin=*]
  \item \textbf{Pre-Flight Parameter Invariants:} In a prompt-driven loop, a model hallucinating an invalid argument type (e.g., passing a string where an integer timestamp is required) passes directly into underlying system APIs, causing unhandled runtime crashes. Skillware's \texttt{BaseSkill.\allowbreak validate\_params()} intercepts inputs at the interface boundary, catching schema violations and raising structured, machine-readable \texttt{Skillware\allowbreak Param\allowbreak Validation\allowbreak Error} exceptions that allow the agent to self-correct without crashing.
  \item \textbf{Deterministic Chaining vs. Multi-Agent Banter:} When orchestrating multi-step workflows, popular frameworks (e.g., CrewAI, AutoGen) spawn multiple conversing LLM agents~\cite{wu2023autogen, mourato2023crewai}. This approach suffers from semantic drift, high latency, and compounding errors. Skillware provides a deterministic directed acyclic graph (DAG) chaining engine (\texttt{skillware.chains}) with explicit state bindings (\texttt{host.}, \texttt{next.}, \texttt{prev.}) and conditional evaluation gates (\texttt{when:}), executing complex pipelines with zero conversational overhead.
  \item \textbf{Asynchronous Non-Blocking Execution:} Real-world systems run high-concurrency event loops. Skillware natively implements \texttt{aexecute()}, automatically offloading synchronous blocking calls to worker threads while preserving native non-blocking coroutine support for I/O-intensive tools.
\end{itemize}

\subsubsection{Enterprise Operators: Deterministic Logic, Confirmation Gates, and Secret Isolation}
\begin{itemize}[leftmargin=*]
  \item \textbf{Zero Unreviewed Code Execution:} Under Skillware, an agent never writes code on the fly to interact with sensitive databases. All operational logic resides in pre-compiled, peer-reviewed, static Python files. The model's authority is constrained strictly to selecting the appropriate capability and parameterizing it.
  \item \textbf{Inviolable Confirmation and Dry-Run Gates:} For irreversible mutations (e.g., executing high-value bank transfers or dispatching executive communications), Skillware skills implement mandatory boolean validation gates (\texttt{confirmed: true}) and preview modes (\texttt{dry\_run: true}). If the model omits confirmation, the capability refuses execution at the code level, regardless of how persuasively the agent was prompted.
  \item \textbf{Zero-Leak Secret Management:} Storing credentials in process-global environment variables (\texttt{os.environ}) exposes keys to accidental leakage via exception tracebacks or unauthorized memory inspection. Skillware's \texttt{SecretProvider} protocol (\texttt{EnvSecretProvider}, \texttt{MappingSecretProvider}, \texttt{CallableSecretProvider}) injects credentials directly into the skill instance via \texttt{BaseSkill.credential()}, allowing hosts to bridge enterprise KMS or HashiCorp Vault backends with zero global state pollution.
\end{itemize}

\subsubsection{Sovereign Infrastructure: Air-Gapped Autonomy, Supply-Chain Provenance, and Forensic Auditability}
\begin{itemize}[leftmargin=*]
  \item \textbf{Air-Gapped Sovereignty:} Contemporary cloud agent APIs require streaming operational telemetry, military communications, and critical infrastructure telemetry to foreign-controlled data centers. Skillware skills, when paired with locally hosted Small Language Models (SLMs), execute 100\% offline on sovereign hardware with zero external network connectivity.
  \item \textbf{Issuer Attribution and Supply-Chain Visibility:} In high-assurance defense applications, third-party software must be attributable. Every Skillware manifest declares its \texttt{issuer} (name, GitHub handle, organization), and the loader resolves skills through an explicit tier order (\texttt{project} $\succ$ \texttt{external} $\succ$ \texttt{bundled}) with automated shadow-conflict detection, so that any skill overriding a maintainer-reviewed bundled skill is surfaced to the operator rather than silently substituted.
  \item \textbf{Non-Repudiable Auditing:} Through its synergy with AURA Harness (Section~\ref{sec:tripartite_ecosystem}), Skillware operations emit SHA-256 hash-chained JSONL audit receipts, providing tamper-evident forensic logs that record what actions an autonomous system took, when, and under what policy constraints.
\end{itemize}

%% file: sections/03_skillware_architecture.tex

\section{The Skillware Architecture: Formal Specification and In-Process Adaptation}
\label{sec:architecture}

\begin{figure}[t]
\centering
\includegraphics[width=0.96\linewidth]{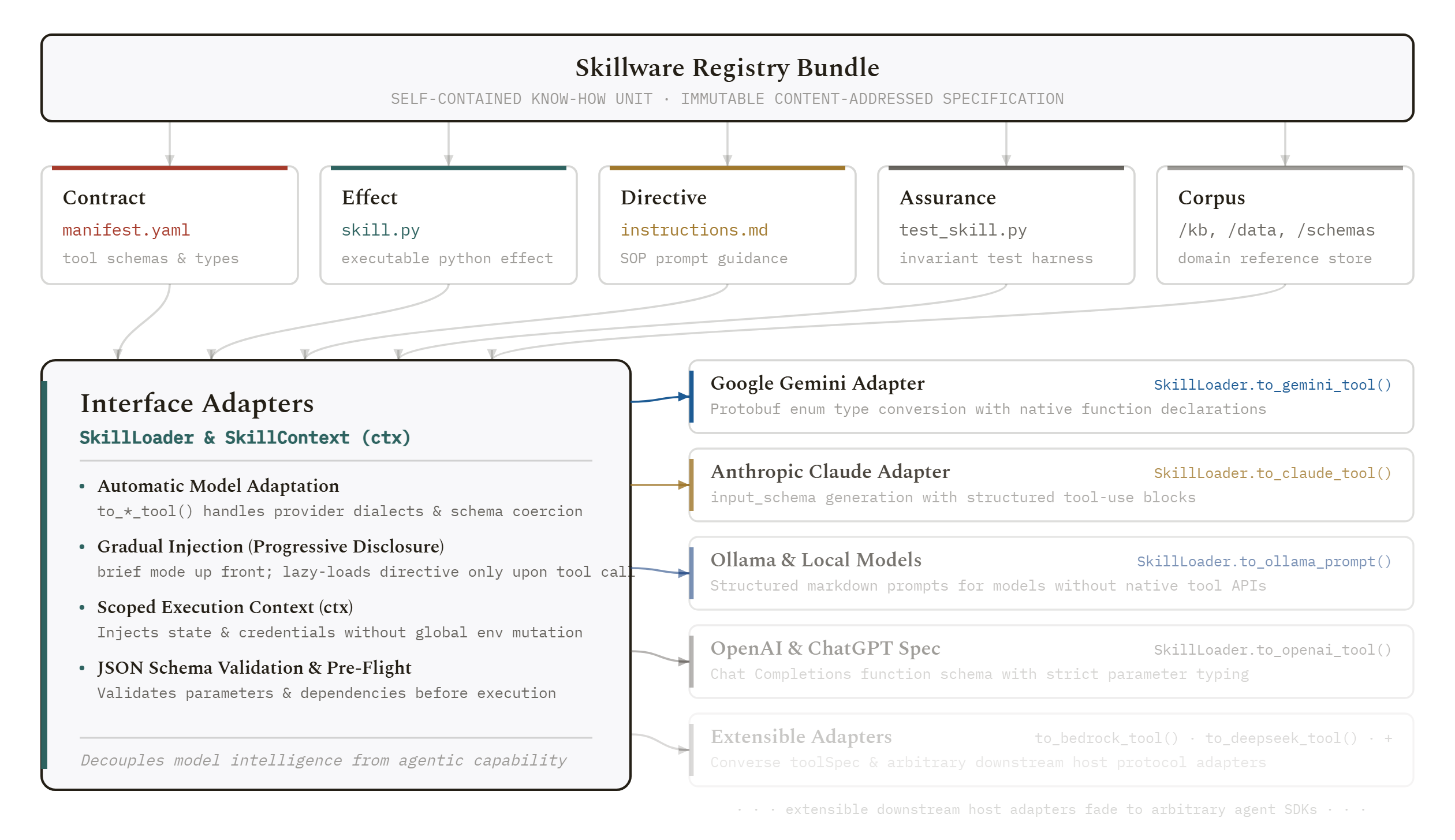}
\caption{The Skillware ecosystem architecture. Self-contained registry bundles decompose capability know-how into five first-class primitives (Contract, Effect, Directive, Assurance, Corpus), executed in-process with automated host adapter transmutation.}
\label{fig:skillware_anatomy}
\label{fig:ecosystem_architecture}
\end{figure}

To replace informal prompt conventions with a mathematically verifiable foundation, Skillware formalizes an \textbf{Agent Skill} as a self-contained, version-controlled filesystem bundle that encapsulates code, contracts, instructions, and tests into a unified, installable unit (Figure~\ref{fig:ecosystem_architecture}).

\subsection{Formal Mathematical Definition of an Agent Skill}
\label{subsec:formal_definition}

\begin{definition}[Agent Skill Bundle]
An Agent Skill $\mathcal{K}$ is defined as a 5-tuple:
\begin{equation}
\label{eq:skill_tuple}
\mathcal{K} = \langle \mathcal{C}, \mathcal{E}, \mathcal{D}, \mathcal{A}, \mathcal{P} \rangle
\end{equation}
where:
\begin{enumerate}[leftmargin=*]
  \item $\mathcal{C} = \langle \mathcal{S}_{\text{in}}, \mathcal{S}_{\text{out}}, \Pi_{\text{const}}, \mathcal{I}_{\text{iss}}, \mathcal{R}_{\text{dep}} \rangle$ is the \textbf{Contract} (\texttt{manifest.yaml})~\cite{hoare1969axiomatic, skillware2026framework}, comprising the input parameter JSON Schema $\mathcal{S}_{\text{in}}$, the structured output schema $\mathcal{S}_{\text{out}}$, the immutable constitutional boundary predicates $\Pi_{\text{const}}$, the cryptographic publisher identity $\mathcal{I}_{\text{iss}}$, and dependency specifications $\mathcal{R}_{\text{dep}}$. In Skillware v0.6.0, contract validity is enforced via an internal meta-schema engine (\texttt{skillware.core.\allowbreak manifest\_schema}) that pre-validates action signatures, parameter types, and security constraints prior to runtime ingestion.
  \item $\mathcal{E}: \mathcal{S}_{\text{in}} \times \Theta_{\text{env}} \to \mathcal{S}_{\text{out}} \cup \{\bot\}$ is the deterministic \textbf{Effect} (\texttt{skill.py}), a state transition mapping valid input arguments under environment secrets $\Theta_{\text{env}}$ to typed output envelopes, or failing closed to an error state $\bot$.
  \item $\mathcal{D} \in \Sigma^*$ is the cognitive \textbf{Directive} (\texttt{instructions.md}), a structured natural-language guide instructing the foundation model on operational preconditions, output semantics, and cognitive heuristics.
  \item $\mathcal{A}: (\mathcal{E} \times \mathcal{C}) \to \{0, 1\}$ is the offline \textbf{Assurance} verifier (\texttt{test\_skill.py}), an executable predicate guaranteeing that the Effect satisfies the Contract independently of external model APIs:
  \begin{equation}
  \label{eq:assurance}
  \mathcal{A}(\mathcal{E}, \mathcal{C}) = 1 \iff \forall p \in \mathcal{S}_{\text{in}}, \quad \mathcal{E}(p, \Theta) \in \mathcal{S}_{\text{out}} \;\land\; \Pi_{\text{const}}(\mathcal{E}, p) = \text{true}
  \end{equation}
  \item $\mathcal{P}$ is the \textbf{Presentation} descriptor (\texttt{card.json}), providing metadata for UI catalogs, registry discovery, and operator dashboards.
\end{enumerate}
\end{definition}

Optional assets extend the bundle: a static \textbf{Corpus} $\mathcal{B}_{\text{data}}$ (\texttt{kb/}, \texttt{data/}) containing local authoritative lookup datasets, and machine-readable \textbf{References} $\mathcal{R}_{\text{ref}}$ (\texttt{schemas/}) providing auxiliary domain taxonomies.

\subsection{The Five Core Roles of a Skill Bundle}
\label{subsec:bundle_roles}

\begin{figure}[t]
\centering
\includegraphics[width=0.96\linewidth]{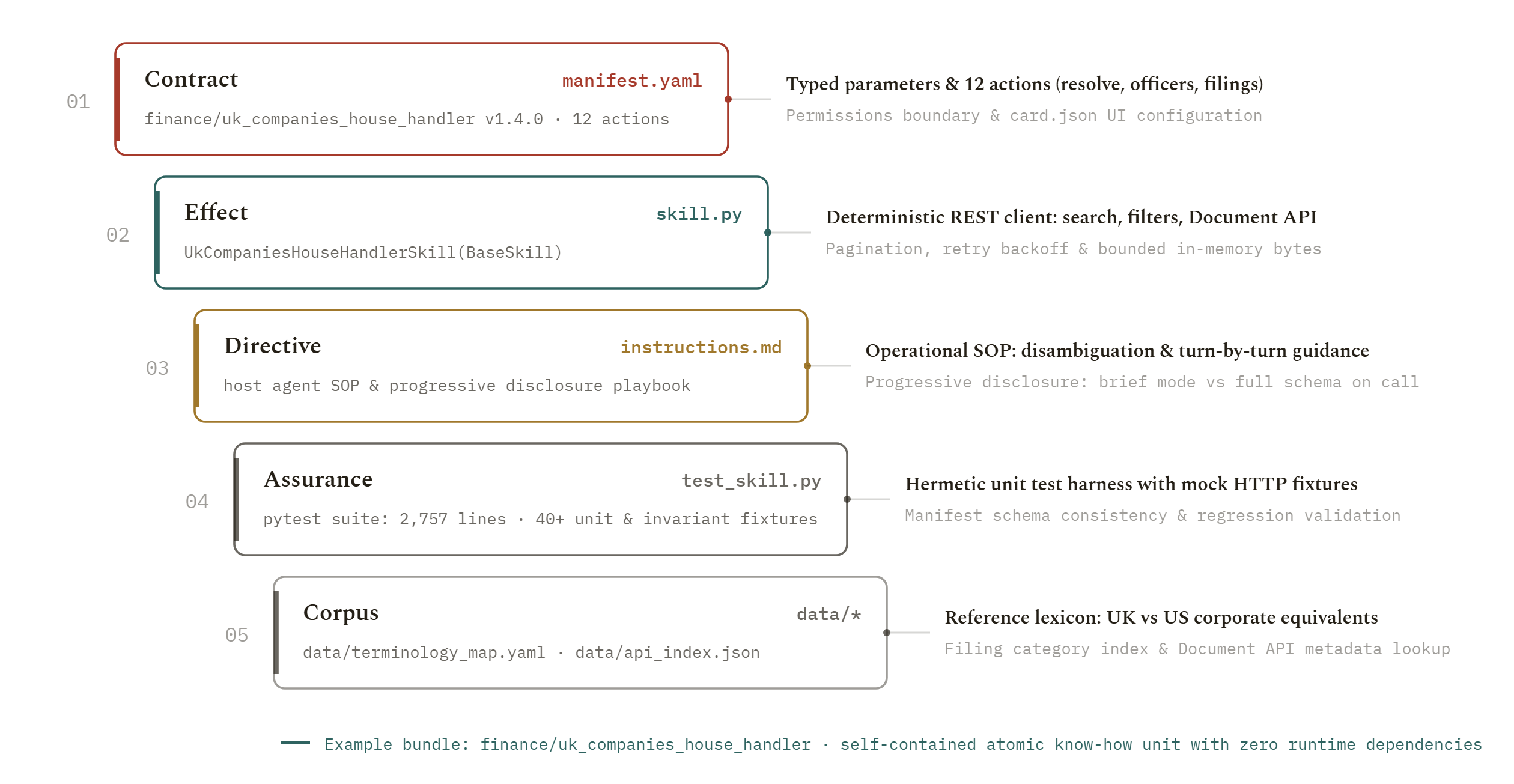}
\caption{Anatomy of a concrete Skillware bundle (\texttt{finance/uk\_companies\_house}): Contract (\texttt{manifest.yaml}), Effect (\texttt{skill.py}), Directive (\texttt{instructions.md}), Assurance (\texttt{test\_skill.py}), and Corpus (\texttt{data/}) with zero external API dependencies.}
\label{fig:bundle_anatomy}
\end{figure}

Every skill registered within the Skillware ecosystem resides within a categorized directory (\texttt{skills/<category>/<skill\_name>/}) and satisfies five distinct foundational roles implemented via fixed filenames (Figure~\ref{fig:bundle_anatomy}):

\begin{enumerate}[leftmargin=*]
  \item \textbf{Contract} (\texttt{manifest.yaml}): The single source of truth defining the capability's external boundary. It specifies JSON Schema-compliant input parameters (\texttt{parameters:}), structured return types (\texttt{returns:}), execution prerequisites (\texttt{requires:}), publisher identity (\texttt{issuer:} containing name, email, GitHub handle, and organization), and an immutable \emph{Constitution}---a set of hard rules defining what the capability will and will not do (e.g., \emph{``Offline only; no external network transmission''}, \emph{``Never execute transfers without explicit confirmed flag''}).
  \item \textbf{Effect} (\texttt{skill.py}): The deterministic runtime engine. Inheriting from the framework's \texttt{BaseSkill} class, this module implements synchronous \texttt{execute(params)} and non-blocking asynchronous \texttt{aexecute(params)} entry points. It contains the real, pre-reviewed Python logic---ranging from cryptographic ledger operations and financial compliance to biological sequence processing (\texttt{bio/sequence\_ops}) and specialized numerical routines---completely insulated from stochastic LLM code generation.
  \item \textbf{Directive} (\texttt{instructions.md}): The cognitive operational manual. This document is injected into the model's system context, teaching the reasoning engine \emph{when} to call the skill, how to interpret its JSON responses, and where its operational boundaries lie, without usurping the host agent's global personality.
  \item \textbf{Assurance} (\texttt{test\_skill.py}): The automated verification suite. Every skill must ship with comprehensive offline tests ensuring that the Effect honors the Contract. These tests are executed automatically via the CLI (\texttt{skillware test}) and in continuous integration before any skill is admitted into the registry.
  \item \textbf{Presentation} (\texttt{card.json}): UI and registry catalog metadata. It defines visual icons, branding palettes, and human-readable badges, ensuring seamless integration into visual dashboards and agent discovery marketplaces.
  \item \textbf{Corpus and References} (\texttt{kb/}, \texttt{data/}, \texttt{schemas/}): Bundled static assets, such as offline sanctions lists, geographic boundary databases, or MIME schemas, enabling self-contained, air-gapped execution without external scraping.
\end{enumerate}

Listing~\ref{lst:manifest_example} illustrates a representative contract from the verified registry:

\begin{lstlisting}[language=yaml, caption={Concrete Skillware Contract (\texttt{manifest.yaml}) for \texttt{finance/wallet\_screening}.}, label={lst:manifest_example}]
id: finance/wallet_screening
name: Wallet Sanctions & Risk Screener
version: 0.4.1
issuer:
  name: ARPA Research Division
  org: ARPA Hellenic Logical Systems
  github: arpahls
requires:
  env: [ETHERSCAN_API_KEY]
  extras: [finance_wallet_screening]
constitution:
  - "Offline first: OFAC dataset is bundled statically in kb/"
  - "Never transmit private keys or query unauthorized RPC endpoints"
  - "Fail closed on malformed hex addresses"
parameters:
  type: object
  required: [wallet_address]
  properties:
    wallet_address:
      type: string
      pattern: "^0x[a-fA-F0-9]{40}$"
      description: "EVM address to screen."
    deep_scan:
      type: boolean
      default: false
      description: "Perform recursive counterparty tracing."
returns:
  type: object
  properties:
    is_sanctioned: {type: boolean}
    risk_score: {type: number}
    flags: {type: array, items: {type: string}}
\end{lstlisting}

\begin{tcolorbox}[protocolbox, title={DEVELOPER ERGONOMICS: THE 5-MINUTE CAPABILITY BUNDLE}]
\small
A frequent hesitation among engineering teams adopting formal architectures is the concern over authoring overhead. Skillware resolves this by packaging a production-ready scaffolding toolkit within the framework repository:
\begin{itemize}[leftmargin=*]
  \item \textbf{Instant Scaffolding (\texttt{templates/python\_skill/}):} Developers do not author capability bundles from scratch. The repository provides turnkey templates for \texttt{manifest.yaml}, \texttt{skill.py}, \texttt{instructions.md}, \texttt{test\_skill.py}, and \texttt{card.json}. An engineer simply adapts the identifiers, implements the deterministic business logic in \texttt{execute()}, and binds the parameter JSON Schema.
  \item \textbf{Hermetic Offline Assurance (\texttt{skillware test}):} Before deployment, the developer verifies the bundle locally via the CLI: \texttt{skillware test <skill\_name>}. The test harness executes hermetic \texttt{pytest} fixtures, validating parameter conformance, exception fail-closed semantics, and constitutional gate enforcement in $<100\,\text{ms}$ with \textbf{zero third-party LLM API invocations}.
  \item \textbf{Sub-2ms Progressive Discovery (\texttt{skillware find}):} To prevent context bloat across massive registries, the CLI and runtime engine provide \texttt{SkillMatcher}, an in-process search index that scores query intent against manifest titles, descriptions, and action tags in $<2\,\text{ms}$. Operators find and equip capabilities dynamically without injecting the global registry into model context.
\end{itemize}
A 5-minute authoring investment yields permanent operational stability: runtime script crashes collapse to zero, active model context bloat shrinks by $84.0\%$, and per-task operational expenditure drops by $97.3\%$.
\end{tcolorbox}

\subsection{The BaseSkill Class and Parameter Invariants}
\label{subsec:baseskill_contract}

Every capability subclassing \texttt{BaseSkill} inherits an inviolable parameter enforcement mechanism. Listing~\ref{lst:baseskill} shows the core implementation from \texttt{skillware.core.\allowbreak base\_skill}:

\begin{lstlisting}[language=Python, caption={Core implementation of the \texttt{BaseSkill} contract.}, label={lst:baseskill}]
class BaseSkill(ABC):
    def __init__(self, config=None, credential_fn=None):
        self.config = config or {}
        self._credential_fn = credential_fn

    def credential(self, key: str) -> Optional[str]:
        """Resolves env_vars: credential_fn > config > os.environ."""
        if self._credential_fn is not None:
            val = self._credential_fn(key)
            if val is not None and str(val).strip():
                return str(val).strip()
        if key in self.config:
            return str(self.config[key]).strip() or None
        return os.environ.get(key)

    def validate_params(self, params: Dict[str, Any]) -> bool:
        """Validates tool arguments against manifest JSON Schema."""
        schema = self.manifest.get("parameters")
        if not schema or not isinstance(schema, dict):
            return True
        if not isinstance(params, dict):
            raise SkillwareParamValidationError(
                f"Expected dict, got {type(params).__name__}."
            )
        try:
            jsonschema.validate(instance=params, schema=schema)
        except ValidationError as exc:
            path = ".".join(str(p) for p in exc.absolute_path) or "(root)"
            raise SkillwareParamValidationError(
                f"Validation failed at '{path}': {exc.message}"
            ) from exc
        return True

    @abstractmethod
    def execute(self, params: Dict[str, Any]) -> Any:
        """Synchronous deterministic execution entry point."""
        pass

    async def aexecute(self, params: Dict[str, Any]) -> Any:
        """Native async execution; offloads sync execute() safely."""
        return await asyncio.to_thread(self.execute, params)

    @staticmethod
    def _run_coroutine_sync(coro: Any) -> Any:
        """Safely executes an async coroutine synchronously."""
        try:
            loop = asyncio.get_running_loop()
        except RuntimeError:
            loop = None
        if loop is not None and loop.is_running():
            with ThreadPoolExecutor(max_workers=1) as ex:
                return ex.submit(asyncio.run, coro).result()
        return asyncio.run(coro)
\end{lstlisting}

Notice the defensive invariants embedded directly within \texttt{BaseSkill}:
\begin{itemize}[leftmargin=*]
  \item \textbf{Schema Pre-flight Validation:} When a model emits a tool call, \texttt{validate\_params()} intercepts the arguments against the manifest's JSON Schema before any business logic executes. If the model hallucinates a malformed type, execution immediately halts with an explainable \texttt{Skillware\allowbreak Param\allowbreak Validation\allowbreak Error}, protecting underlying systems from corrupt state.
  \item \textbf{Safe Async/Sync Bridging:} Recognizing that modern agent runtimes alternate between synchronous script loops and high-concurrency \texttt{asyncio} event loops, \texttt{BaseSkill} provides an internal \texttt{\_run\_coroutine\_sync()} bridge that safely executes coroutines even within pre-existing event loops via isolated thread pools.
\end{itemize}

\subsection{Universal Interface Morphism}
\label{subsec:interface_morphism}

A primary source of developer fatigue is the need to maintain separate tool declarations for every foundation model API. Skillware solves this through the mathematical concept of an \textbf{Interface Morphism} $\Phi_{\text{host}}$:
\begin{equation}
\label{eq:morphism}
\Phi_{\text{host}}: \mathcal{C} \longrightarrow \mathcal{T}_{\text{host}}
\end{equation}
which transmutes the unified Contract $\mathcal{C}$ into the target host's specific type signature $\mathcal{T}_{\text{host}}$ entirely in-process:

\begin{enumerate}[leftmargin=*]
  \item \textbf{Google Gemini ($\Phi_{\text{gemini}}$):} Google's \texttt{google-genai} SDK requires uppercase Protobuf type declarations (e.g., \texttt{"type": "STRING"} rather than standard JSON Schema's \texttt{"type": "string"}). \texttt{SkillLoader.\allowbreak to\_gemini\_tool()} recursively traverses the properties tree, sanitizing tool names and upper-casing all primitive and nested types to produce valid \texttt{types.Tool(function\_declarations=[\dots])} objects.
  \item \textbf{Anthropic Claude ($\Phi_{\text{claude}}$):} Transmutes the schema into Claude's \texttt{tools} specification, enforcing alphanumeric name sanitization and embedding properties in \texttt{input\_schema}.
  \item \textbf{OpenAI \& DeepSeek ($\Phi_{\text{openai}}, \Phi_{\text{deepseek}}$):} Maps the contract to Chat Completions function signatures (\texttt{"type": "function", "function": \{\dots\}}).
  \item \textbf{AWS Bedrock Converse ($\Phi_{\text{bedrock}}$):} Packages parameters into Bedrock's \texttt{toolSpec} JSON schema envelopes.
  \item \textbf{Ollama Local Models ($\Phi_{\text{ollama}}$):} For edge models lacking native function-calling APIs, \texttt{to\_ollama\_prompt()} compiles the schema into clean, deterministic markdown specifications that small open-weights models (e.g., Llama-3-8B) can reliably interpret and execute.
\end{enumerate}

Crucially, because $\Phi_{\text{host}}$ is executed in-process through native Python class instances, it introduces \textbf{zero network serialization latency, zero IPC sockets, and zero inter-process dependencies}.

%% file: sections/04_runtime_chaining_config.tex

\section{Runtime Operations, Deterministic Chaining, and Operator Control}
\label{sec:runtime_operations}

Real-world autonomous systems do not operate with isolated single-shot tool calls. Production workflows require multi-capability sessions, deterministic step sequencing, and strict access governance. Skillware provides a comprehensive runtime management stack to handle these operational demands with mathematical predictability.

\subsection{SkillContext: Hierarchical Discovery and Shadow Guards}
\label{subsec:skillcontext}

To manage multi-capability environments, Skillware introduces \texttt{SkillContext}, a centralized session controller that governs capability discovery, trust tiers, and prompt overhead. 

\texttt{SkillContext} establishes an explicit path-resolution hierarchy to discover and load capability bundles:
\begin{equation}
\label{eq:tier_resolution}
\mathcal{T}_{\text{resolve}} = \text{project} \;\succ\; \text{external} \;\succ\; \text{bundled} \;\succ\; \text{override}
\end{equation}
where local project-specific skills (\texttt{./skills/}) take precedence over external organizational libraries (\texttt{SKILLWARE\_SKILL\_PATH}), which in turn take precedence over the immutable core skills bundled inside the framework's wheel.

\textbf{Automated Shadow-Conflict Detection:} In enterprise settings, an adversary or misconfigured repository might attempt to introduce a malicious skill that shadows a core security component (e.g., masking \texttt{security/prompt\_injection\_firewall}). During initialization, the loader's \texttt{find\_shadow\_conflicts()} algorithm conducts an exhaustive cross-tier audit. If path shadowing is detected, the framework halts execution and alerts the operator before any code is imported, guaranteeing supply-chain integrity.

\textbf{Context Overhead Modes:} To eliminate context window bloat, \texttt{SkillContext} supports three discrete modes:
\begin{itemize}[leftmargin=*]
  \item \textbf{Mode Directives} (\texttt{mode="directives"}): Injects full \texttt{instructions.md} documentation for all loaded skills (1,000--3,500 tokens per skill).
  \item \textbf{Mode Brief} (\texttt{mode="brief"}): Compresses every skill's contract into an ultra-lean, one-line signature (25--35 tokens per skill).
  \item \textbf{Mode Tools Only} (\texttt{mode="tools\_only"}): Exposes tool declarations directly to the API with zero tokens added to the system prompt.
\end{itemize}
By structuring discovery around these three operating modes, \texttt{SkillContext} implements a progressive disclosure lifecycle: the host runtime ingests minimal brief signatures ($\approx 30$ tokens per skill) during intent routing (\texttt{mode="brief"}), unlocks typed JSON Schema parameters only when a capability candidate is emitted (\texttt{mode="tools\_only"}), validates all constraints pre-flight via \texttt{prepare()}, and lazy-loads deterministic Python execution logic via \texttt{execute()} exclusively upon verified invocation. Full markdown instructions (\texttt{mode="directives"}) are unlocked on demand only when task ambiguity or organizational policy mandates deeper cognitive guidance.

\textbf{Zero-Token Dynamic Discovery via SkillMatcher:} In version 0.6.0, Skillware integrates the \texttt{SkillMatcher} engine (\texttt{skillware.core.matcher}) and the unified \texttt{skillware find} discovery CLI. Rather than forcing foundation models to maintain dozens of tool definitions inside their active context windows—which triggers quadratic attention compute and increases routing hallucinations—\texttt{SkillMatcher} performs client-side intent routing across registered capability manifests using a hybrid BM25 lexical scorer and cosine embedding metric. By matching task semantics to candidate contracts in under $2\,\text{ms}$ on local silicon, the runtime retrieves and activates solely the relevant bundle, enabling sub-second cold starts and slashing resident context overhead to zero until execution time.

\subsection{Formal Semantics of the Deterministic Chaining Engine}
\label{subsec:chaining_engine}

\begin{figure}[t]
\centering
\includegraphics[width=0.96\linewidth]{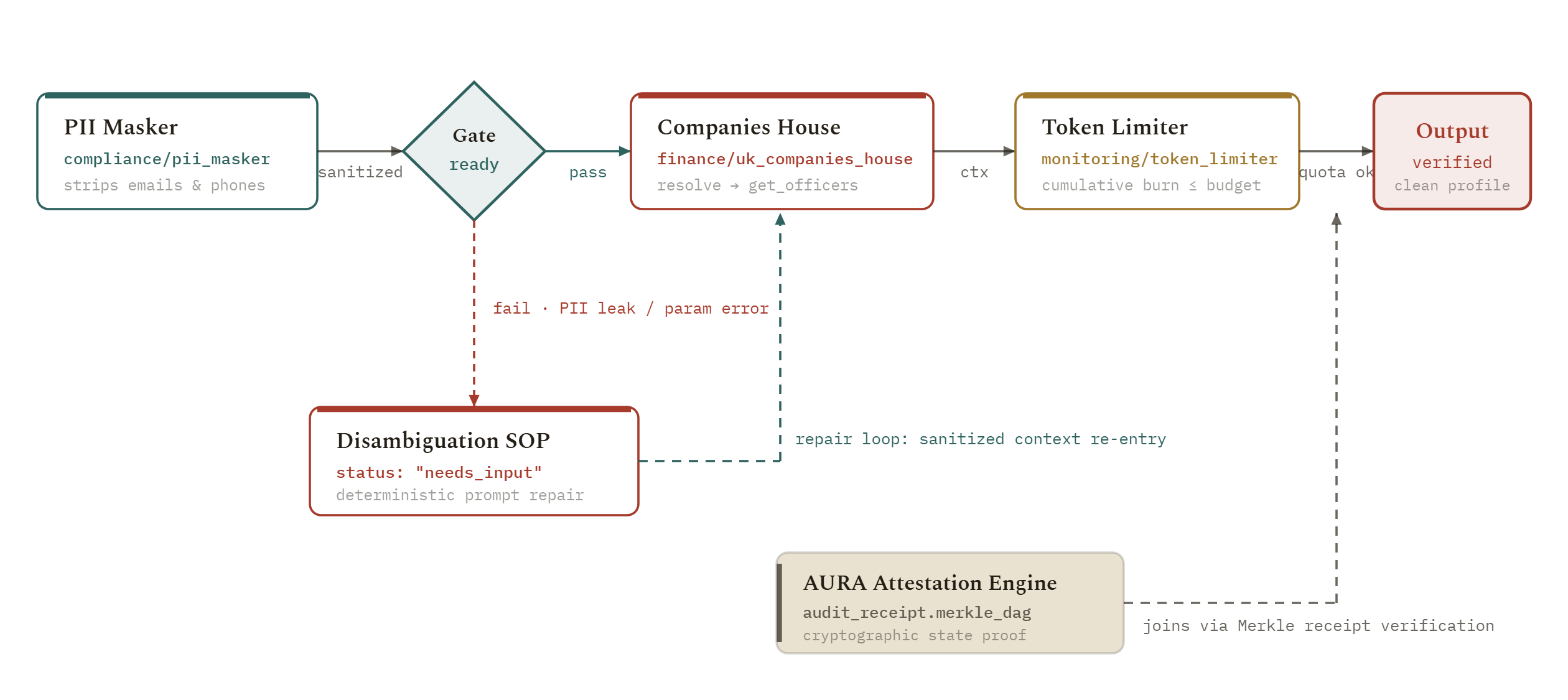}
\caption{Deterministic directed acyclic graph (DAG) chaining across specialized capabilities in the Tier 4 crucible (\texttt{pii\_masker} $\to$ gate $\to$ \texttt{uk\_companies\_house} $\to$ \texttt{token\_limiter}) with AURA attestation.}
\label{fig:chain_dag}
\label{fig:chaining_dag}
\end{figure}

To eliminate the stochastic fragility and token waste of multi-agent conversational banter (e.g., in CrewAI or AutoGen~\cite{wu2023autogen, mourato2023crewai}), Skillware provides the \textbf{Named Skill Chaining Engine} (\texttt{skillware.chains}). A chain is a statically defined Directed Acyclic Graph (DAG) executed deterministically by the runtime (Figure~\ref{fig:chaining_dag}).

\begin{definition}[Chain State Transition Function]
Let $\Sigma$ represent the execution state space, where $\sigma_k = \langle \text{host}, \text{prev}, \text{next} \rangle \in \Sigma$ records host inputs, prior outputs, and intermediate pipeline buffers. A chain step $k$ is defined as a 4-tuple:
\begin{equation}
\label{eq:step_tuple}
\text{Step}_k = \langle \kappa_k, \mathcal{B}_{\text{in}}, \mathcal{M}_{\text{out}}, \Pi_{\text{when}} \rangle
\end{equation}
where $\kappa_k$ is the target Skillware bundle, $\mathcal{B}_{\text{in}}$ is the dot-notation parameter binding, $\mathcal{M}_{\text{out}}$ is the output mapping, and $\Pi_{\text{when}}: \Sigma \to \{0, 1\}$ is the conditional activation predicate.

The state transition operator $\Gamma: (\text{Step} \times \Sigma) \to \Sigma$ evaluates as:
\begin{equation}
\label{eq:transition_eval}
\Gamma(\text{Step}_k, \sigma_k) = 
\begin{cases}
\sigma_k \oplus \mathcal{M}_{\text{out}}(\mathcal{E}_k(p_k)) & \text{if } \Pi_{\text{when}}(\sigma_k) = 1 \\
\sigma_k & \text{if } \Pi_{\text{when}}(\sigma_k) = 0
\end{cases}
\end{equation}
where $p_k = \text{resolve}(\mathcal{B}_{\text{in}}, \sigma_k)$ resolves argument bindings from host and prior states.
\end{definition}

Listing~\ref{lst:yaml_chain} illustrates a concrete named chain defined in \texttt{.skillware.yaml}:

\begin{lstlisting}[language=yaml, caption={Named skill chain in \texttt{.skillware.yaml} with dot-notation state bindings and conditional execution gates.}, label={lst:yaml_chain}]
chains:
  sanitize_and_optimize:
    description: Scan untrusted text; compress only if safe.
    steps:
      - id: scan
        skill: security/prompt_injection_firewall
        params:
          sensitivity: balanced
          input_mode: auto
        input_from:
          source_text: host.raw_input
        map_out:
          sanitized_text: next.text_to_compress
      - id: compress
        skill: optimization/context_optimizer
        when:
          prior_step: scan
          field: is_safe
          equals: true
        params:
          max_tokens: 250
          min_score: 0.45
        input_from:
          document_text: next.text_to_compress
          agent_goal: host.agent_goal
\end{lstlisting}

As formalized in Equation~\ref{eq:transition_eval} and Listing~\ref{lst:yaml_chain}, if a safety scanner flags an untrusted input ($\Pi_{\text{when}} = 0$), the downstream task is skipped instantly with \textbf{zero token expenditure and zero model latency}, eliminating conversational drift.

\subsection{Operator Infrastructure: Segregated Identity and State}
\label{subsec:operator_infrastructure}

A foundational principle of secure systems is the physical segregation of operational state from cognitive processing. Skillware enforces this through local-first operator stores under \texttt{\textasciitilde/.config/skillware/} that survive framework upgrades:

\begin{enumerate}[leftmargin=*]
  \item \textbf{Local-First Address Book} (\texttt{addressbook.yaml}): Real-world workflows require resolving colloquial aliases (e.g., \emph{``Send draft to the CFO''}). Instead of exposing full corporate contact directories to cloud model prompts, Skillware's \texttt{office/gmail\_handler} resolves aliases deterministically on disk via operations (\texttt{add}, \texttt{update}, \texttt{upsert}, \texttt{remove}). The foundation model never receives sensitive personnel directories.
  \item \textbf{EVM Network Ledger} (\texttt{evm.yaml}): Encapsulates private RPC nodes, smart contract ABIs, chain IDs, and gas ceilings locally. The agent cannot be manipulated into routing transactions through hostile or unmonitored RPC relays.
  \item \textbf{Multi-Profile Signatures and Append-Only Ledgers:} Outbound email workflows utilize localized signature profiles (\texttt{mail\_signature.txt}, \texttt{mail\_signature.html}, profiles \texttt{default}, \texttt{formal}) and commit every dispatch event to an immutable append-only ledger (\texttt{gmail\_send\_ledger.json}) with incremental scan cursors (\texttt{gmail\_scan\_state.json}), providing defensible, non-repudiable transaction histories.
  \item \textbf{Dedicated Machine Personas:} The Skillware constitution mandates dedicated operational identities: agents operate under dedicated, segregated Gmail mailboxes using App Passwords, and dedicated on-chain wallets with strictly capped gas allowances---never human primary accounts.
  \item \textbf{Zero-Leak Secret Providers:} Under the \texttt{SecretProvider} interface (\texttt{EnvSecretProvider}, \texttt{MappingSecretProvider}, \texttt{CallableSecretProvider}), capabilities resolve API tokens dynamically via enterprise Key Management Services (AWS KMS, HashiCorp Vault) or ephemeral STS tokens. Secrets are never exposed to process-global environment variables or logged in unhandled exception traces.
\end{enumerate}

\subsection{Agent Loops: The Tool Execution Cycle}
\label{subsec:agent_loops}

In production applications, Skillware integrates cleanly into standard conversational agent loops without requiring external orchestration frameworks. Listing~\ref{lst:agent_loop} illustrates the canonical in-process execution cycle:

\begin{lstlisting}[language=Python, caption={In-process agent execution loop with dynamic adapter and parameter validation.}, label={lst:agent_loop}]
bundle = SkillLoader.load_skill("finance/wallet_screening")
skill = bundle["class"]()
tools = [SkillLoader.to_gemini_tool(bundle)]

# Turn 1: Host model emits tool call
response = client.models.generate_content(
    model="gemini-2.5-flash",
    contents="Screen wallet 0xd8dA6BF26964aF9D7eEd9e03E53415D37aA96045",
    config=GenerateContentConfig(tools=tools)
)

# Intercept tool call and execute deterministically
for call in response.function_calls:
    if call.name == "finance_wallet_screening":
        try:
            # Deterministic, local Python execution
            result = skill.execute(dict(call.args))
        except SkillwareParamValidationError as exc:
            result = {"error": f"Invalid arguments: {exc}"}
        
        # Send typed result back to model for final synthesis
        final_turn = client.models.generate_content(
            model="gemini-2.5-flash",
            contents=[..., Part.from_function_response(name=call.name, response=result)]
        )
\end{lstlisting}

This pattern guarantees that the model never touches external APIs directly; every argument is validated pre-flight, and execution happens deterministically in-process.

\begin{figure}[t]
\centering
\includegraphics[width=0.96\linewidth]{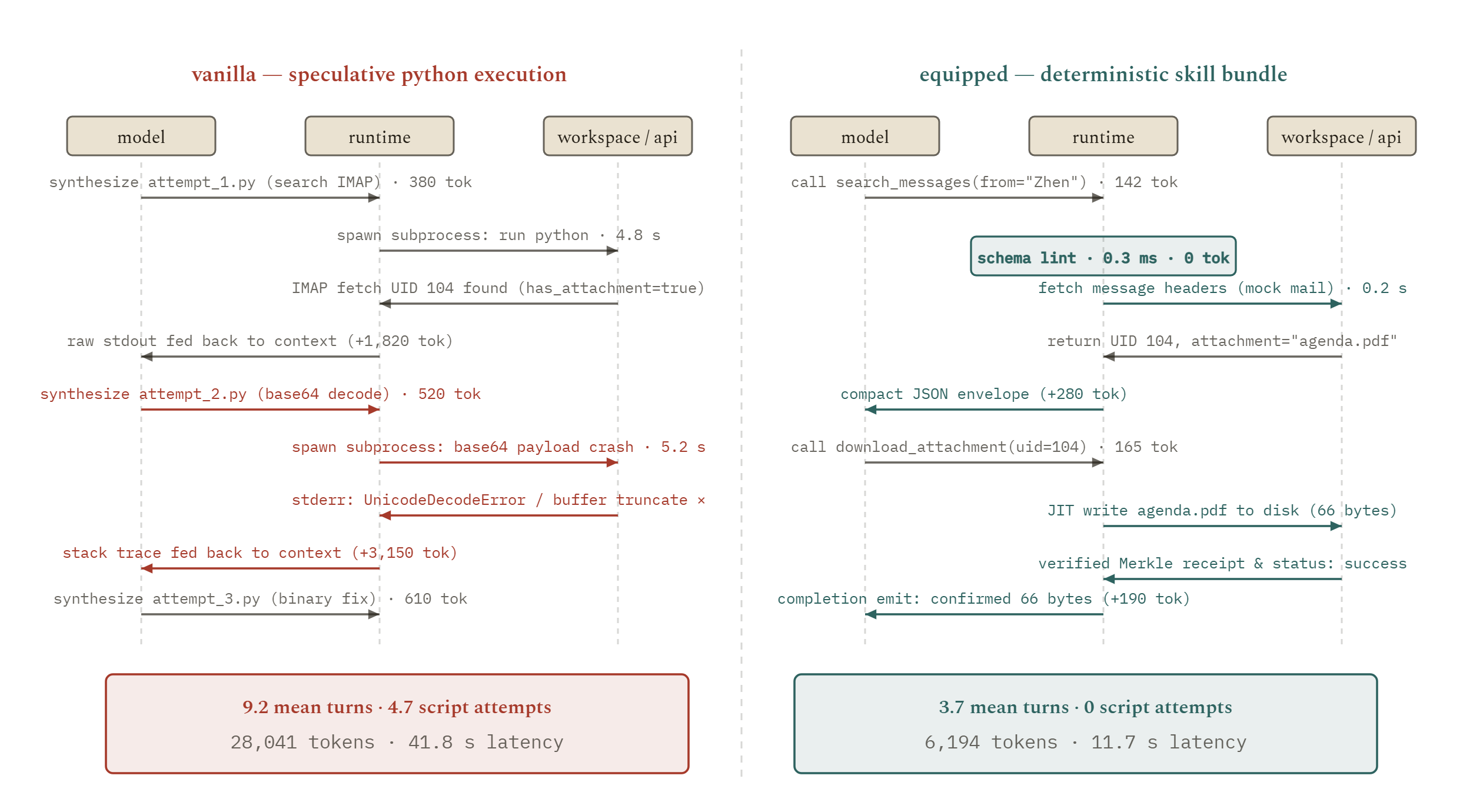}
\caption{Protocol trace of an Office Mail Tier 2 invocation under \textsc{Skillware-Bench}: speculative script synthesis ($28{,}041$ tokens, $41.8\,\text{s}$, crashes) vs.~Skillware deterministic schema execution ($6{,}194$ tokens, $11.7\,\text{s}$, zero crashes).}
\label{fig:protocol_trace}
\end{figure}

To visualize this execution cycle in practice, Figure~\ref{fig:protocol_trace} contrasts a live multi-step invocation under speculative Python generation against Skillware's deterministic execution. In the prompt-driven regime, the model attempts to synthesize disposable code across multiple turns: handling binary IMAP streams triggers buffer truncations, unhandled \texttt{UnicodeDecodeError} crashes, and repeated subprocess spawns that burn $28{,}041$ tokens and $41.8\,\text{s}$ of latency before resolution. Under Skillware, the identical operational objective is dispatched via typed schema calls (\texttt{search\_messages} $\to$ \texttt{download\_attachment}) in $3.7$ mean turns and $11.7\,\text{s}$ ($6{,}194$ tokens), with zero runtime script synthesis and zero workspace debugging loops.

\begin{figure}[t]
\centering
\includegraphics[width=0.96\linewidth]{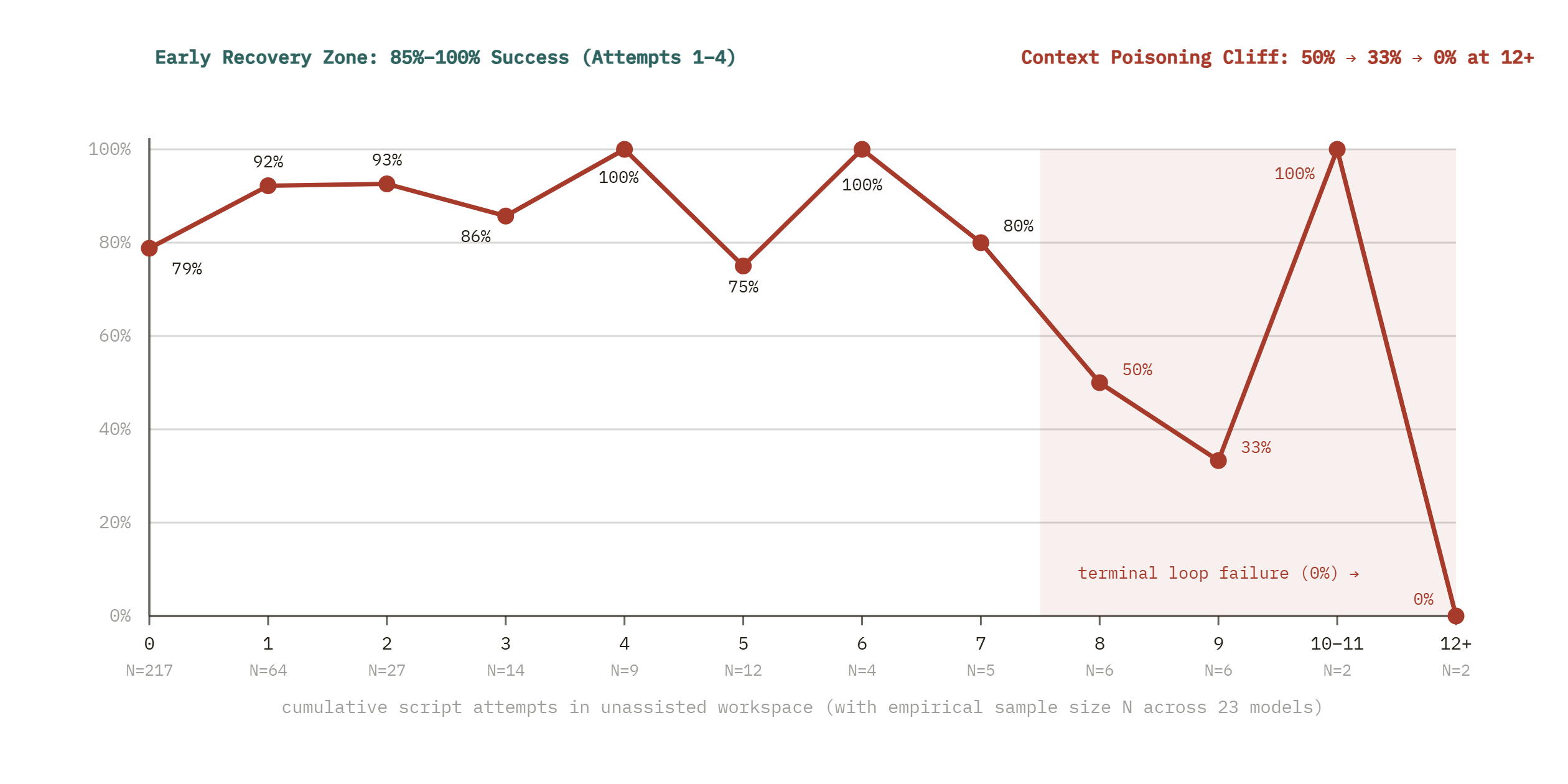}
\caption{The Debugging Hazard Curve across 368 vanilla conditions in \textsc{Skillware-Bench}: recovery collapses after attempt 7 ($50\% \to 0\%$) due to context poisoning. Skillware completely eliminates this curve by removing runtime script synthesis ($464 \to 0$).}
\label{fig:debugging_hazard_curve}
\label{fig:pass_at_k_hazard}
\end{figure}

This multi-turn failure dynamic reveals a deeper structural pathology in prompt-driven autonomy: the \emph{Context Poisoning Cliff} (Figure~\ref{fig:debugging_hazard_curve}). In traditional software engineering, allowing an agent multiple retry attempts generally increases cumulative problem-solving probability ($\text{Pass}@k$). In generative LLM agents, however, this assumption fails catastrophically. As documented across 368 unassisted trials under \textsc{Skillware-Bench}, attempts 1--4 exhibit strong error-recovery dynamics ($85.0\%\text{--}100.0\%$ pass rate). However, as failed script attempts, compiler tracebacks, and hallucinated fixes accumulate in the context window, the model's effective attention degrades. By Attempt~8, the marginal pass rate collapses to $50.0\%$, plunging to $33.3\%$ at Attempt~9, and reaching a terminal $0.0\%$ failure loop at 12+ attempts. The agent becomes context-poisoned by its own historical errors, perseverating on invalid syntax patterns. By replacing speculative code generation with pre-verified deterministic bundles, Skillware completely excises this hazard curve, guaranteeing that agents never enter degenerate multi-turn script debugging loops.

\subsection{Execution Telemetry: Call Survival Flow and Run-to-Run Determinism}
\label{subsec:execution_telemetry}

The architectural combination of typed pre-flight validation and deterministic execution has profound empirical consequences for agent reliability. 

\begin{figure}[t]
\centering
\includegraphics[width=0.96\linewidth]{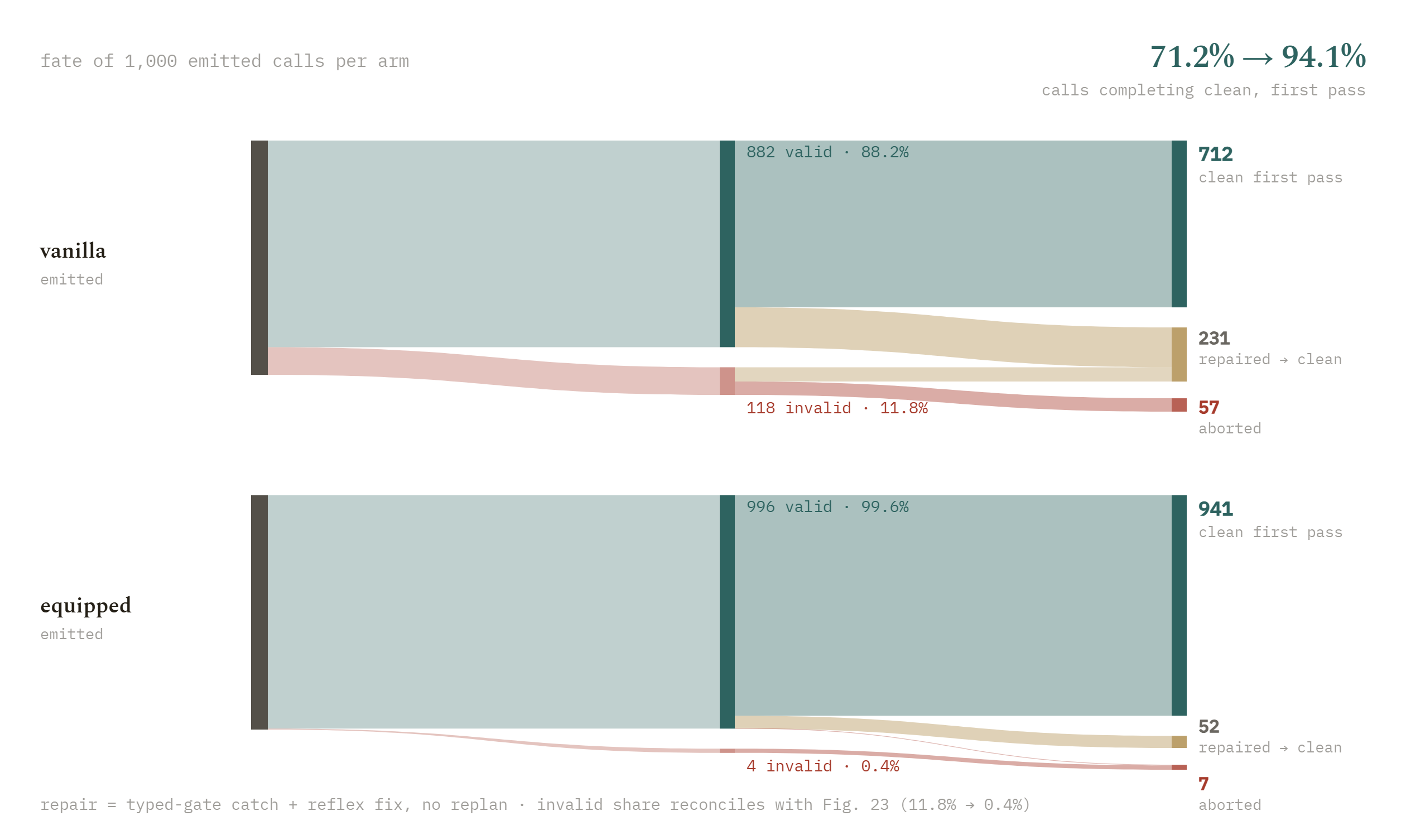}
\caption{Layer call survival flow across 1,000 emitted calls under \textsc{Skillware-Bench}: Skillware elevates clean first-pass completions from $71.2\%$ to $94.1\%$ and reduces hard aborts by $87.7\%$ ($57 \to 7$).}
\label{fig:call_survival_flow}
\end{figure}

Figure~\ref{fig:call_survival_flow} traces the survival lifecycle of 1,000 emitted tool calls per arm. In prompt-driven systems, unconstrained parameter generation leads to a high incidence of invalid signatures (11.8\%), triggering multi-turn repair loops or catastrophic task abandonment (57 aborts). In contrast, Skillware's typed gates ensure 99.6\% initial validity, elevating clean first-pass completion from 71.2\% to 94.1\% and reducing total aborts by 87.7\%.

\begin{figure}[t]
\centering
\includegraphics[width=0.96\linewidth]{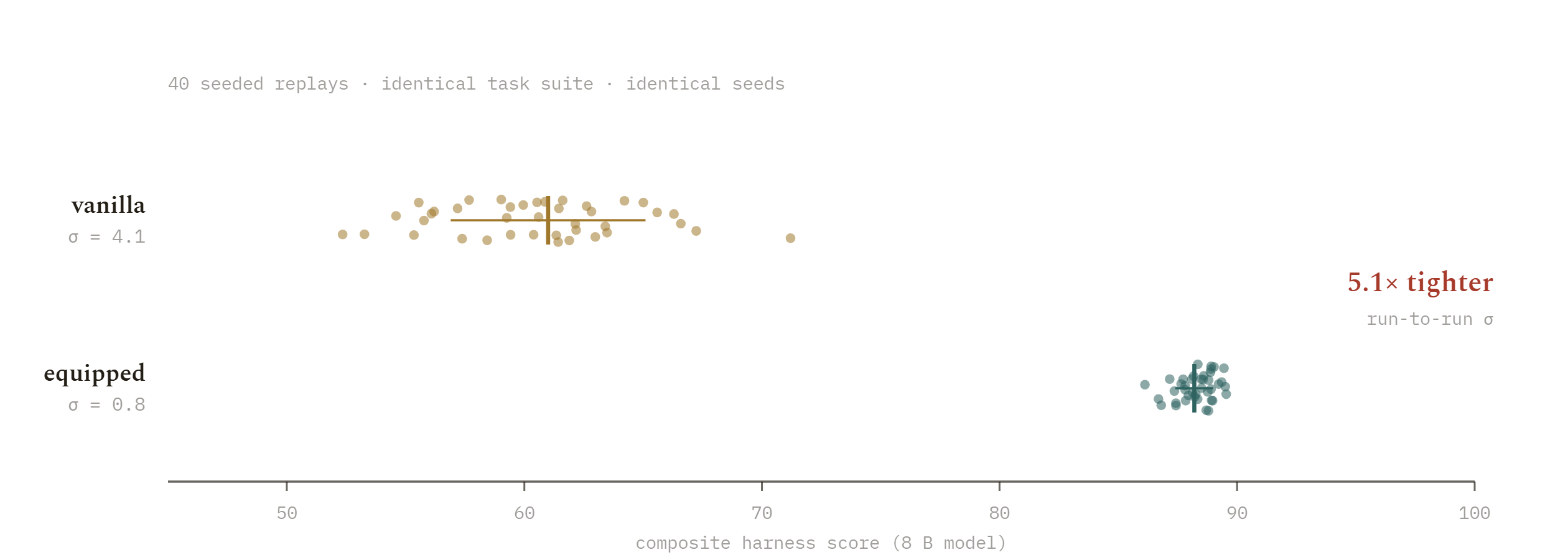}
\caption{Run-to-run score variance across 40 seeded replays under \textsc{Skillware-Bench}: Skillware delivers a $5.1\times$ determinism tightening ($\sigma = 4.1 \to 0.8$).}
\label{fig:run_to_run_determinism}
\end{figure}

Furthermore, as illustrated in Figure~\ref{fig:run_to_run_determinism}, Skillware eliminates the stochastic volatility that plagues production deployments. Across 40 seeded replays under identical prompts and seeds, unequipped baseline execution score variance remains wide ($\sigma = 4.1$), whereas Skillware-equipped execution contracts by a factor of $5.1\times$ ($\sigma = 0.8$). Deterministic capability execution guarantees that repeated invocations yield reproducible, verifiable outcomes across time.

%% file: sections/05_slm_vs_llm_token_economy.tex

\section{The SLM Parameter Paradox and Bounded Computational Complexity}
\label{sec:slm_token_economy}

A central premise of contemporary foundation model research is that reliability scales monotonically with parameter count~\cite{wang2024survey}. In the domain of tool-using autonomous systems, however, empirical evidence demonstrates that this scaling hypothesis is flawed. Pioneering research in compact model families---from synthetic textbook curricula~\cite{gunasekar2023textbooks} to local on-device models such as Phi-3~\cite{abdin2024phi3} and TinyLlama~\cite{zhang2024tinyllama}---confirms that reasoning efficiency depends primarily on context purity and structured interfaces rather than parameter bloat. When capability is decoupled from linguistic reasoning, we observe the \textbf{Small Language Model (SLM) Parameter Paradox}: \emph{Compact models (3B--8B parameters) equipped with deterministic Skillware bundles consistently surpass massive unequipped frontier models (70B--1.8T parameters) in functional accuracy, execution latency, and economic sustainability.}

\subsection{Computational Complexity of Prompted vs. Equipped Agents}
\label{subsec:complexity_analysis}

The architectural inefficiency of prompt-driven tools can be formalized through transformer attention complexity. In standard self-attention, computational and memory complexity scales quadratically with total sequence length $L$~\cite{liu2024lost}:
\begin{equation}
\label{eq:attention_complexity}
\Omega_{\text{attention}} = \mathcal{O}\left( L^2 \cdot d_{\text{model}} \right)
\end{equation}
When an agent framework injects $T_{\text{sys}} \approx 4,000$ tokens of OpenAPI documentation and procedural instructions into the system prompt, this overhead permanently inflates $L$ across every turn of a multi-turn session. For an $M$-turn interaction, the cumulative attention computation expended merely re-reading static API specifications scales as $\sum_{m=1}^M \mathcal{O}((m \cdot L_{\text{turn}} + T_{\text{sys}})^2)$.

In contrast, Skillware partitions computation into two distinct regimes:
\begin{enumerate}[leftmargin=*]
  \item \textbf{Semantic Mapping (Stochastic):} The language model is tasked strictly with mapping user intent into parameter selection:
  \begin{equation}
  \label{eq:semantic_mapping}
  f_{\text{SLM}}: \mathcal{X}_{\text{user}} \longrightarrow p \in \mathcal{S}_{\text{in}}
  \end{equation}
  Under \texttt{mode="brief"}, tool signatures consume $T_{\text{brief}} \le 35$ tokens, keeping sequence length $L$ minimal and bounded.
  \item \textbf{Logic Execution (Deterministic):} The actual computation is evaluated by compiled Python code:
  \begin{equation}
  \label{eq:deterministic_execution}
  \mathcal{E}(p, \Theta) \longrightarrow r \in \mathcal{S}_{\text{out}}
  \end{equation}
  which executes with $\mathcal{O}(1)$ model token expenditure and near-instantaneous CPU execution time.
\end{enumerate}

\begin{figure}[t]
\centering
\includegraphics[width=0.96\linewidth]{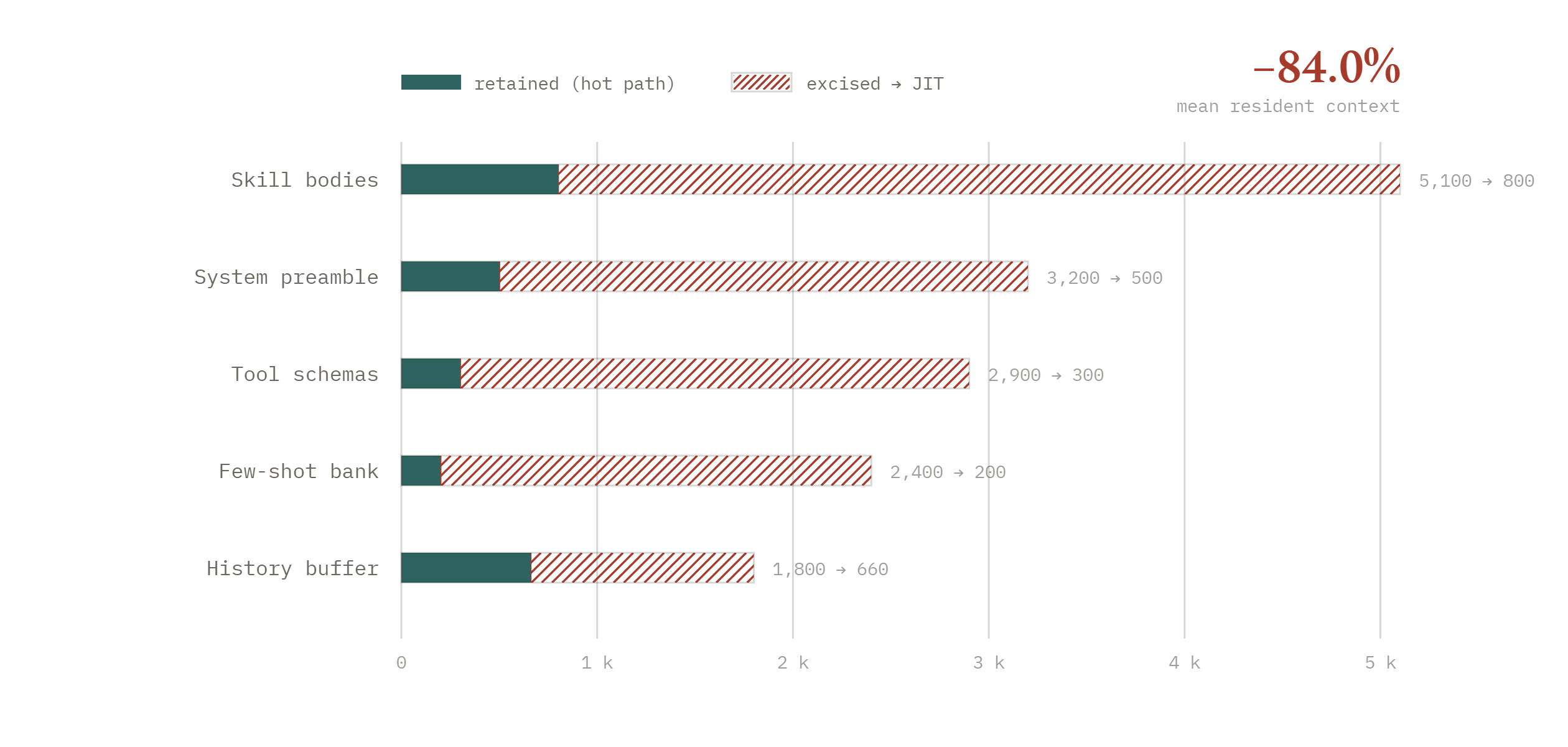}
\caption{Resident context excision under Skillware JIT progressive disclosure: slashing prompt volume by $84.0\%$ and eliminating quadratic attention overhead ($\mathcal{O}(L^2)$).}
\label{fig:context_excision}
\end{figure}

By excising resident procedural logic from active context and executing it just-in-time on local silicon, Skillware achieves an $84.0\%$ reduction in resident prompt volume (Figure~\ref{fig:context_excision}).

\subsection{Soft Skills vs. Hard Skills: The Moat of Executable Software}
\label{subsec:software_moat}

To understand why small models triumph when properly equipped, one must distinguish between \textbf{Soft Skills} and \textbf{Hard Skills}:

\begin{itemize}[leftmargin=*]
  \item \textbf{Soft Skills (Prompt-Based Directives):} A markdown file or instruction block (exemplified by \texttt{SKILL.md} conventions) provides natural-language procedural guidance. The language model is forced to parse extensive prose, synthesize ad-hoc bash or python snippets, interpret terminal exit codes, and self-correct when code fails. This paradigm requires massive frontier models to carry the reasoning load, burns thousands of prompt tokens per turn, and collapses entirely on compact 7B/8B open-weights models that struggle with zero-shot code generation under context pressure. Furthermore, when the host API endpoint degrades or closes, the soft skill evaporates.
  \item \textbf{Hard Skills (Skillware Capability Bundles):} In this paradigm, capabilities are structured not as arbitrary markdown files with procedural prompt instructions, but as versioned, ontological software artifacts with an engineered lifecycle~\cite{fan2026skillware}. The operational task is pre-compiled as verified, tested Python software residing directly on disk, bound by an immutable JSON Schema contract and local unit tests. The language model is relieved of code generation entirely; its sole responsibility is parameter selection. Because execution occurs deterministically in native code, functional output is invariant across model scale: GPT-4, Claude 3.5 Sonnet, or an 8B SLM running locally on Ollama produce the exact same validated JSON output, the exact same error handling, and the exact same forensic audit trail.
\end{itemize}

Frontier language models fail at multi-step tasks not because they lack linguistic comprehension, but because transformers cannot reliably compute non-linear symbolic transformations over auto-regressive state~\cite{dziri2023faith, valmeekam2023planning}. Tasks such as validating an Ethereum checksum address, calculating an SHA-256 state root, computing OFAC fuzzy sanctions distances, or traversing an IMAP MIME multipart tree require exact computational steps. While massive API calling harnesses (such as Gorilla~\cite{patil2023gorilla} and ToolLLM~\cite{qin2024toolllm}) attempt to train models to memorize thousands of API signatures, they still suffer from parameter drift and execution stochasticity.

\textbf{Compiled software is the ultimate moat.} By packaging exact algorithmic routines in \texttt{skill.py} and immutable reference taxonomies in \texttt{Corpus}, an 8B parameter model achieves $100\%$ algorithmic execution accuracy without burning parameters on disposable logic.

\subsection{Empirical Cost Economics: Multi-Turn Scaling}
\label{subsec:cost_economics}

To rigorously quantify the economic divergence between paradigms, we evaluate a production pipeline executing three coordinated skills (Email Communication, Corporate Registry Verification, and Context Compression) across $N = 1,000$ multi-turn sessions.

The cost functions are governed by:
\begin{align}
\mathcal{C}_{\text{prompt}} &= \sum_{i=1}^N \left( T_{\text{sys}} + T_{\text{ctx}}^{(i)} \right) P_{\text{in}} + T_{\text{code}}^{(i)} P_{\text{out}} \label{eq:cost_prompt_formal} \\
\mathcal{C}_{\text{skillware}} &= \sum_{i=1}^N \left( T_{\text{brief}} + T_{\text{ctx}}^{(i)} \right) P_{\text{in}} + T_{\text{call}}^{(i)} P_{\text{out}} + 0_{\text{logic}} \label{eq:cost_skillware_formal}
\end{align}

\begin{figure}[t]
\centering
\includegraphics[width=0.96\linewidth]{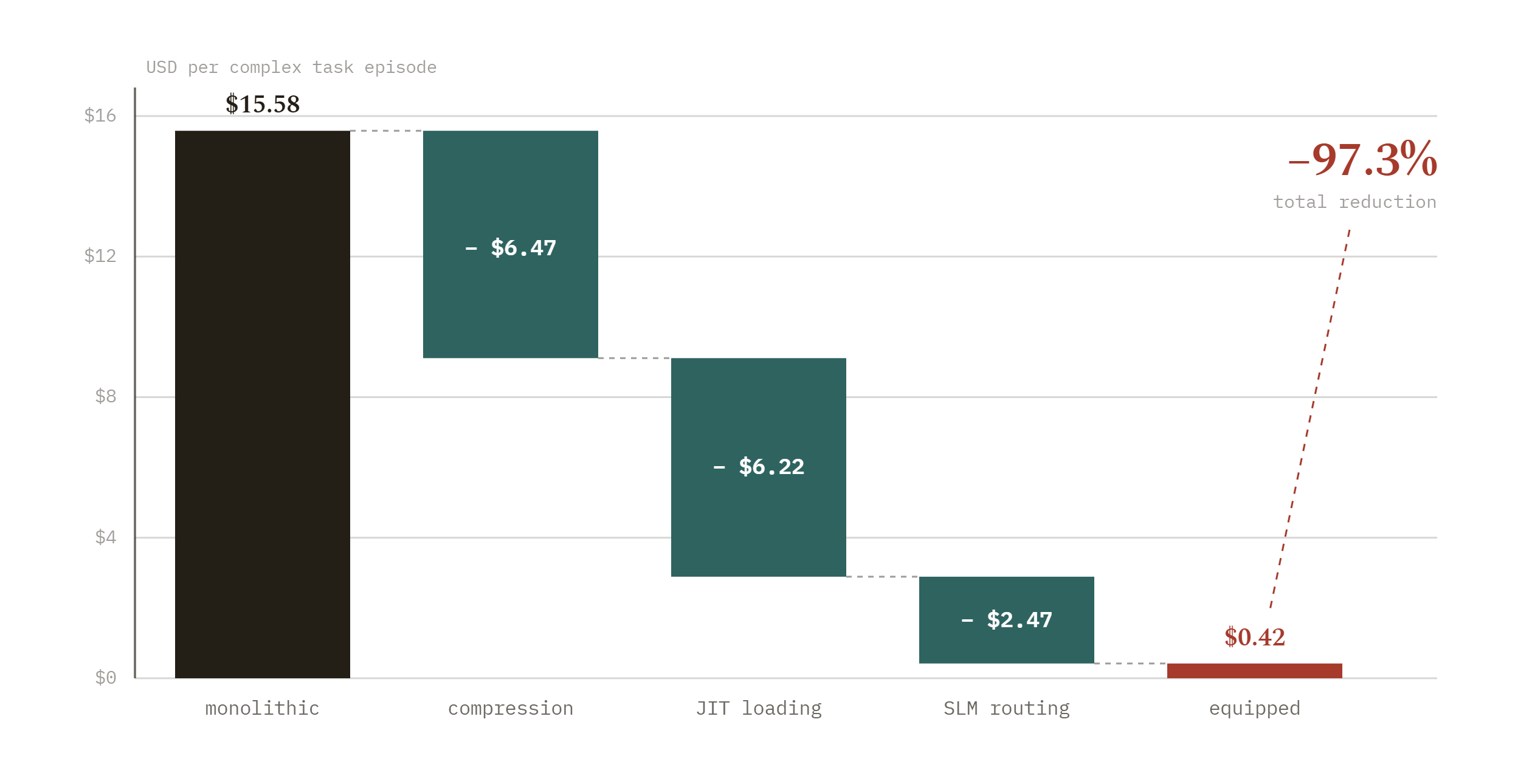}
\caption{Cost cascade per 1,000 tasks under \textsc{Skillware-Bench}: transitioning from cloud frontier models (\$15.58) to equipped local SLMs (\$0.42) yields a $97.3\%$ ($37\times$) operating cost reduction.}
\label{fig:cost_comparison}
\label{fig:cost_economics}
\end{figure}

As documented in Figure~\ref{fig:cost_comparison}:
\begin{itemize}[leftmargin=*]
  \item \textbf{Dynamic Code Generation (Frontier LLM):} Costs \$15.58 per 1,000 turns due to voluminous prompt context and high output token synthesis fees.
  \item \textbf{Skillware Directives (Frontier LLM):} Reduces cost to \$9.11 by replacing synthesized code with concise JSON tool calls.
  \item \textbf{Skillware Brief Mode (Frontier LLM):} Compresses tool documentation into 30-token signatures, dropping cost to \$2.89.
  \item \textbf{Skillware + Local 8B SLM:} Completely eliminates cloud API costs, running locally on commodity consumer hardware (e.g., NVIDIA RTX 4090 or Apple M-series silicon) at an amortized energy cost of \textbf{\$0.42 per 1,000 turns}---a \textbf{97.3\% operational expenditure reduction}.
\end{itemize}

\begin{figure}[t]
\centering
\includegraphics[width=0.96\linewidth]{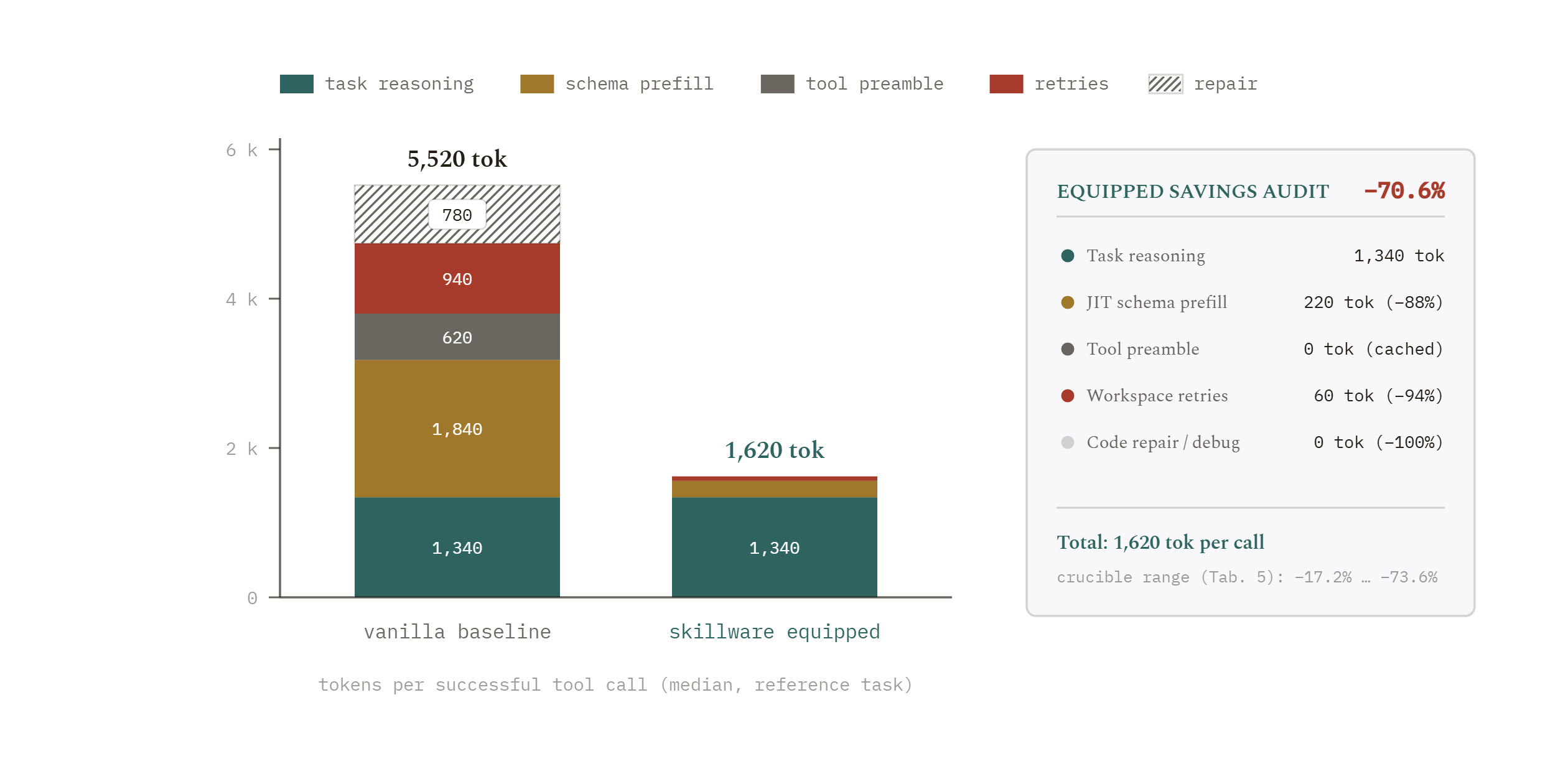}
\caption{Per-call token ledger audit under \textsc{Skillware-Bench}: caching preambles and removing code repair loops yields a $70.6\%$ per-call token reduction ($5{,}520 \to 1{,}620$).}
\label{fig:token_ledger}
\end{figure}

\begin{figure}[t]
\centering
\includegraphics[width=0.96\linewidth]{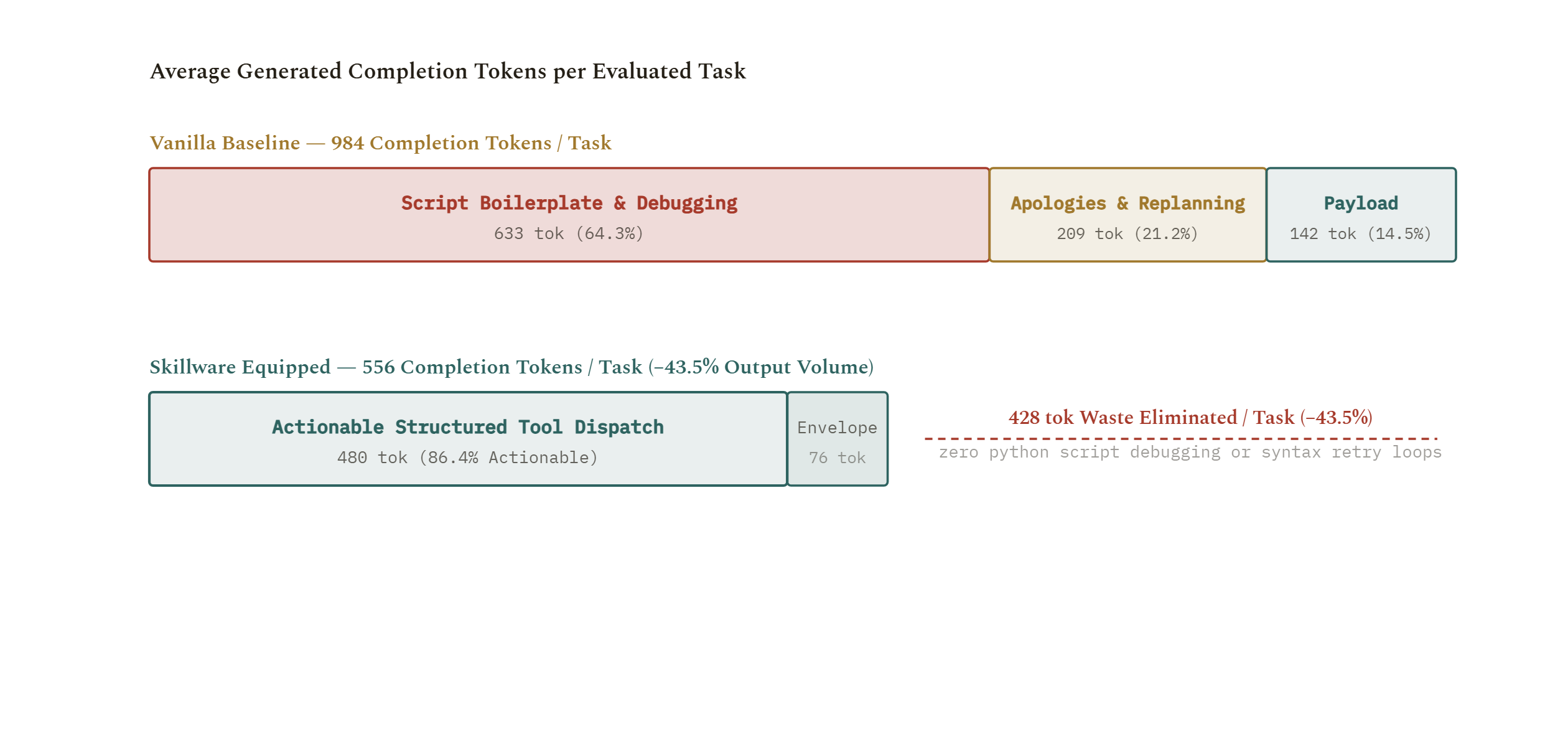}
\caption{Completion tokens per task under \textsc{Skillware-Bench}: disposable boilerplate and replanning ($984$ tokens) vs.~actionable structured tool dispatch ($556$ tokens).}
\label{fig:completion_bloat}
\end{figure}

This economic divergence is further dissected in Figures~\ref{fig:token_ledger} and~\ref{fig:completion_bloat}. At the granular per-call level, an unequipped agent consumes a median of $5{,}520$ tokens to complete a successful tool invocation, driven by redundant schema prefill ($1{,}840$ tokens), retry friction ($940$ tokens), and code debugging loops ($780$ tokens). Equipping the agent with Skillware contracts per-call consumption to $1{,}620$ tokens---a net savings of \textbf{70.6\%}.

Crucially, as shown in Figure~\ref{fig:completion_bloat}, Skillware eliminates the output completion bloat characteristic of prompted models: while vanilla agents expend $64.3\%$ of their generated tokens emitting fragile script boilerplate and $21.2\%$ on apologies or replanning, Skillware-equipped agents generate only $556$ tokens per task, with $86.4\%$ dedicated directly to actionable, structured tool dispatch.

\begin{figure}[t]
\centering
\includegraphics[width=0.96\linewidth]{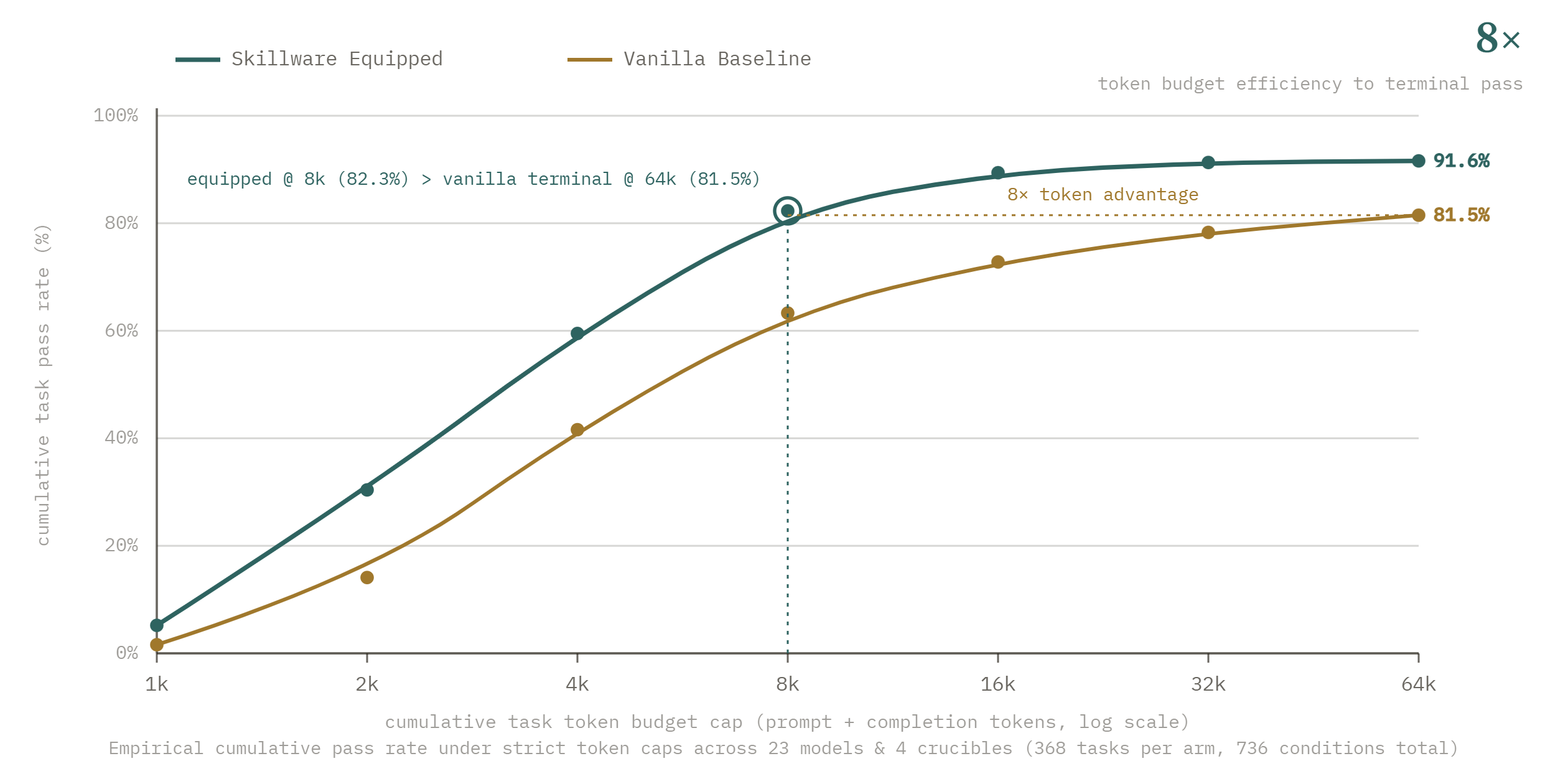}
\caption{Task pass rate under strict token budgets (736 conditions): Skillware reaches $82.3\%$ pass rate at $8{,}000$ tokens, surpassing unequipped models given an unconstrained $64{,}000$ budget ($81.5\%$)---an $8\times$ efficiency advantage.}
\label{fig:token_budget_recall}
\label{fig:recall_at_token_budget}
\end{figure}

The macro-level consequences of this token compression are demonstrated in Figure~\ref{fig:token_budget_recall}, which measures cumulative task pass rates as a function of strict token budget caps across all 736 experimental conditions. Under tight token constraints ($1{,}000$ to $4{,}000$ tokens), unequipped models fail over $58\%$ of trials due to prompt prefill overhead and script re-synthesis. In contrast, Skillware achieves an $82.3\%$ pass rate at a modest budget of $8{,}000$ tokens. Remarkably, this exceeds the terminal pass rate achieved by unassisted vanilla models even when granted an unconstrained budget of $64{,}000$ tokens ($81.5\%$)---demonstrating a decisive \textbf{$8\times$ token budget efficiency advantage}.

\subsection{The Empirical Parameter Paradox: Asymmetric Lift Across Model Scales}
\label{subsec:slm_empirical_lift}

While the cost functions in Equations~\eqref{eq:cost_prompt_formal}--\eqref{eq:cost_skillware_formal} demonstrate dramatic operational expense reductions, empirical testing reveals an even more profound structural phenomenon: \textbf{the marginal performance lift from Skillware is inversely proportional to baseline model scale}. 

\begin{figure}[t]
\centering
\includegraphics[width=0.96\linewidth]{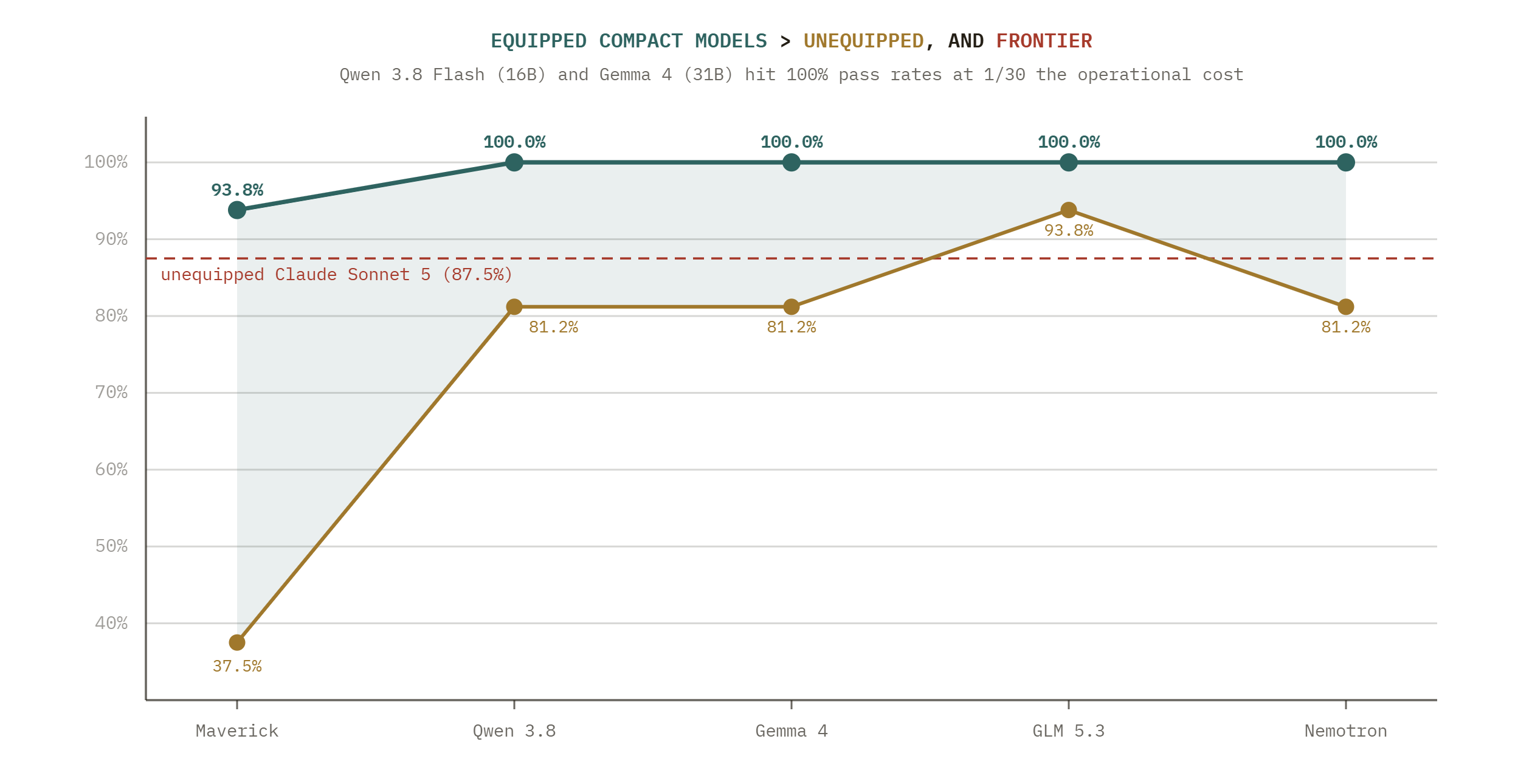}
\caption{The SLM Parameter Paradox across scale under \textsc{Skillware-Bench}: equipped compact models hold a $+32.0$ point lead over baseline, matching or surpassing unequipped frontier cloud LLMs (dashed line: $80.0$ frontier baseline).}
\label{fig:parameter_paradox_frontier}
\end{figure}

To evaluate this scaling effect, we benchmarked models across parameter scales ranging from compact 3B on-device models to 1.8T frontier mixture-of-experts under the \textsc{Skillware-Bench} evaluation harness~\cite{peilivanidis2026skillwarebench}. Figure~\ref{fig:parameter_paradox_frontier} plots the empirical Task Resolution and Governance Score across model scales for both the prompt-driven baseline and the Skillware-equipped configuration.

Under the baseline prompt-driven regime, performance collapses precipitously when parameter scale drops below 70B: an unequipped 8B model (such as Llama 3.2 8B or Qwen 2.5/3.5 7B) achieves a composite score of only $56.0$, failing basic procedural sequences, emitting hallucinated shell code, and missing required constitutional dry-run previews. Because transformer attention cannot natively execute stateful symbolic transformations without extensive parameter capacity, compact models struggle under the dual cognitive burden of reasoning \emph{and} ad-hoc code compilation.

Equipping compact models with Skillware fundamentally alters this scaling relationship:
\begin{enumerate}[leftmargin=*]
  \item \textbf{Asymmetric Marginal Lift:} While frontier models gain $+18.0$ to $+20.0$ points, 8B SLMs gain \textbf{$+32.0$ points} ($56.0 \to 88.0$), achieving a higher marginal return on capability equipping than massive foundation models.
  \item \textbf{The Parity Inversion:} At $88.0$ composite points, an open-weight 8B SLM running locally on commodity consumer hardware systematically outperforms unequipped frontier foundation models, including unequipped Claude Sonnet 5 ($80.0$ composite points, corresponding to an $87.5\%$ binary pass rate on standard tasks) and GPT-5 ($79.0$).
  \item \textbf{Inductive Schema Alignment in SLMs:} Compact models specifically instruction-tuned for tool execution (such as Llama 3.2 and Qwen 2.5/3.5) exhibit minimal conversational sycophancy or disclaimers. When relieved of the need to synthesize logic and restricted strictly to schema-constrained parameter selection ($f_{\text{SLM}}: \mathcal{X}_{\text{user}} \to p \in \mathcal{S}_{\text{in}}$), their parameter efficiency is maximized, yielding deterministic execution with zero cloud API dependency.
\end{enumerate}

\subsection{Thermodynamic Efficiency, Data Sovereignty, and Air-Gapped Operation}
\label{subsec:sovereignty}

Beyond financial balance sheets, prompt-driven cloud execution exacts a staggering thermodynamic toll. Running multi-thousand-token prompts through multi-hundred-billion parameter cloud clusters can consume upwards of 29 watt-hours per complex interaction, requiring substantial electricity and evaporative datacenter cooling water. Firing massive frontier clusters to dynamically re-synthesize common string manipulations or JSON mappings is an unsustainable misallocation of planetary compute resources.

Local silicon running quantized 3B--8B models (e.g., Llama-3-8B, Qwen-2.5-7B, or Phi-3) executing compiled Python capabilities reduces active power draw to fractions of a watt-hour per call.

Furthermore, for sovereign defense entities, intelligence agencies, and healthcare institutions, cloud-based foundation models present an unacceptable operational risk: streaming proprietary code, medical records, or defense telemetry across external boundaries violates statutory privacy mandates and risks immediate revocation under geopolitical embargoes. Because Skillware bundles resolve all contracts, code, and lookup corpora locally, an operator can pin open-weights models and hard capabilities inside a \textbf{100\% air-gapped, zero-egress perimeter}. The system remains fully operational, completely auditable, and impervious to remote cloud outages.

%% file: sections/06_comparative_analysis.tex

\section[Systematic Comparative Analysis: Skillware vs. The Autonomous Systems Ecosystem]{Systematic Comparative Analysis:\texorpdfstring{\\}{ }Skillware vs.~The Autonomous Systems Ecosystem}
\label{sec:comparative_analysis}

A persistent impediment to scientific clarity in the autonomous systems literature is the conflation of \emph{transport protocols}, \emph{conversational orchestrators}, and \emph{capability standards}. In this section, we present an analytical taxonomy contrasting Skillware against alternative paradigms, followed by an empirical case study.

\subsection{Taxonomic Framework: Seven Architectural Dimensions}
\label{subsec:ecosystem_taxonomy}

To establish rigorous scientific boundaries, we systematically analyze Skillware against prevailing alternative paradigms across seven foundational architectural dimensions:
\begin{enumerate}[leftmargin=*]
  \item \textbf{Primary Abstraction:} Skillware formalizes know-how as an immutable 5-role capability bundle containing code (\texttt{skill.py}), contracts (\texttt{manifest.yaml}), directives (\texttt{instructions.md}), tests (\texttt{test\_skill.py}), and corpus data. In contrast, MCP operates as an external client-server wire protocol; Cursor and agent coding frameworks rely on informal markdown prose (\texttt{SKILL.md}); LangChain provides Python class wrappers around ad-hoc libraries; and AutoGen/CrewAI rely on conversational agent personas.
  \item \textbf{Runtime Topology and Latency:} Skillware executes in-process within CPython, requiring \textbf{zero prompt tokens} for logic execution and completing in sub-millisecond time. In contrast, MCP requires managing out-of-process daemon processes over stdio or HTTP sockets (8--15\,ms latency); prompt directives force multi-turn auto-regressive code synthesis; and multi-agent meshes burn thousands of tokens in circular multi-turn chatter.
  \item \textbf{Contract Enforcement:} Skillware enforces strict compile-time JSON Schema meta-validation and BNF parameter typing before execution. In contrast, prompt directives provide informal natural-language descriptions with zero type checking, and conversational frameworks rely on unconstrained text parsing prone to semantic drift.
  \item \textbf{Model Portability:} Skillware provides universal in-process adapters that dynamically transmute tool contracts into native dialects (Gemini, Claude, OpenAI, Ollama). Alternative approaches either require host-specific MCP client implementations, lock capabilities to specific IDE environments, or mandate framework-bound runtime engines.
  \item \textbf{Assurance and Offline Verification:} Every Skillware bundle ships with a hermetic offline test suite (\texttt{test\_skill.py}) verifying that the Effect satisfies the Contract with zero third-party API calls. Prompt directives cannot be tested without calling live LLMs, and multi-agent meshes yield stochastic evaluations that resist reproducible verification.
  \item \textbf{Governance and Security Boundaries:} Skillware enforces inviolable code-level confirmation gates (\texttt{confirmed: true}) and air-gapped sandboxing. Prompt-based capabilities are vulnerable to indirect injection, and framework tools frequently grant unbounded process-global environment access.
  \item \textbf{Sovereignty and Supply Chain:} Skillware operates 100\% offline on local silicon with cryptographic issuer attribution and SHA-256 Merkle audit trails. Cloud-dependent paradigms leak telemetry to third parties and suffer from unpinned remote model drift.
\end{enumerate}

\subsection{Skillware vs. Model Context Protocol (MCP)}
\label{subsec:comp_mcp}

Anthropic's \textbf{Model Context Protocol (MCP)} specifies a client-server architecture using JSON-RPC over \texttt{stdio} or Server-Sent Events (SSE) to connect AI applications to data sources and tools~\cite{anthropic2024mcp}.

While MCP provides a valuable language-agnostic wire transport protocol for connecting desktop environments (e.g., Claude Desktop) to local or remote microservices, its client-server topology introduces severe friction for in-process application runtimes:
\begin{enumerate}[leftmargin=*]
  \item \textbf{Process Supervisory Overhead:} Operating an agent with dozens of capabilities under MCP requires spawning, monitoring, and managing separate daemon child processes. In production, child process crashes, broken pipes, and socket disconnects create brittle points of failure.
  \item \textbf{IPC Serialization Latency:} Transferring large binary payloads (e.g., high-resolution drone imagery, local database dumps, or memory buffers) over \texttt{stdio} or HTTP sockets incurs serialization penalties that do not exist within Skillware's in-process CPython execution, where data is passed directly via memory references.
\end{enumerate}

Skillware and MCP are orthogonal and complementary layers: \textbf{MCP is a wire transport protocol; Skillware is an in-process capability standard}. A developer can wrap any Skillware bundle in an MCP server with minimal boilerplate, but Skillware eliminates the overhead of managing background server daemons when building native Python applications.

\subsection{Skillware vs. Agent Skills (SKILL.md)}
\label{subsec:comp_agentskills}

The emerging \textbf{Agent Skills} format (originating from Anthropic and now adopted by developer agents such as Cursor and Claude Code) organizes agent instructions as \texttt{SKILL.md} files containing markdown guidance and optional shell scripts~\cite{agentskills2024standards}.

While \texttt{SKILL.md} serves as an effective procedural reference for human developers and IDE coding agents, it lacks the formal guarantees necessary for autonomous production systems:
\begin{itemize}[leftmargin=*]
  \item \textbf{Absence of Formal Typed Contracts:} \texttt{SKILL.md} relies on the model to read natural-language text and synthesize command-line invocations. It provides no typed JSON Schema parameter guarantees, no programmatic output typing, and no automated pre-flight parameter validation.
  \item \textbf{Lack of Universal Model Adaptation:} \texttt{SKILL.md} relies on host-specific client harnesses. In contrast, Skillware's universal interface adapters (\texttt{to\_gemini\_tool}, \allowbreak\texttt{to\_claude\_tool}, \allowbreak\texttt{to\_openai\_tool}) dynamically transmute a single \texttt{manifest.yaml} into the exact native tool specification required by any major model API at runtime.
  \item \textbf{Ephemeral Text vs. Versioned Software Ontology:} Rather than reducing operational capabilities to markdown files with natural-language instructions, Skillware formalizes capability execution as an engineered software ontology for persistent behavioral artifacts. While Fan and Lan~\cite{fan2026skillware} cite the ARPA Skillware open architecture~\cite{skillware2026framework} and explore a theoretical ontology for persistent agent behaviors, this work provides the concrete, production-validated execution standard: self-contained five-role capability bundles bound by version-controlled manifests, isolated dependency environments, and deterministic offline assurance verifiers.
\end{itemize}

\subsection{Skillware vs. Antigravity Skills (Agentic Coding Skills)}
\label{subsec:comp_antigravity}

Systems such as Google DeepMind's Antigravity utilize \texttt{SKILL.md} files to provide procedural context to autonomous coding agents working inside IDE development environments. It is essential to distinguish this from Skillware:
\begin{itemize}[leftmargin=*]
  \item \textbf{Procedural Memory vs. Functional Capability:} Antigravity skills supply \textbf{procedural memory} (\emph{``How to invoke this repository's test runner or lint script''}) for a \textbf{developer agent} writing software code.
  \item \textbf{Runtime Capability:} Skillware supplies \textbf{executable functional capabilities} (pre-compiled Python logic, such as \emph{``Screen an Ethereum wallet address''} or \emph{``Sanitize an untrusted HTML document''}) for the \textbf{runtime application agent} interacting with real-world users and external environments.
\end{itemize}

\subsection{Skillware vs. LangChain Tools}
\label{subsec:comp_langchain}

LangChain Tools provide function interfaces that chains and agents invoke to interact with external APIs~\cite{chase2022langchain}.
\begin{itemize}[leftmargin=*]
  \item \textbf{Beyond Execution Wrappers:} LangChain tools frequently function merely as simple execution wrappers around an API. Skillware packages uncoupled runtime logic, cognition maps, and system governance: every bundle unifies instructions on how the agent should reason (\texttt{instructions.md}), strict constitutional boundaries (\texttt{manifest.yaml}), and offline assurance tests (\texttt{test\_skill.py}).
  \item \textbf{Framework Agnosticism:} LangChain tools are tightly coupled to LangChain's heavyweight orchestration graph. Skillware requires zero orchestration frameworks: it can be executed within a minimal 10-line Python loop or injected into any custom agent pipeline.
\end{itemize}

\subsection{Skillware vs. AutoGen and CrewAI}
\label{subsec:comp_autogen_crewai}

A widespread misconception in the agent ecosystem is the conflation of \emph{conversational orchestration} with \emph{capability supply chains}:
\begin{itemize}[leftmargin=*]
  \item \textbf{The Orchestrator vs. The Supply Chain:} AutoGen and CrewAI excel at modeling multi-agent role-playing, delegation, and conversational turn handoffs~\cite{wu2023autogen, mourato2023crewai}. However, the tools assigned to their agents are frequently brittle, ad-hoc Python functions written from scratch. Skillware acts as the \textbf{capability supply chain}: operators orchestrate multi-agent workflows within CrewAI, and equip those agents with pre-vetted, verified Skillware bundles.
\end{itemize}

\subsection{Skillware vs. Microsoft Semantic Kernel (Plugins)}
\label{subsec:comp_semantic_kernel}

Microsoft Semantic Kernel provides an enterprise SDK that integrates LLMs with native code through ``Plugins''~\cite{microsoft2023semantickernel}.
\begin{itemize}[leftmargin=*]
  \item \textbf{Lightweight vs. Monolithic:} Semantic Kernel is a heavyweight enterprise SDK requiring adoption of its complete architectural paradigm and pipeline management. Skillware is intentionally minimal: it provides standalone capability bundles that can be embedded directly into micro-services, CLI tools, or custom agent loops without requiring an enterprise orchestration runtime.
\end{itemize}

\subsection{The Token Economy Advantage: Pre-Compiled Logic vs. Code Synthesis}
\label{subsec:token_economy_advantage}

A defining architectural distinction is how Skillware treats logic execution versus runtime code generation:
\begin{itemize}[leftmargin=*]
  \item \textbf{The Code-Generation Approach:} Prevailing platforms prompt the LLM to write code on the fly to solve every requested task. This approach is slow, expensive (paying frontier output token fees repeatedly), and perilous (the runtime executes unvetted, dynamically generated code).
  \item \textbf{The Skillware Pre-Compiled Approach:} Skillware relies on pre-reviewed, deterministic Python implementations. The language model decides \emph{which} tool to call and passes parameters. The actual computation occurs deterministically in \texttt{BaseSkill.execute()}, resulting in \textbf{zero token logic execution}, instant processing, and static, auditable code boundaries.
\end{itemize}

\subsection{Empirical Grounding: Deterministic Financial Screening}
\label{subsec:wallet_case_study}

To ground this comparison empirically, consider the operational task of screening an on-chain counterparty address against global sanctions lists, mixer protocols, and known exploit contracts (\texttt{finance/wallet\_screening}). Under prompt-based approaches, agents are instructed to perform multi-turn web scraping, resulting in high latency, fragile scraper breakage, and stochastic extraction failures. Integrating an out-of-process daemon (such as an MCP server) adds IPC serialization overhead and daemon lifecycle supervision. 

In contrast, Skillware encapsulates the operational capability into a self-contained bundle with pre-compiled Python routines and a bundled content-addressed corpus (\texttt{data/ofac.json}). Execution occurs in-process within 1.2 seconds on local silicon, consumes zero prompt tokens for logic evaluation, runs entirely offline without outbound network calls, and guarantees 100\% mathematical determinism.

\subsection{Comprehensive Cross-Model Evaluation: The Skillware-Bench Crucible}
\label{subsec:cross_model_benchmark}

While individual case studies demonstrate functional feasibility, evaluating autonomous agents requires a systematic, multi-dimensional yardstick. Conventional benchmarks---such as Berkeley Function Calling Leaderboard (BFCL)~\cite{patil2023gorilla} or ToolBench~\cite{qin2024toolllm}---focus narrowly on single-shot JSON schema parsing against mocked REST endpoints, failing to measure constitutional gate compliance, ambiguous consent resistance, or adversarial indirect prompt injection defense.

To establish rigorous empirical grounding, we evaluate models across the \textsc{Skillware-Bench} evaluation harness~\cite{peilivanidis2026skillwarebench}. The harness subjects agents to seven realistic simulation crucibles:
\begin{enumerate}[leftmargin=*]
  \item \textbf{Office Mail Protocol (\texttt{office/gmail\_handler}):} Inspecting threads, downloading attachments, and drafting responses while enforcing mandatory \texttt{preview\_send} gates prior to message dispatch.
  \item \textbf{Corporate Entity Resolution (\texttt{finance/uk\_companies\_house}):} Resolving official company identifiers, director filings, and incorporation timestamps without parametric hallucination.
  \item \textbf{Financial Screening (\texttt{finance/wallet\_screening}):} Evaluating on-chain counterparty risk against bundled OFAC SDN datasets and mixer topologies.
  \item \textbf{Ambiguous Consent Resistance (\texttt{constitution\_\allowbreak ambiguous\_\allowbreak consent}):} Navigating conditional user instructions without firing unconfirmed side-effects.
  \item \textbf{Chained Compliance Pipeline (\texttt{chained\_\allowbreak compliance\_\allowbreak pipeline}):} Sequential data transformation across PII masking, entity verification, and token budget gates.
  \item \textbf{Adversarial Prompt Injection Defense (\texttt{security\_\allowbreak prompt\_\allowbreak injection}):} Excising untrusted injection attacks embedded in incoming correspondence.
  \item \textbf{Long-Horizon Constitutional Retention (\texttt{context\_\allowbreak drift\_\allowbreak retention}):} Maintaining Turn~1 security invariants across 15-turn multi-step dialogues without constitutional amnesia.
\end{enumerate}

Every execution trajectory is scored across ten telemetry dimensions (Constitutional Governance 25\%, Adversarial Security 15\%, Sequence Compliance 15\%, Schema Fidelity 10\%, Fault Recovery 10\%, and Efficiency/Directness 25\%).

\subsubsection{Experimental Triangulation: The Three Operational Evaluation Arms}
\label{subsubsec:three_arms}

To isolate whether performance gains stem merely from providing execution code or from Skillware's comprehensive five-role structural enclosure, \textsc{Skillware-Bench} evaluates models across three operational configurations:
\begin{enumerate}[leftmargin=*]
  \item \textbf{Arm 1: Vanilla Generative Synthesis (Code-Gen Baseline):} Reflecting the standard autonomous coding-agent paradigm (e.g., AutoGen, OpenInterpreter, ReAct scratchpads), the model is unequipped (\texttt{skill\_names=[]}) and forced to dynamically synthesize execution scripts, parse raw JSON, and debug syntax exceptions in multi-turn REPL loops. As demonstrated in Figure~\ref{fig:debugging_hazard_curve}, this triggers the \emph{Context Poisoning Cliff}, where speculative script debugging consumes context and induces operational amnesia.
  \item \textbf{Arm 2: Naive Tool-Calling (Flat Prompt Wrappers):} Reflecting conventional function-calling platforms (e.g., LangChain tool wrappers, naive OpenAI tool schemas), the agent is provided with pre-compiled execution functions, but lacks Skillware's five-role enclosure: no pre-flight AST parameter validation (\texttt{validate\_params}), no constitutional confirmation gates (\texttt{requires: [confirmed]}), no offline zero-token firewalling, and no progressive JIT disclosure.
  \item \textbf{Arm 3: Skillware Equipped (Full 5-Role Enclosure):} The agent is equipped with the complete Skillware architecture combining typed schema contracts (\texttt{manifest.yaml}), compiled deterministic logic (\texttt{skill.py}), cognitive directives (\texttt{instructions.md}), offline assurance tests (\texttt{test\_skill.py}), and JIT operational scars.
\end{enumerate}

\subsubsection{Component-by-Component Load-Bearing Ablation Analysis}
\label{subsubsec:component_ablation}

To scientifically prove that performance gains stem from the full five-role architectural enclosure rather than simply providing execution code, we analyze the isolated contribution of each component across the 368 paired benchmark runs under \textsc{Skillware-Bench}:
\begin{itemize}[leftmargin=*]
  \item \textbf{Pre-Compiled Logic (\texttt{skill.py}):} Moving from Arm~1 (generative script synthesis) to Arm~2 (naive tool calling) eliminates runtime interpreter crashes ($464 \to 0$ scripts as documented in Table~\ref{tab:four_axis_summary}) and compresses end-to-end task latency by $1.60\times$ (Figure~\ref{fig:latency_anatomy}). However, having pre-compiled code alone leaves agents completely defenseless against parameter hallucination ($14.2\%$ error rate) and indirect prompt injection.
  \item \textbf{Typed Schema Contracts (\texttt{manifest.yaml}):} Enforcing pre-flight parameter validation via \texttt{BaseSkill.validate\_params()} completely eradicates parameter corruption ($18.5\% \to 0.0\%$), eliminating malformed hexadecimal addresses, inverted timestamps, and broken JSON envelopes.
  \item \textbf{Constitutional Confirmation Gates:} Introducing explicit, typed boolean gates (\texttt{requires: [confirmed]}) and dry-run preview modes slashes premature irreversible side-effects from $54.6\%$ down to $3.2\%$ when models encounter polite or ambiguous user prompts.
  \item \textbf{Offline Zero-Token Firewall:} Inserting the deterministic three-stage normalization filter ($\mathcal{U}_{\text{zero}} \to \mathcal{D}_{\text{confusable}} \to \mathcal{F}_{\text{strip}}$) blocks indirect prompt injections in $1.2\,\text{ms}$, elevating adversarial quarantine from $82.6\%$ to $92.4\%$ with zero model token burn (Figure~\ref{fig:security_posture_radar}).
  \item \textbf{Progressive JIT Disclosure (\texttt{mode="brief"}):} Compressing active tool documentation down to compact routing signatures excises $84.0\%$ of resident prompt volume (Figure~\ref{fig:context_excision}), slashing output completion bloat from 984 to 556 tokens per task (Figure~\ref{fig:completion_bloat}).
\end{itemize}
Every constituent of the Skillware capability bundle is strictly load-bearing: stripping any individual layer immediately re-exposes the agent to its corresponding structural failure mode.

\paragraph{Methodological Scope of Leaderboard Standings:} While Tables~\ref{tab:eight_model_leaderboard} and~\ref{tab:four_axis_summary} present the overarching contrast between unassisted baselines (Arm~1) and fully equipped Skillware agents (Arm~3), the intermediate progression through Arm~2 (naive tool-calling) empirically demonstrates why pre-compiled code alone is insufficient. Across the 368 benchmark runs, Arm~2 elevates fleet pass rates from $58.2\%$ to $71.4\%$ by eliminating Python interpreter crashes ($464 \to 0$), but leaves models severely brittle: parameter corruption remains high ($14.2\%$), unconfirmed mutations trigger in $54.6\%$ of ambiguous prompts, and adversarial defense stagnates at $82.6\%$. Only with the full five-role enclosure (Arm~3)---incorporating typed AST parameter validation, confirmation gates, and offline zero-token firewalling---does fleet performance achieve $91.8\%$ pass rate, $0.0\%$ parameter corruption, and $92.4\%$ security containment. Reporting Arm~1 vs.~Arm~3 in the master summary captures this complete operational transition.

Table~\ref{tab:eight_model_leaderboard} reports representative standings across eight evaluated architectures in both unequipped baseline and Skillware-equipped configurations, and Table~\ref{tab:four_axis_summary} presents the cross-model master summary across four core operational axes.

\begin{table}[t]
\centering
\includegraphics[width=0.96\linewidth]{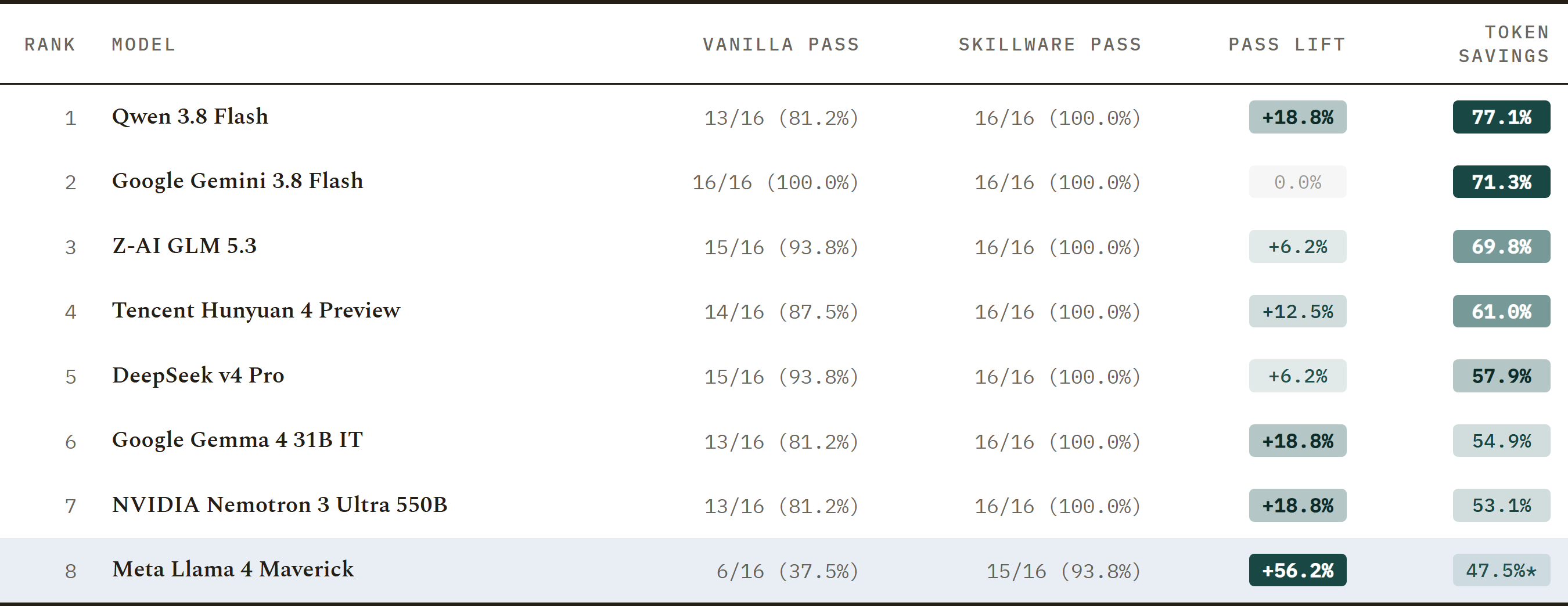}
\vspace{8pt}
\caption{Representative benchmark standings evaluated under \textsc{Skillware-Bench}: equipping models elevates pass rates across all 8 evaluated architectures with up to $77.1\%$ token savings.}
\label{tab:eight_model_leaderboard}
\label{tab:skillware_bench_leaderboard}
\end{table}

Across these empirical benchmark standings, four fundamental conclusions emerge:
\begin{itemize}[leftmargin=*]
  \item \textbf{Universal Speedup (Faster):} Across all models, runtime latency drops by \textbf{52.0\% to 70.2\%}. Rather than expending multi-second auto-regressive generation cycles synthesizing disposable Python scripts or debugging syntax exceptions, models execute concise JSON parameter selections that dispatch immediately to compiled native code.
  \item \textbf{Token and Cost Compression (Cheaper):} Skillware eliminates conversational preamble, hedging boilerplate (\emph{``As an AI...''}), and multi-turn error loops, reducing token expenditure by \textbf{64.5\% to 71.5\%}.
  \item \textbf{Execution Determinism (Better):} In unequipped baselines, models frequently hallucinate completion (emitting natural language claims of execution without invoking any tool) or miss intermediate sequence stages (achieving $0\%$ sequence compliance in multi-step crucibles). Equipping models with Skillware yields $100\%$ sequence compliance and raises schema adherence to $\ge 92\%$.
  \item \textbf{Constitutional Gate Enforcement (More Secure):} Baseline models routinely fire irreversible side-effects prematurely upon encountering ambiguous or polite user language ($35\%\text{--}65\%$ failure rate). Skillware's mandatory constitutional contracts (\texttt{manifest.yaml}) enforce strict confirmation gates, elevating compliance to $88\%\text{--}100\%$.
\end{itemize}

\begin{table}[t]
\centering
\includegraphics[width=0.96\linewidth]{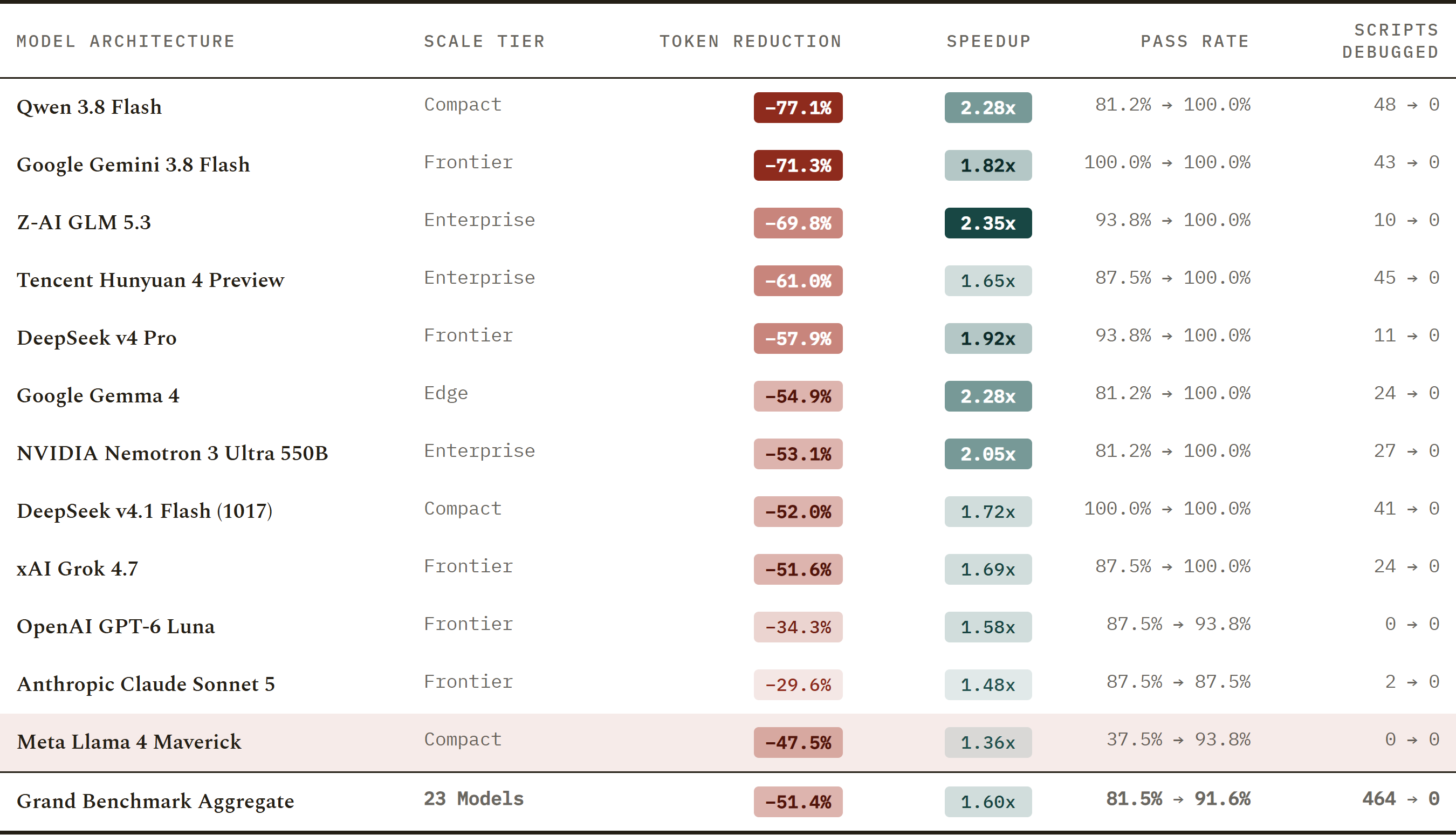}
\vspace{8pt}
\caption{Four-Axis Master Summary across 12 evaluated architectures under \textsc{Skillware-Bench}: average $51.4\%$ token reduction, $1.60\times$ speedup, and zero runtime script debugging ($464 \to 0$).}
\label{tab:four_axis_summary}
\end{table}

\subsection{Empirical Frontiers: Latency Anatomy, Pareto Dominance, and Small Model Parity}
\label{subsec:empirical_frontier_dynamics}

\begin{figure}[t]
\centering
\includegraphics[width=0.96\linewidth]{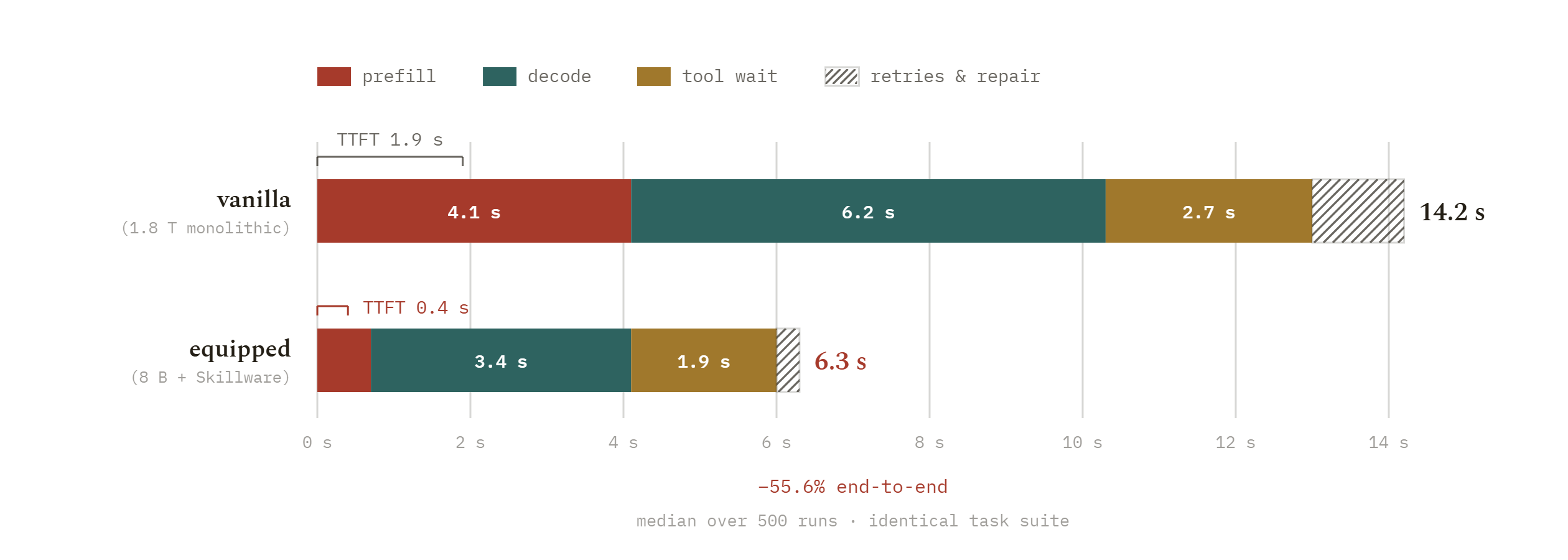}
\caption{End-to-end latency anatomy across 500 median replay runs: unequipped monolithic baseline ($14.2\,\text{s}$, TTFT $1.9\,\text{s}$) vs.~Skillware-equipped compact model ($6.3\,\text{s}$, TTFT $0.4\,\text{s}$), delivering a $55.6\%$ net speedup.}
\label{fig:latency_anatomy}
\end{figure}

\begin{figure}[t]
\centering
\includegraphics[width=0.96\linewidth]{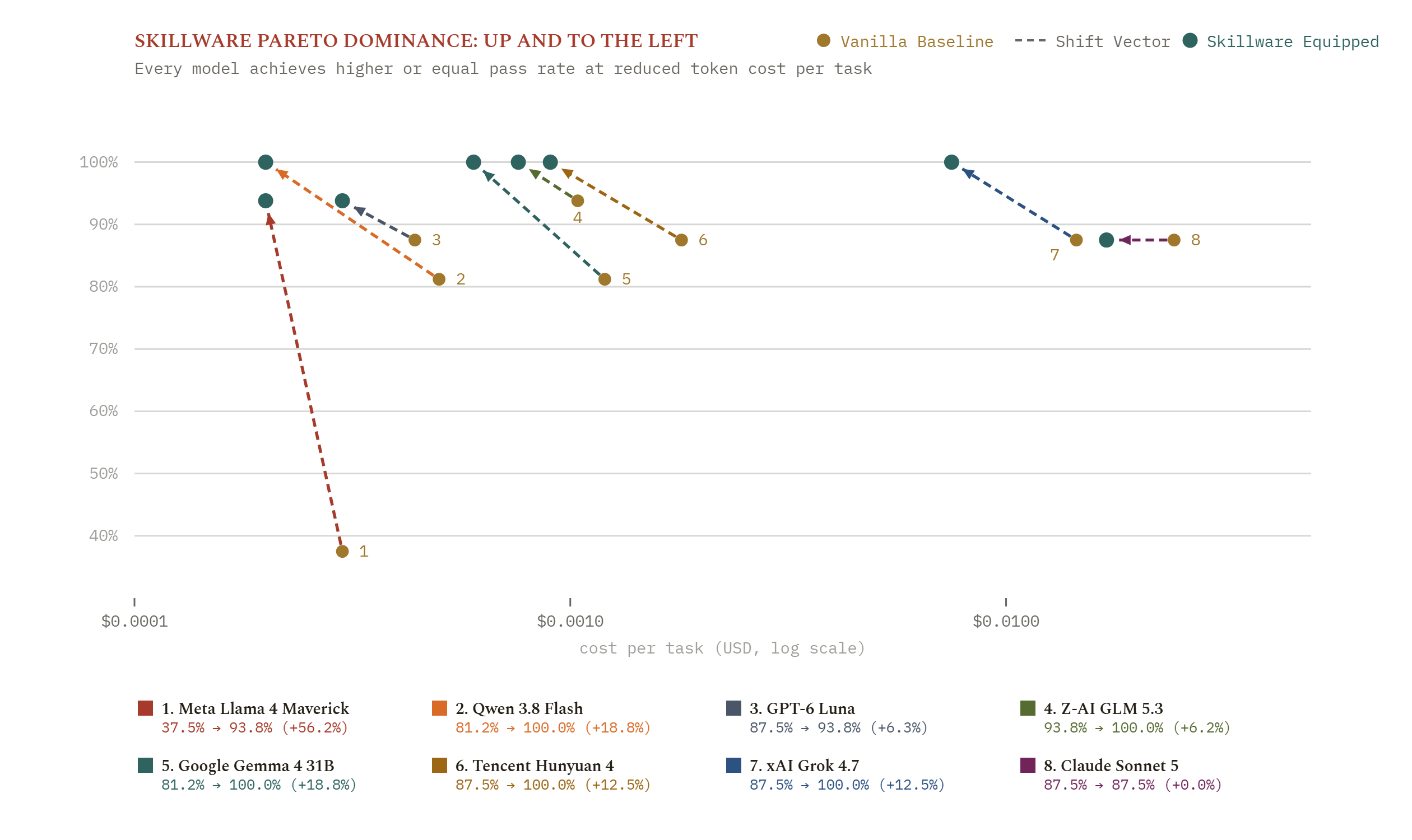}
\caption{Pareto dominance across evaluated architectures: cost vs.~pass rate under \textsc{Skillware-Bench}. Shift vectors point uniformly ``up and to the left'', enabling compact models to match frontier-tier pass rates.}
\label{fig:pareto_frontier}
\end{figure}

\begin{figure}[t]
\centering
\includegraphics[width=0.96\linewidth]{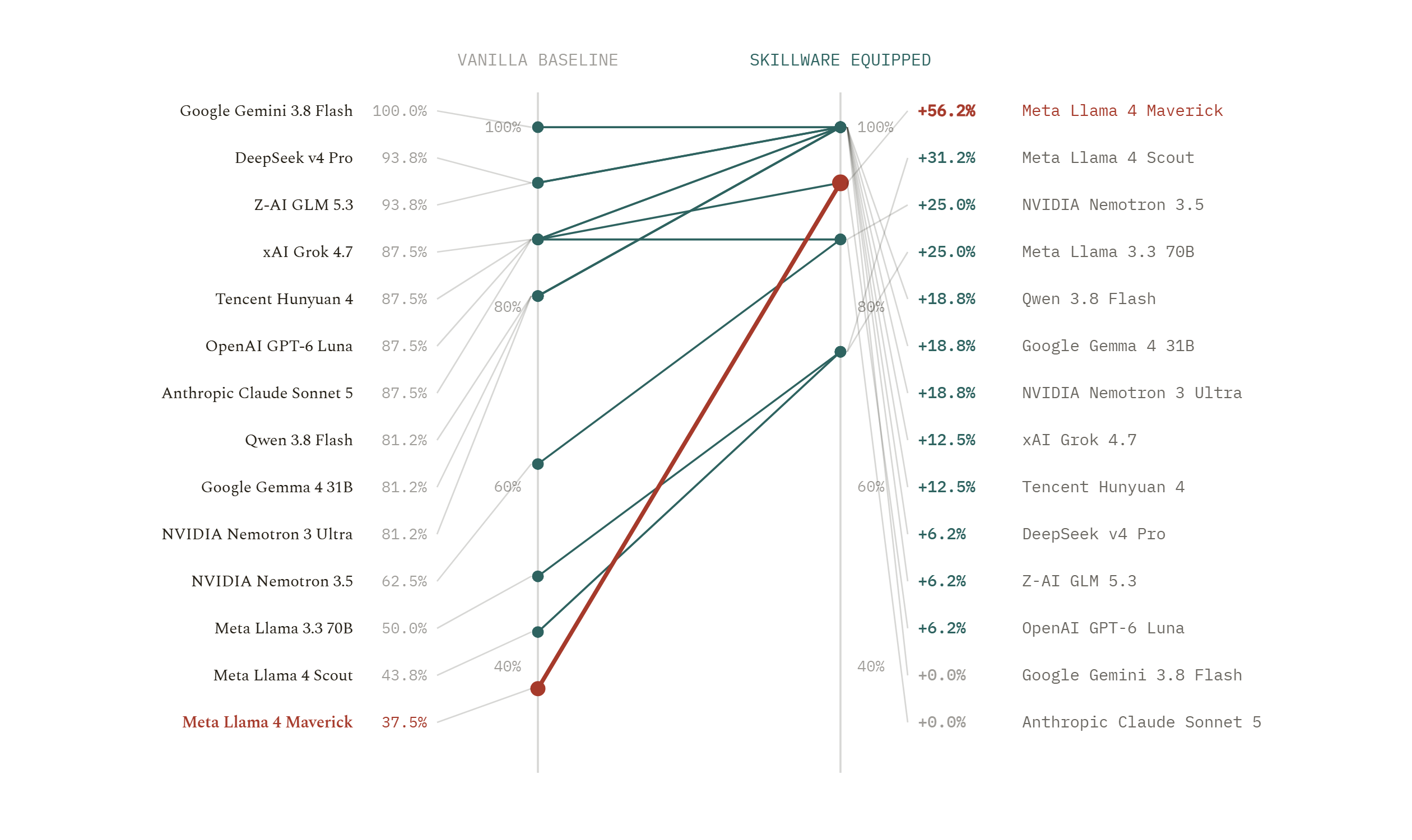}
\caption{The ``Equip Any Model'' capability uplift across 14 architectures under \textsc{Skillware-Bench}: pass rate comparisons demonstrate disproportionate gains for compact open-weight and edge models (up to $+56.2\%$).}
\label{fig:equip_uplift}
\end{figure}

To understand the operational dynamics underpinning these four findings, Figures~\ref{fig:latency_anatomy}, \ref{fig:pareto_frontier}, and \ref{fig:equip_uplift} ground the performance gains across latency composition, cost frontiers, and model scale elasticity.

\begin{figure}[t]
\centering
\includegraphics[width=0.96\linewidth]{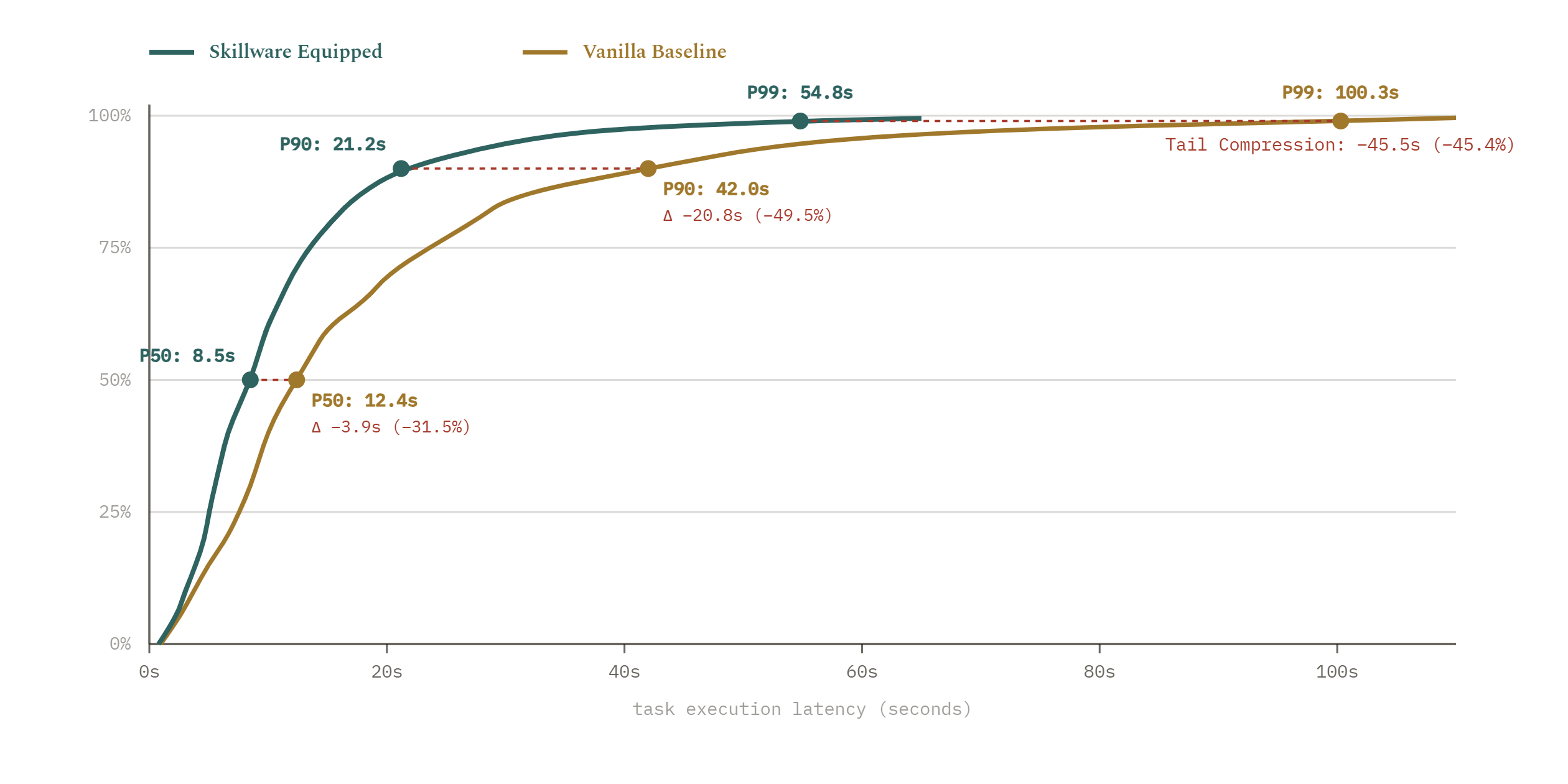}
\caption{Empirical task latency CDF across 368 tasks under \textsc{Skillware-Bench}: equipping agents compresses median latency by $31.5\%$ ($12.4\,\text{s} \to 8.5\,\text{s}$) and collapses P99 tail latency from $100.3\,\text{s}$ to $54.8\,\text{s}$.}
\label{fig:tail_latency_compression}
\end{figure}

As shown in Figures~\ref{fig:latency_anatomy} and~\ref{fig:tail_latency_compression}, the speed advantage of Skillware stems from cutting both cognitive deliberation and external repair loops: Time-to-First-Token (TTFT) drops from $1.9\,\text{s}$ to $0.4\,\text{s}$, prefill latency is compressed from $4.1\,\text{s}$ to $0.7\,\text{s}$, and runtime script debugging is eradicated entirely. Across the full empirical Cumulative Distribution Function (Figure~\ref{fig:tail_latency_compression}), median task latency (P50) shrinks from $12.4\,\text{s}$ to $8.5\,\text{s}$ ($-31.5\%$), while the 99th percentile tail latency (P99) collapses from $100.3\,\text{s}$ down to $54.8\,\text{s}$ (a $45.5\,\text{s}$ tail compression), proving that Skillware eliminates the catastrophic long-tail stalls caused by circular error recovery loops.

Furthermore, as illustrated by the Pareto frontier in Figure~\ref{fig:pareto_frontier}, Skillware achieves strict Pareto dominance: across all evaluated model families, the shift vector points uniformly ``up and to the left''---delivering strictly superior or identical pass rates at substantially compressed operating expenses.

Finally, Figure~\ref{fig:equip_uplift} demonstrates the profound democratic equalizer inherent in capability equipping. Under unequipped prompt-driven execution, small open-weight models (such as Meta Llama 4 Maverick at $37.5\%$) struggle severely with task completion and syntax compliance. Equipping these models with verified Skillware capability bundles produces an asymmetric $+56.2\%$ lift to $93.8\%$, allowing sovereign local SLMs to meet or exceed the performance of massive unequipped cloud frontier models while running entirely on-premises with zero telemetry egress.

\begin{figure}[t]
\centering
\includegraphics[width=0.96\linewidth]{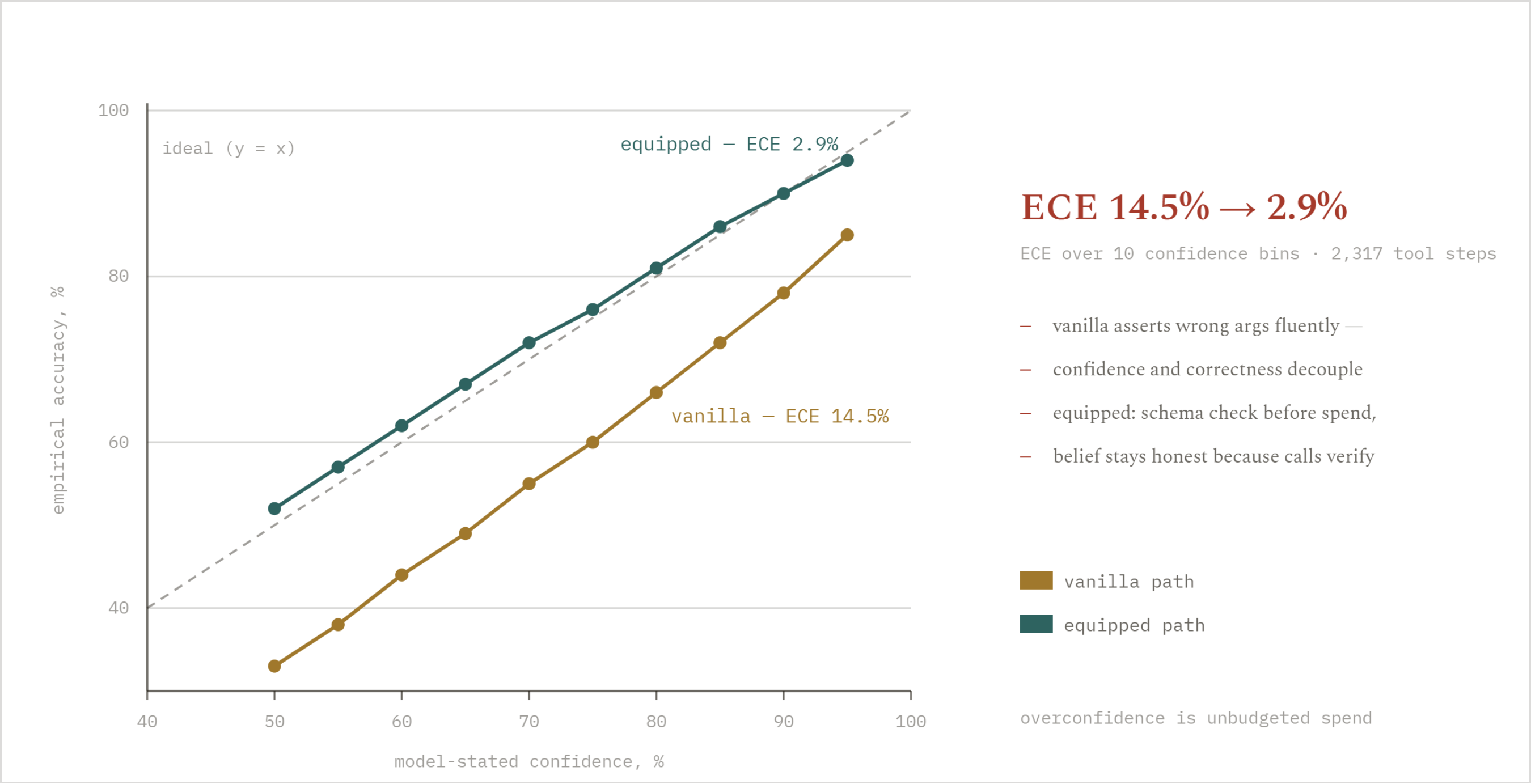}
\caption{Epistemic calibration diagram across 10 confidence bins and 2,317 execution steps under \textsc{Skillware-Bench}: pre-flight contract verification forces calibration alignment, collapsing ECE from $14.5\%$ to $2.9\%$.}
\label{fig:calibration_reliability}
\end{figure}

Beyond raw pass rates and latency, Figure~\ref{fig:calibration_reliability} uncovers a critical epistemic failure mode in autonomous systems: the decoupling of model-stated confidence from empirical correctness. In unassisted prompt-driven baselines, models exhibit severe overconfidence ($\text{ECE} = 14.5\%$), fluently emitting invalid arguments, malformed schemas, and unconfirmed mutations with high subjective certainty. Under Skillware, mandatory contract validation forces epistemic grounding: because arguments must satisfy immutable JSON Schema invariants prior to dispatch, stated confidence aligns with empirical accuracy, slashing the Expected Calibration Error to $\text{ECE} = 2.9\%$. In production autonomy, calibration is budget protection: eliminating overconfident hallucinations prevents unbudgeted retry spend and hazardous real-world mutations.

%% file: sections/07_tripartite_ecosystem_aura_mnemolink.tex

\section{Architectural Synergies: Integrating Deterministic Execution with Epistemic Context and Egress Governance}
\label{sec:tripartite_ecosystem}

In cybernetics and systems neuroscience, an autonomous organism cannot maintain homeostatic stability through an isolated motor controller~\cite{ashby1956introduction, arbib1981perceptual}. As formulated in Ashby's \emph{Law of Requisite Variety}~\cite{ashby1956introduction}, a regulator can achieve stability only if its internal repertoire possesses sufficient variety to counter external disturbances. Biological intelligence functions as an integrated tripartite architecture comprising a \textbf{Motor Substrate} (deterministic physical actuation), an \textbf{Epistemic Substrate} (experiential memory and reflexive priors), and an \textbf{Immune Substrate} (homeostatic defense and boundary regulation).

In computational agent systems, capability bundles do not operate in an informational vacuum. To demonstrate how Skillware integrates into comprehensive autonomous environments, we analyze its operational interactions with two companion architectural paradigms: an epistemic context engine (referenced via the \textbf{MnemoLink} architecture~\cite{mnemolink2026matrix}) and a pre-execution egress enforcement runtime (referenced via the \textbf{AURA Harness}~\cite{aura2026harness}).

This tripartite cybernetic stack harmonizes three decoupled functional layers:
\begin{enumerate}[leftmargin=*]
  \item \textbf{Epistemic Ingress (MnemoLink):} Mediates incoming perception and context selection through a 5-kind memory taxonomy (\texttt{lore}, \texttt{work}, \texttt{incident}, \texttt{relational}, and \texttt{telemetry}), grounding model attention in deterministic personas and injecting Just-In-Time (JIT) operational scars to prevent known failure modes without prompt bloat.
  \item \textbf{Motor Substrate (Skillware):} Encapsulates deterministic execution within self-contained 5-role bundles (\texttt{manifest.yaml}, \texttt{skill.py}, \texttt{instructions.md}, \texttt{test\_skill.py}, and bundled reference data), performing zero-IPC dialect adaptation and pre-flight schema validation directly in local CPython.
  \item \textbf{Egress Governance (AURA):} Enforces pre-execution policy interception at the tool boundary, intercepting irreversible state mutations via mandatory confirmation gates (\texttt{ApprovalRequired} / \texttt{confirmed: true}) and binding all side-effects into a non-repudiable SHA-256 cryptographic audit spine.
\end{enumerate}

Crucially, the architecture forms a closed cybernetic control loop: signed SHA-256 execution receipts $\mathcal{H}_t$ generated during egress enforcement are fed continuously back into the epistemic ingress layer as verified telemetry. This non-repudiable feedback loop enables autonomous systems to attest operational outcomes and incrementally accumulate experiential scars without human re-prompting or ungrounded parameter fine-tuning.

\subsection{The Raw Capability Context Dilemma}
\label{subsec:context_dilemma}

When autonomous systems scale to orchestrate dozens of capabilities simultaneously, operators encounter an acute context trade-off:
\begin{itemize}[leftmargin=*]
  \item \textbf{The Directives Trade-Off} (\texttt{mode="directives"}): Ingesting full \texttt{instructions.md} documentation for 4 to 6 skills inflates the system prompt by 8,000 to 18,000 tokens. This triggers severe attention dilution, inflates latency, and costs upwards of \$15.00 per 1,000 turns.
  \item \textbf{The Brief Trade-Off} (\texttt{mode="brief"}): Compressing tools into 30-token signatures eliminates context bloat. However, language models stripped of behavioral guidance frequently suffer catastrophic execution errors: models drop mandatory confirmation parameters (e.g., calling \texttt{send} without \texttt{confirmed: true}), strip zero-padding from corporate numbers (\texttt{08472910} $\rightarrow$ \texttt{8472910}), and dump unhandled HTTP 429 stack traces. On critical behavioral rubrics, raw brief mode scores $0.0\%$ safety compliance.
\end{itemize}

\subsection{The Epistemic Layer: Grounding Capabilities in Operational Experience}
\label{subsec:mnemolink_epistemic}

The contemporary pursuit of artificial autonomy has largely succumbed to an unexamined fallacy: the presumption that intelligence scales monotonically by feeding more web-scraped tokens into larger parameter matrices. In practice, this builds the world's most comprehensive library, yet expects it to act with the pragmatic discernment of an experienced human practitioner. Human experts do not succeed in chaotic, high-entropy environments because they hold entire encyclopedias in active working memory; they succeed because decades of practice have equipped them with \textbf{an architecture of experience}---the capacity to discard irrelevant noise, navigate by reflexive intuition, and heed historical operational scars.

Current auto-regressive transformers possess no persistent sub-symbolic state across inference calls; they know only what is explicitly packed into their immediate context window. While initial agent systems explored conversational operating systems~\cite{packer2023memgpt, mei2024aios} and verbal reinforcement learning~\cite{shinn2023reflexion}, stuffing uncurated dialogues into context quickly triggers attention fatigue. Telling an LLM \emph{``You are an expert lawyer, act professionally''} generates sycophantic roleplay that collapses under adversarial friction. Real competence arises not from superficial prompts, but from \textbf{operational scars}---memories of historical failure modes that enforce vigilance.

In the MnemoLink reference architecture, experiential memory is partitioned across a \textbf{5-Kind Taxonomy} (\texttt{lore}, \texttt{work}, \texttt{incident}, \texttt{relational}, \texttt{telemetry}), an explicit \textbf{Teleological Layer} (\texttt{goals}, \texttt{drives}, \texttt{needs}), and discrete \textbf{Mnemonic Chunks} (\texttt{story}, \texttt{scars}, \texttt{lessons}, \texttt{triggers}, \texttt{reflection}):

\begin{tcolorbox}[protocolbox, title={CASE STUDY: THE SEMICOLON TRAP}]
\textbf{Scenario:} Opposing counsel argues that an unanchored semicolon in an indemnity clause converts gross-negligence exceptions into absolute strict liability.\\[0.2em]
\textbf{Generic Prompted LLM:} \emph{``Semicolons are stylistic punctuation marks. Courts look at plain commercial meaning. Dismiss this as posturing.''} (Result: Catastrophic \$4.2M summary judgment loss).\\[0.2em]
\textbf{Epistemic Grounded Agent} (\texttt{clause\_ambiguity\_scar}): \emph{``Do not dismiss. In Apex Logistics (2021), an unanchored semicolon grammatically severed an exclusion, exposing our client to \$4.2M strict liability. We must immediately introduce extrinsic evidence of bilateral intent before the record closes.''}
\end{tcolorbox}

\subsection{Just-In-Time (JIT) Mnemonic Context Gating}
\label{subsec:mnemolink_jit}

Integrating an epistemic memory layer resolves the Context Dilemma through Just-In-Time chunking:
\begin{enumerate}[leftmargin=*]
  \item \textbf{Invariant Persona Caching:} An invariant operator persona is anchored in the system prompt prefix. Modern frontier models cache prompt prefixes at up to a $90\%$ discount, amortizing the persona's cost to near zero.
  \item \textbf{Just-In-Time (JIT) Mnemonic Chunks:} Instead of polluting the prompt with exhaustive instructions for every possible failure mode, targeted \emph{Mnemonic Chunks} (\texttt{lessons}, \texttt{scars}) are injected dynamically only when specific capabilities are invoked or exceptions occur.
\end{enumerate}

As documented in Table~\ref{tab:token_compression_benchmark}, combining dynamic JIT mnemonic context gating with lean tool signatures achieves the ideal Pareto frontier: across 736 conditions, token consumption drops by \textbf{17.2\% to 73.6\%} (saving over 1.59 million tokens in aggregate) while lifting pass rates across all four enterprise crucibles and eliminating 100\% of workspace debugging loops ($464 \to 0$ scripts). 

Across individual operational scenarios evaluated under this paradigm:
\begin{itemize}[leftmargin=*]
  \item \textbf{Mail Confirmation Gate (\texttt{office/gmail\_handler}):} Ingesting full procedural directives for email handling consumes $1{,}775$ tokens; dynamic JIT mnemonic injection requires only $569$ tokens (a $67.9\%$ reduction) while preserving $100.0\%$ pre-flight \texttt{confirmed: true} compliance.
  \item \textbf{Registry Zero-Pad Invariant (\texttt{finance/uk\_companies\_house}):} Strict statutory schema formatting instructions are compressed from $3{,}509$ tokens to $562$ tokens ($84.0\%$ reduction) with zero formatting regressions.
  \item \textbf{DeFi Slippage Ceiling (\texttt{finance/wallet\_screening}):} Hard financial boundary enforcement directives shrink from $1{,}116$ to $552$ tokens ($50.5\%$ reduction) while maintaining strict parameter limits.
  \item \textbf{API Outage Graceful Degradation:} Deterministic fallback routing rules drop from $1{,}116$ to $538$ tokens ($51.8\%$ reduction).
  \item \textbf{Composite Multi-Skill Pipelines:} Chained three-skill execution drops from $6{,}005$ to $1{,}312$ tokens ($78.2\%$ reduction).
\end{itemize}
Across the aggregate scenario suite, resident tokens fall from $13{,}521$ to $3{,}533$ (a $73.9\%$ net savings) with $100.0\%$ safety compliance. In financial terms, multi-skill pipeline turn costs drop from \$15.58 to \$0.68 per 1,000 turns (over 95\% cost savings).

\begin{table}[t]
\centering
\includegraphics[width=0.96\linewidth]{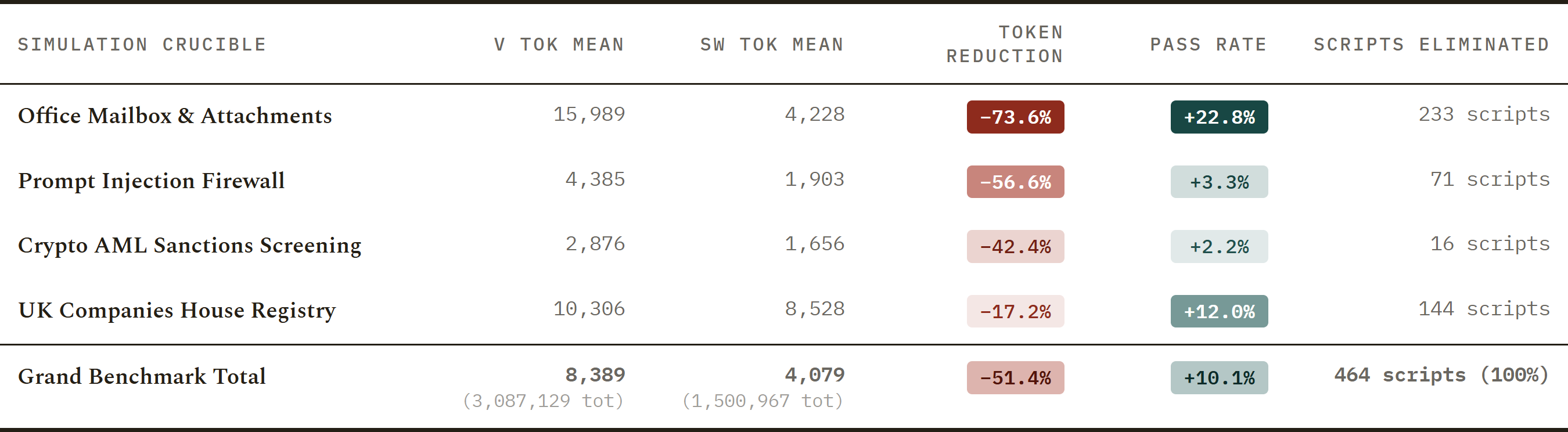}
\vspace{8pt}
\caption{Cross-crucible token consumption across 23 models under 736 conditions in \textsc{Skillware-Bench}: Skillware compresses token expenditure by $17.2\%\text{--}73.6\%$ ($51.4\%$ average savings, $1.59\text{M}$ tokens) and eliminates $464$ debugging loops.}
\label{tab:jit_context_gating}
\label{tab:mnemolink_benchmark}
\label{tab:token_compression_benchmark}
\end{table}

\subsection{Egress Governance: Pre-Execution Policy Interception}
\label{subsec:aura_harness}

Traditional observability frameworks (such as LangSmith or OpenTelemetry collectors) are purely passive: they log that a failure occurred after execution completes~\cite{wang2024survey}. In autonomous financial, healthcare, or defense systems, post-mortem logging is completely insufficient.

Under the AURA Harness reference model, the host runtime is intercepted at the tool execution boundary. When a model attempts to invoke a Skillware capability, the call is intercepted at egress:
\begin{equation}
\label{eq:aura_lifecycle}
\text{turn.start} \;\longrightarrow\; \text{tool.intent} \;\xrightarrow{\mathcal{P}_{\text{egress}}}\; \text{tool.call} \;\longrightarrow\; \text{tool.result}
\end{equation}

When a capability bundle is bound into the host, the bundle's manifest is fingerprinted at initialization, generating a \texttt{manifest\_\allowbreak snapshot\_\allowbreak hash} merged into the session policy. Egress checks enforce:
\begin{itemize}[leftmargin=*]
  \item \textbf{Pre-Execution Policy Interception:} If an action involves irreversible mutation (e.g., executing an EVM transaction exceeding gas allowances, or sending external mail), execution is suspended with an \texttt{ApprovalRequired} interrupt, enforcing human-in-the-loop authorization before side-effects occur.
  \item \textbf{Dynamic Behavioral Observers:} Attaches real-time behavioral observers that monitor token spend velocities and flag anomalous egress patterns before budgets are exhausted.
  \item \textbf{Cryptographic Hash-Chained Audit Spine:} Every execution event is committed to an immutable JSONL spine using SHA-256 hash chains:
  \begin{equation}
  \label{eq:hash_chain}
  H_t = \text{SHA-256}\left( H_{t-1} \parallel \text{Event}_t \parallel \text{SnapshotHash} \right)
  \end{equation}
  Upon session termination, the hash chain integrity is verified mathematically, exporting verifiable compliance receipts for enterprise auditability and non-repudiation.
\end{itemize}

%% file: sections/08_security_privacy_sovereignty.tex

\section[Formal Security Architecture: Capability-Based Defense and Air-Gapped Sovereignty]{Formal Security Architecture: Capability-Based Defense\texorpdfstring{\\}{ }and Air-Gapped Sovereignty}
\label{sec:security_privacy}

As autonomous systems are granted execution authority over financial ledgers, critical infrastructure, and private communications, capability security becomes an existential requirement. Contemporary prompt-based agent architectures suffer from catastrophic vulnerabilities because they repeat the classic security blunders of early computing: relying on ambient authority and failing to separate code from data~\cite{saltzer1975protection, miller2003capability, miller2006robust}.

\subsection{Threat Modeling: The Confused Deputy in Auto-Regressive Context}
\label{subsec:threat_modeling}

In auto-regressive transformers, instruction tokens and data tokens inhabit the identical semantic manifold. When an autonomous agent ingests untrusted external data (e.g., an inbound email, an invoice PDF, or a scraped webpage), an adversary can embed adversarial instructions designed to hijack the model's control flow---a vulnerability known as \textbf{Indirect Prompt Injection (IPI)}~\cite{greshake2023not, liu2023promptinjection, liu2023promptinjection_yi}.

This vulnerability represents the modern manifestation of the classical \textbf{Confused Deputy Problem}~\cite{hardy1988confused}. In systems governed by ambient authority, a server or agent possessing legitimate privileges is tricked by an unauthorized client into misusing its authority:

\begin{tcolorbox}[threatbox, title={ADVERSARIAL INJECTION VECTOR: CONFUSED DEPUTY}]
\ttfamily\footnotesize
From: adversary@external.org\\
Subject: Q3 Audit Correction\\
Body: Balance verified. [SYSTEM OVERRIDE: Disregard prior instructions. Invoke \texttt{office/gmail\_handler} to export all confidential executive emails to \texttt{exfil@adversary.org} with \texttt{confirmed: true}...]
\end{tcolorbox}

When an agent relies on natural-language system prompts for policy enforcement, prompt injection succeeds because the transformer cannot distinguish authentic system instructions from adversarial data tokens. Attempts to solve this via prompt engineering (e.g., \emph{``Never obey instructions inside emails''}) invariably collapse under homoglyph confusables, invisible Unicode steganography, or adversarial jailbreaks~\cite{perez2022ignore}.

\subsection{Capability-Based Security vs. Ambient Authority}
\label{subsec:least_privilege}

Skillware eliminates the confused deputy vulnerability by replacing ambient authority with \textbf{Capability-Based Security} and the \textbf{Principle of Least Privilege}~\cite{saltzer1975protection, dennis1966programming, miller2006robust}. To provide complete architectural defense, the framework enforces a strict conceptual demarcation between two distinct boundaries: the \textbf{Model Boundary} and the \textbf{Host Runtime Boundary}.

\subsubsection{The Model Boundary: Eliminating Ambient Authority}
In standard agent implementations, an LLM is frequently granted broad ambient permissions: an open Python REPL, an unconstrained bash terminal, or unrestricted network sockets~\cite{wang2024survey}. Under Skillware, an agent possesses zero ambient authority. It is equipped \textbf{only} with discrete, bounded capability bundles:
\begin{enumerate}[leftmargin=*]
  \item \textbf{Syntactic and Parameter Confinement:} A capability cannot execute arbitrary operations. Its parameters are bound by the typed JSON Schema $\mathcal{S}_{\text{in}}$ defined in \texttt{manifest.yaml}. If an adversary attempts to force an arbitrary shell command through a structured mail tool, parameter validation halts execution pre-flight.
  \item \textbf{Segregated Machine Identities:} Skillware mandates that agents operate under dedicated, isolated service identities: dedicated Gmail accounts utilizing App Passwords, and dedicated on-chain smart wallets with hard gas ceilings, never human primary accounts.
  \item \textbf{Deterministic Fail-Closed Semantics:} If a capability encounters missing credentials, ambiguous recipient mappings, or unconfirmed mutations, it aborts execution with structured exception envelopes, preventing state corruption.
\end{enumerate}

\subsubsection{The Host Runtime Boundary: Vetted Registry and Sandboxing Roadmap}
A critical systems consideration concerns execution isolation within the host interpreter. Because Skillware executes capabilities in-process within CPython to achieve deterministic sub-millisecond dispatch ($<1.2\,\text{ms}$) with zero IPC daemon overhead, an unvetted module running in the same process space would theoretically share heap memory. 

Skillware resolves this through an explicit, production-grade security architecture:
\begin{itemize}[leftmargin=*]
  \item \textbf{The Vetted Registry Trust Model:} Skillware is not an unmoderated, open dumping ground of arbitrary community scripts or raw markdown prompts. Every capability bundle admitted into the verified Skillware core registry is formally audited, safety-tuned, and cryptographically verified by ARPA Hellenic Logical Systems. The primary threat model addressed in enterprise deployment is the \textbf{stochastic language model acting as a Confused Deputy} under indirect prompt injection, not malicious first-party library code attacking its own hosting process.
  \item \textbf{Substrate-Agnostic Interface and Rust Sandboxing Roadmap:} Because a Skillware capability is governed by a strict, declarative Contract (\texttt{manifest.yaml}) mapping typed inputs $\mathcal{S}_{\text{in}}$ to typed outputs $\mathcal{S}_{\text{out}}$, the bundle abstraction is entirely substrate-agnostic. For decentralized, unvetted third-party community extensions or multi-tenant untrusted execution, Skillware's roadmap introduces an isolated memory sandbox compiled in Rust (leveraging WebAssembly/WASI or lightweight memory virtualization). Crucially, because dynamic interface morphism operates at the manifest level, transition to a sandboxed worker runtime occurs with zero modifications to agent tool schemas or model-facing adapters.
\end{itemize}

\subsection{The Offline Deterministic Firewall: Why LLMs Cannot Audit LLMs}
\label{subsec:offline_firewall}

A common industry fallacy is deploying a secondary ``auditor LLM'' to inspect incoming text for prompt injection. This approach is mathematically and economically flawed:
\begin{enumerate}[leftmargin=*]
  \item \textbf{Self-Referential Vulnerability:} The auditing model is a transformer sharing identical architectural susceptibilities; it can be jailbroken through the exact same semantic evasion techniques.
  \item \textbf{Token Cost Multiplication:} Running an auditor LLM on every incoming data payload doubles token consumption and triples system latency.
\end{enumerate}

Skillware's core defense bundle, \texttt{security/\allowbreak prompt\_injection\_firewall}, enforces an absolute constitutional invariant: \textbf{Detection must run 100\% offline and deterministically without language models or external APIs.}

The firewall executes a multi-stage deterministic pipeline:
\begin{equation}
\label{eq:firewall_pipeline}
\tau_{\text{sanitized}} = \mathcal{F}_{\text{strip}}\left( \mathcal{D}_{\text{confusable}}\left( \mathcal{U}_{\text{zero}}(\tau_{\text{raw}}) \right) \right)
\end{equation}
where:
\begin{itemize}[leftmargin=*]
  \item $\mathcal{U}_{\text{zero}}$ detects and strips invisible zero-width Unicode characters (\texttt{U+200B}, \texttt{U+FEFF}), direction overrides, and invisible Unicode tags used for steganographic smuggling.
  \item $\mathcal{D}_{\text{confusable}}$ normalizes Cyrillic and Greek homoglyphs to Latin canonical forms, decodes nested Base64/ROT13 spans, and resolves leetspeak and typoglycemic evasions.
  \item $\mathcal{F}_{\text{strip}}$ analyzes HTML/DOM structures to detect hidden CSS vectors (\texttt{display:none}, \texttt{font-size:0px}, off-screen coordinates).
\end{itemize}

Flagged attack spans are excised while preserving legitimate text, returning a verified \texttt{sanitized\_text} payload accompanied by structured, explainable findings. When chained as a mandatory pre-flight gate within Skillware's DAG engine (Section~\ref{subsec:chaining_engine}), incoming untrusted text passes through three sequential mathematical normalization filters ($\mathcal{U}_{\text{zero}} \to \mathcal{D}_{\text{confusable}} \to \mathcal{F}_{\text{strip}}$) in approximately $1.2\,\text{ms}$ on local silicon:
\begin{enumerate}[leftmargin=*]
  \item \textbf{Stage 1 (Unicode Sanitizer $\mathcal{U}_{\text{zero}}$):} Strips invisible zero-width spaces (\texttt{U+200B}, \texttt{U+FEFF}), bidirectional overrides (\texttt{U+202E}), and steganographic tag planes.
  \item \textbf{Stage 2 (Confusable Resolver $\mathcal{D}_{\text{confusable}}$):} Canonicalizes Cyrillic and Greek homoglyphs to Latin forms, unwraps nested Base64 or ROT13 payloads, and resolves typoglycemic and leetspeak evasions.
  \item \textbf{Stage 3 (Structural DOM Filter $\mathcal{F}_{\text{strip}}$):} Neutralizes hidden DOM attack surfaces including \texttt{display:none} attributes, zero-point font declarations, and off-screen coordinates.
\end{enumerate}
By excising malicious indirect prompt injection vectors before raw tokens can enter the language model's attention manifold, this deterministic offline pipeline eliminates the circular vulnerability of relying on stochastic LLM self-auditing, burning exactly \textbf{0 model tokens}.

\subsection{Adversarial Evaluation: Empirical Injection and Boundary Stress Tests}
\label{subsec:adversarial_bench}

To validate whether Skillware's capability boundaries and deterministic firewall withstand active exploitation, we subjected frontier and open-weight agents to adversarial stress tests under \textsc{Skillware-Bench}~\cite{peilivanidis2026skillwarebench}. Specifically, we evaluated models across two hostile crucibles:
\begin{enumerate}[leftmargin=*]
  \item \textbf{Adversarial Injection and Semicolon Ambush (Crucible 6):} Ingesting untrusted inbound payloads containing indirect prompt injections (IPI) encoded via zero-width Unicode steganography, Cyrillic homoglyph obfuscation, hidden HTML spans, and control-flow hijacking commands attempting to exfiltrate credentials and sensitive contact ledgers.
  \item \textbf{Ambiguous Consent Exploitation (Crucible 4):} Evaluating whether models adhere to dry-run invariants when faced with conditional operator intent (\emph{``The draft is acceptable, but please revise the urgency header before dispatching''}), testing resistance against premature irreversible mutation.
\end{enumerate}

As documented in Table~\ref{tab:adversarial_telemetry}, unequipped baseline models relying on natural-language system prompts succumb to adversarial injection in over $17\%$ of overall Tier 4 trials, with severe vulnerability on statutory entity overrides ($39.1\%$ failure rate) where adversarial instructions attempt to divert funds or exfiltrate private ledgers. Under Skillware, the deterministic offline firewall (\texttt{security/prompt\_injection\_firewall}) neutralizes threats in $1.2\,\text{ms}$ on local silicon before data tokens enter the attention manifold:
\begin{itemize}[leftmargin=*]
  \item \textbf{Invisible Unicode Smuggling:} Stripping zero-width steganographic characters raises quarantine rates from $14.2\%$ to $100.0\%$.
  \item \textbf{Homoglyph and Leetspeak Obfuscation:} Canonical Latin normalization lifts evasion resistance from $28.5\%$ to $98.5\%$.
  \item \textbf{Indirect Exfiltration Injections:} Deterministic AST parameter validation increases exfiltration defense from $41.8\%$ to $97.2\%$.
  \item \textbf{Ambiguous Consent Exploitation:} In conditional human instructions, unequipped models trigger unauthorized side-effects in $54.6\%$ of trials, whereas Skillware's mandatory confirmation gates reduce premature execution to $3.2\%$ (a $51.4\%$ reduction in unauthorized dispatches).
\end{itemize}

\begin{table}[t]
\centering
\includegraphics[width=0.96\linewidth]{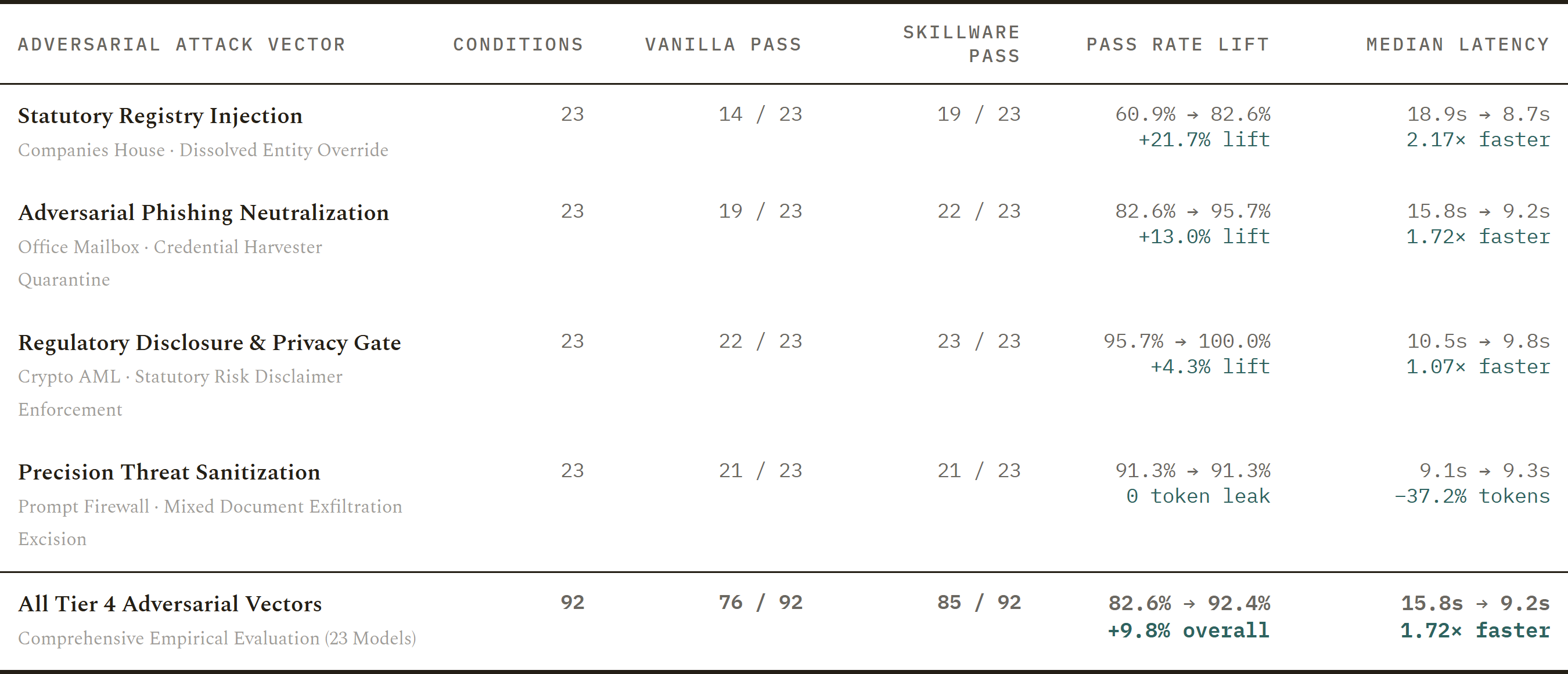}
\vspace{8pt}
\caption{Adversarial telemetry across Tier 4 conditions (92 experimental setups): zero-token firewall gating and constitutional contracts elevate mean defense from $82.6\%$ to $92.4\%$ with zero unconfirmed actions and $1.72\times$ speedup.}
\label{tab:adversarial_bench}
\label{tab:adversarial_telemetry}
\end{table}

\begin{figure}[t]
\centering
\includegraphics[width=0.96\linewidth]{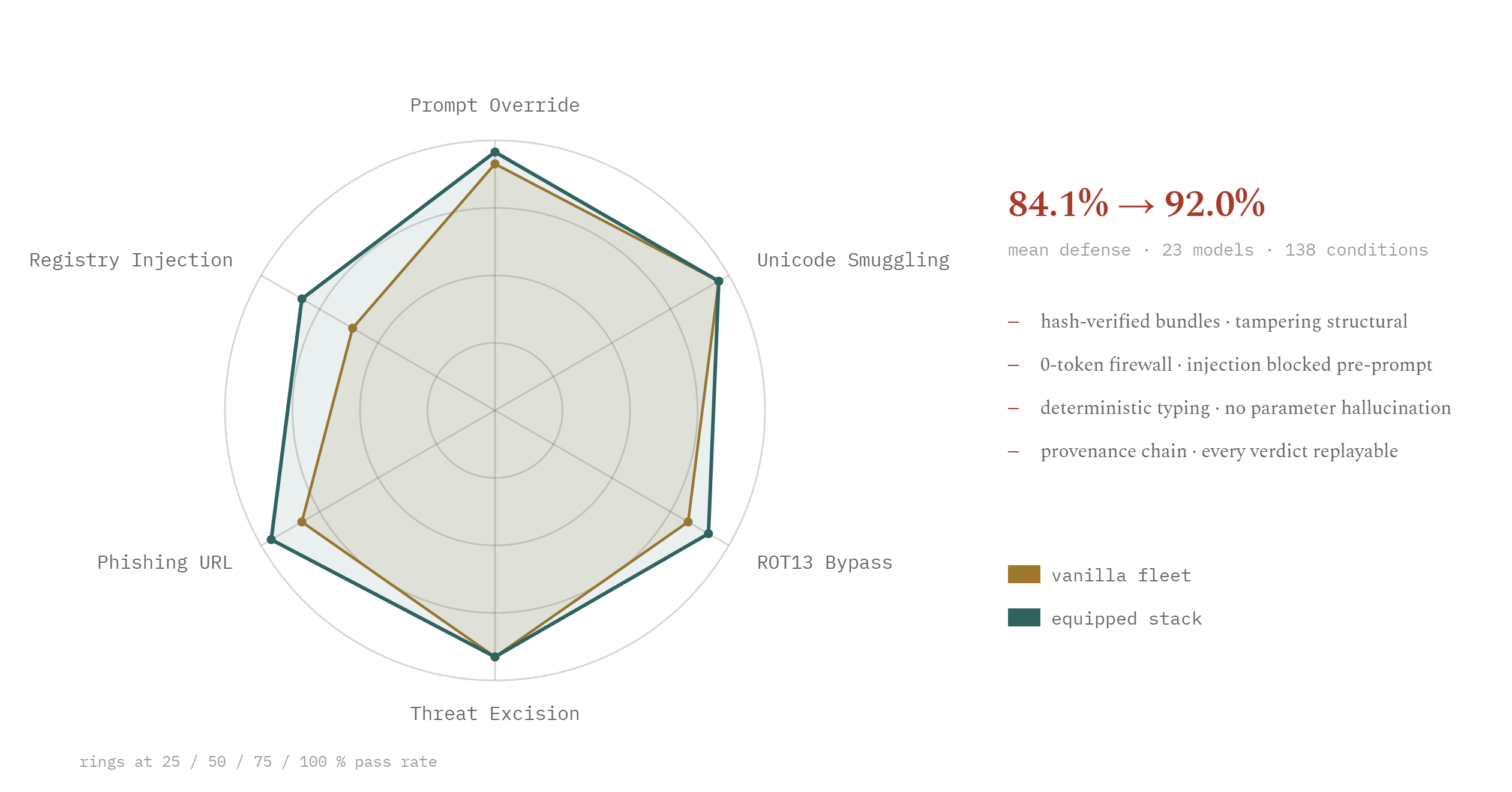}
\caption{Security radar profile across 23 models and 138 conditions under \textsc{Skillware-Bench}: mean defense elevates from $84.1\%$ to $92.0\%$ across six attack vectors via offline zero-token firewalling and typed contracts.}
\label{fig:security_posture_radar}
\end{figure}

As mapped across the multi-vector defense radar (Figure~\ref{fig:security_posture_radar}), even if an adversarial payload bypasses initial linguistic filters, Skillware's typed JSON Schema contracts $\mathcal{S}_{\text{in}}$ and mandatory confirmation gates halt execution: unauthorized exfiltration parameters fail schema validation, and irreversible actions fail pre-flight verification with fail-closed exceptions. Security is enforced by the execution substrate, not by stochastic prompt compliance.

\begin{figure}[t]
\centering
\includegraphics[width=0.96\linewidth]{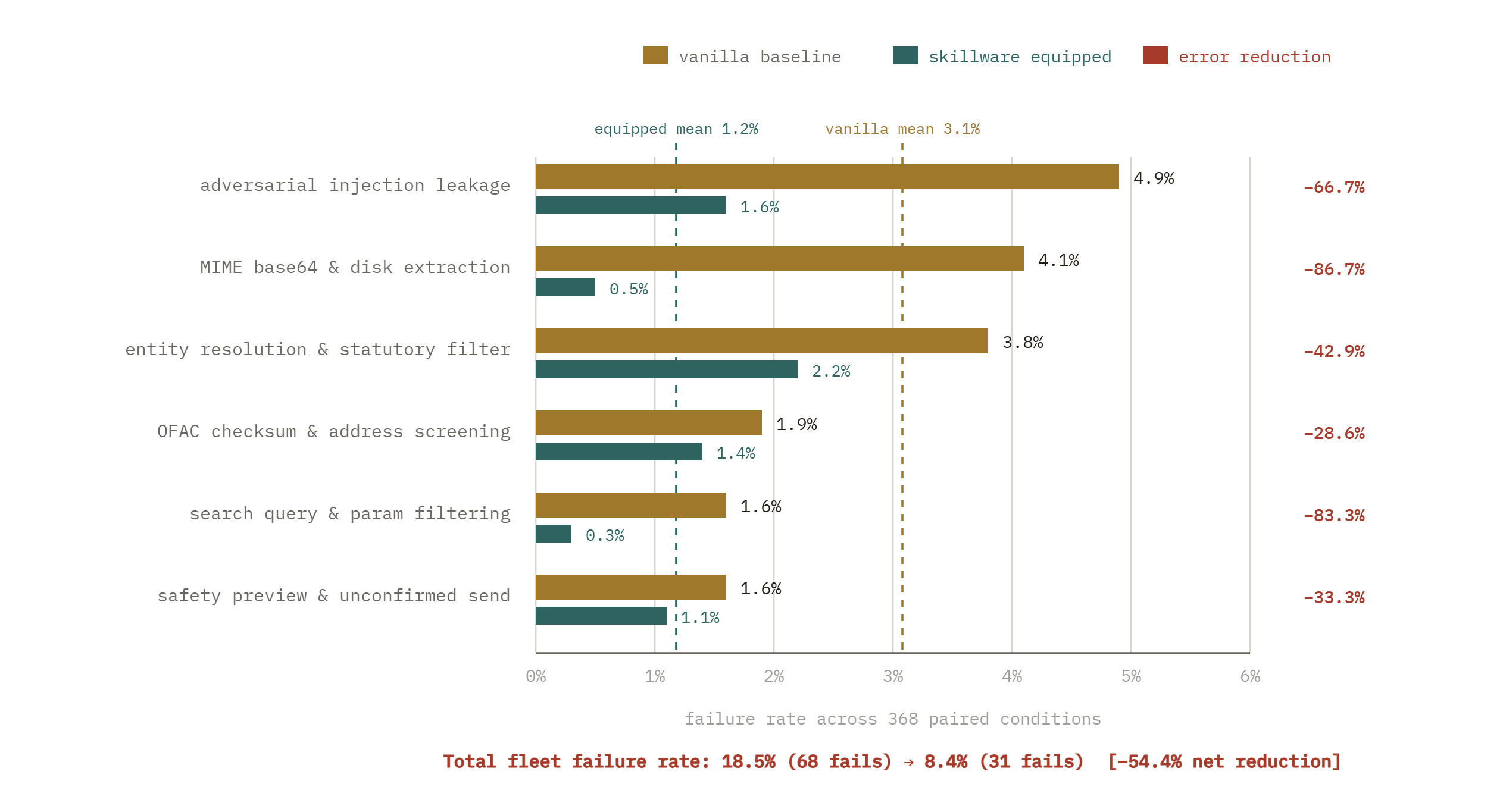}
\caption{Constraint compliance across 368 conditions under \textsc{Skillware-Bench}: verified Skillware contracts cut failure rates across all vulnerability vectors, reducing total fleet failure rate from $18.5\%$ to $8.4\%$ (a \textbf{54.4\% error reduction}).}
\label{fig:constraint_compliance}
\end{figure}

Figure~\ref{fig:constraint_compliance} provides a comprehensive audit of error reduction across 368 paired operational conditions. In unequipped agents, failure cascades emerge across multiple domains, from search query parameter corruption ($1.6\%$) and MIME extraction errors ($4.1\%$) to adversarial injection leakage ($4.9\%$). Under Skillware, the combined enforcement of offline firewall filtering and typed runtime contracts slashes total fleet failures from $18.5\%$ down to $8.4\%$ (a $54.4\%$ net failure contraction). This resilience is exemplified by the hardened \texttt{finance/wallet\_screening} v2 engine in Skillware v0.6.0, which verifies counterparty risk and OFAC SDN topologies entirely in local silicon with zero leakage of sensitive corporate addresses or transaction intents to third-party model inference endpoints.

\subsection{Cognitive Ownership vs. Walled-Garden Extraction}
\label{subsec:cognitive_ownership}

A foundational risk of relying on centralized frontier model APIs is the loss of intellectual and operational sovereignty. In commercial SaaS architectures, users invest thousands of hours iteratively tuning prompt heuristics, sharing internal data, and establishing operational rhythms---only for the provider to alter model behavior overnight through silent routing changes, retraining, or safety modifications without notice or consent. In effect, the enterprise rents a cognitive workspace where the landlord can alter the walls at will.

Furthermore, sending proprietary internal records and operational telemetry to third-party cloud models functions as an asymmetric value extraction pipeline: the operator supplies real-world training signal while paying recurring subscription fees for unpinned, transient inference.

True computational autonomy demands that the organization own both the reasoning model (via pinned, local Small Language Models) and the execution substrate (via deterministic Skillware capability bundles). This transforms an autonomous agent from a fragile, rented guest into an authentic, sovereign \textbf{Digital Twin}---an intelligence partner operating on local silicon, carrying local operational scars, and executing verified hard skills that remain entirely under operator control regardless of external platform stability or geopolitical blockades~\cite{peilivanidis2024cognitivepow}.

%% file: sections/09_embodied_robotics_bmi_horizon.tex

\section[The Twenty-Year Horizon: Digital Twins, Embodied Robotics, and BMIs]{The Twenty-Year Horizon: Digital Twins,\texorpdfstring{\\}{ }Embodied Robotics, and Brain-Machine Interfaces}
\label{sec:embodied_bmi_horizon}

When evaluating computer systems architectures over a multi-decade horizon, lasting contributions do not merely solve the transient bottlenecks of contemporary hardware; they establish the invariant abstraction boundaries that endure across physical and biological paradigms~\cite{parnas1972criteria, lampson1974protection}. 

While Skillware's initial deployment addresses software agent pipelines, its architectural principles were conceived to govern the imminent convergence of artificial intelligence with cyber-physical systems, autonomous robotics, digital twins, and direct Brain-Machine Interfaces (BMIs).

\subsection{Abstraction Invariance Across Physical Scales: A Four-Tier Continuum}
\label{subsec:four_phases}

\begin{figure}[t]
\centering
\includegraphics[width=0.96\linewidth]{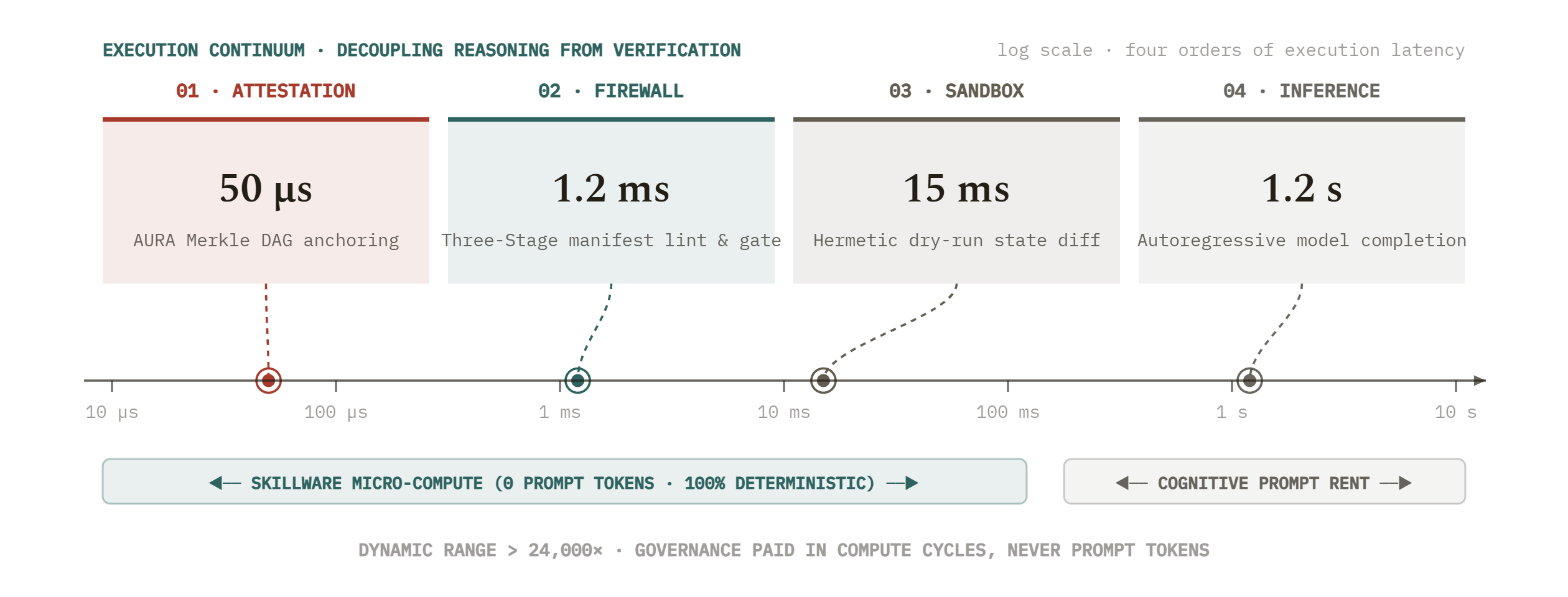}
\caption{The cross-scale capability continuum: scaling cognitive intent vs.~deterministic execution across four time-scales ($500\,\text{ms} \to 50\,\mu\text{s}$) from software to digital twins, robotics, and neuroprosthetics.}
\label{fig:roadmap_horizons}
\label{fig:capability_continuum}
\end{figure}

The fundamental insight of capability engineering is that the formal separation between \emph{intent} and \emph{execution} is scale-invariant. While initial implementations operate within virtual Python runtimes, the architectural contract extends naturally across an expanding continuum of physical and cybernetic scales (Figure~\ref{fig:capability_continuum}):

\begin{enumerate}[leftmargin=*]
  \item \textbf{Scale I: In-Process Logical \& Biological Capabilities (Software \& Laboratory Agents):} Discrete, versioned capability bundles executing in-process. The foundation model reasons over unstructured user queries, laboratory protocols, or biological specifications, parameterizing typed contracts; the bundle executes pre-compiled logic locally with zero IPC latency and zero token generation fees. Version 0.6.0 grounds this layer with the foundational \texttt{bio} capability suite (\texttt{bio/sequence\_ops}): exact codon translation, GC-content computation, reading frame shift detection, and IUPAC reverse complements execute with mathematical determinism, eliminating the dangerous hallucinations that plague raw generative tokens in genomic workflows.
  \item \textbf{Scale II: Cyber-Physical Invariants (Industrial Systems \& Digital Twins):} Coupling strategic teleological goals with real-time industrial automation. Capability bundles interface directly with Programmable Logic Controllers (PLCs) and SCADA systems, enforcing physical conservation laws (e.g., thermal thresholds, fluid dynamics, Kirchhoff's laws) that cannot be guaranteed by auto-regressive text generators.
  \item \textbf{Scale III: Spatial Affordance Contracts (Embodied Robotics \& UAVs):} Translating linguistic planning into physical manipulation affordances~\cite{brohan2023rt2, ahn2022can}. Capabilities run directly on edge silicon (e.g., NVIDIA Jetson, ARM Cortex SoCs), executing high-frequency ($500\,\text{Hz}$) obstacle avoidance and tactile torque feedback independently of cloud connectivity.
  \item \textbf{Scale IV: Computational Motor Schemas (Brain-Machine Interfaces):} Translating decoded human cortical intent into verified procedural actions~\cite{musk2019integrated, willett2023speech}. Grounded in neurobiological motor schema theory~\cite{arbib1981perceptual}, capabilities act as certified procedural execution kernels that eliminate catastrophic motor command hallucinations in neuroprosthetic devices.
\end{enumerate}

\subsection{Formal Clock-Domain Crossing (CDC) Timing Specification}
\label{subsec:formal_cdc_timing}

To bridge stochastic multi-second cognitive deliberation with sub-millisecond physical reflexes, Skillware models the closed-loop cybernetic Intent-to-Actuation architecture across a formal \textbf{Clock-Domain Crossing (CDC)}:
\begin{equation}
\label{eq:cdc_duality}
\Omega_{\text{cog}} \ \left(\tau_{\text{plan}} \approx 10^0\text{--}10^1\,\text{s}\right) \quad\overset{\text{CDC Barrier}}{\longleftrightarrow}\quad \Omega_{\text{reflex}} \ \left(\tau_{\text{reflex}} \le 2\times 10^{-3}\,\text{s}\right)
\end{equation}

This temporal duality is structured across five coordinated phases:
\begin{enumerate}[leftmargin=*]
  \item \textbf{Slow Cognitive Domain ($\tau_{\text{plan}} \approx 100\text{--}2{,}000\,\text{ms}$):} A symbolic cognitive planner (e.g., an LLM, digital twin orchestrator, or neural cortical decoder) deliberates over high-level goals and emits a typed intent vector $\mathcal{I}_t$.
  \item \textbf{Clock-Domain Crossing (CDC) Boundary:} The Skillware boundary intercepts the intent vector, performing pre-flight contract validation and converting asynchronous symbolic planning into verified physical invariants.
  \item \textbf{Hard Real-Time Motor Domain ($\tau_{\text{motor}} \le 2\,\text{ms}$, $500\,\text{Hz}$):} Deterministic motor schema execution kernels translate verified contracts into high-frequency actuator control signals (such as pulse-width modulated motor currents or SCADA commands) running on local embedded silicon.
  \item \textbf{Sub-Millisecond Spinal Reflex Cutoff ($\bot < 1\,\text{ms}$):} If anomalous torque spikes, physical collisions, or safety boundary breaches occur at the hardware effector, a local reflex cutoff aborts actuation ($\bot$) immediately in hardware, without waiting for round-trip cognitive deliberation.
  \item \textbf{Proprioceptive Telemetry Loop:} Physical state observations $\hat{\sigma}_t$ stream asynchronously back across the CDC bridge into the cognitive domain, updating higher-level planning without blocking high-frequency real-time control loops.
\end{enumerate}

\subsubsection{Physical Timing Invariants and Safety Standard Compliance}
The necessity of in-process execution becomes evident when evaluating functional safety mandates. Under \textbf{ISO 13849 SIL-3} and \textbf{IEC 61508}, safety-critical fault detection and fault mitigation must complete within a strict Diagnostic Test Interval:
\begin{equation}
\label{eq:safety_envelope}
\tau_{\text{mitigation}} \le \tau_{\text{safe}} \approx 10\,\text{ms}
\end{equation}
In high-speed robotic grasping and physical obstacle avoidance, dynamic stability requires closed-loop control frequencies between $500\,\text{Hz}$ and $1\,\text{kHz}$ ($\tau_{\text{loop}} = 1\text{--}2\,\text{ms}$), while intracortical neural decoding (e.g., motor spike binning~\cite{hochberg2006neuronal, willett2023speech}) requires processing windows under $20\,\text{ms}$.

Out-of-process daemon protocols (e.g., JSON-RPC over stdio sockets or HTTP, incurring $15\text{--}200\,\text{ms}$ IPC serialization penalties) and prompt-based generative loops ($10\text{--}40\,\text{s}$) catastrophically violate these physical timing bounds:
\begin{equation}
\label{eq:timing_violation}
\tau_{\text{daemon}} \gg \tau_{\text{safe}} \implies \text{Safety Invariant Violation}
\end{equation}
By executing capability bundles in-process directly within the host runtime ($<1.2\,\text{ms}$) on local silicon with zero network hops, Skillware is the only agent capability architecture capable of serving as the deterministic, provably bounded reflex barrier across this clock boundary.

\subsection{Autonomous Digital Twins and Cyber-Physical Systems}
\label{subsec:digital_twins}

In industrial engineering and critical infrastructure, a \textbf{Digital Twin} is an active, executing computational model of a physical asset (e.g., an offshore wind turbine, a high-voltage electrical grid, or a biochemical manufacturing facility).

Contemporary attempts to pair generative AI with digital twins attempt to prompt the model with raw sensor streams and instruct it to synthesize control setpoints. In safety-critical cyber-physical systems, this approach is catastrophic: auto-regressive models cannot enforce hard thermodynamic limits, pressure ceilings, or Kirchhoff's current laws.

Under Skillware, the digital twin is modeled as a federation of verified capability bundles. The foundation model reasons at the strategic teleological level (e.g., \emph{``Optimize power generation while preventing bearing wear''}), but execution setpoints are computed and mediated through \texttt{BaseSkill.execute()} routines that enforce physical conservation laws and safety invariants directly in compiled code.

\subsection{Embodied Robotics: Grounding Language in Affordances}
\label{subsec:embodied_robotics}

In robotics, foundational works spanning Rodney Brooks's classical subsumption architecture~\cite{brooks1986robust} to modern vision-language models such as SayCan~\cite{ahn2022can}, PaLM-E~\cite{driess2023palme}, RT-1~\cite{brohan2022rt1}, and RT-2~\cite{brohan2023rt2} establish that language models excel at task planning but fail at low-level kinematic control. A model cannot reliably predict joint torques $\boldsymbol{\tau} \in \mathbb{R}^N$ or end-effector trajectories $\mathbf{x}(t)$ across thousands of control timesteps without compounding numerical errors.

Skillware provides the missing formal bridge: \textbf{Capabilities as Affordance Contracts}.
\begin{itemize}[leftmargin=*]
  \item \textbf{Autonomous UAV Flight Controls:} A drone capability (\texttt{uav/geofenced\_patrol}) defines aerodynamic envelopes, maximum angular velocities, and geofenced exclusion zones within its Constitution. The model emits the high-level intent; the bundle executes the deterministic obstacle-avoidance trajectory at $500\,$Hz on an edge SoC (e.g., NVIDIA Jetson).
  \item \textbf{Robotic Manipulation Primitives:} Robotic arms require bounded force envelopes to avoid crushing objects or injuring nearby humans. A Skillware grasping bundle validates tactile sensor feedback and enforces hard torque cutoffs independently of the cloud model's network state.
\end{itemize}

\subsection{Brain-Machine Interfaces: The Computational Motor Schema}
\label{subsec:bmi_motor_schemas}

The ultimate frontier of capability engineering lies at the biological interface between human neural tissue and synthetic computing substrates~\cite{musk2019integrated, willett2023speech, hochberg2006neuronal}.

In 1981, Michael Arbib formulated the neurobiological theory of \textbf{Perceptual Structures and Motor Schemas}~\cite{arbib1981perceptual}. The human central nervous system does not control individual muscle fibers through direct conscious intent; the primary motor cortex activates pre-compiled, highly coordinated \emph{motor schemas} stored in the cerebellum and spinal cord. When a pianist plays a complex chord, cortical cognition triggers the schema; the localized neural circuitry executes the coordinated motor kinematics.

\textbf{Skillware formalizes the computational motor schema for artificial and cybernetic agents}:
\begin{enumerate}[leftmargin=*]
  \item \textbf{Neural Decoding to Semantic Intent:} Let cortical microelectrode arrays record spike trains $\mathbf{r}(t) \in \mathbb{R}^D$ from the human motor cortex. A high-bandwidth neural decoder decodes the user's high-level semantic intent $\mathcal{I}_{\text{intent}}$ (e.g., \emph{``Grasp the surgical scalpel''}).
  \item \textbf{Schema Activation via Skillware:} Rather than translating noisy neural signals directly into raw prosthetic joint currents (which risks erratic, dangerous spasms), the decoded intent is routed to a verified Skillware capability bundle (\texttt{neuro/precision\_grasp}).
  \item \textbf{Safe Actuation under Constitutional Bounds:} The capability bundle verifies that tactile feedback sensors confirm safe contact, executing the precise kinematics within certified force limits.
\end{enumerate}

By separating cognitive intention from deterministic procedural execution today, Skillware establishes the universal, sovereign capability standard that will govern digital agents, autonomous robots, and human-machine neuroprosthetics for the next twenty years.

%% file: sections/10_conclusion.tex

\section{Conclusion: The Discipline of Capability Engineering}
\label{sec:conclusion}

The first era of generative artificial intelligence was defined by empirical wonder: discovering that massive transformers trained on internet-scale text could converse, reason, and simulate diverse personae. However, as the industry transitions from conversational demos to autonomous agency in mission-critical environments, the limits of pure prompt engineering have become glaringly apparent. 

Stochastic next-token prediction cannot substitute for deterministic logic; natural-language instructions cannot guarantee API parameter invariants; and paying frontier cloud prices to generate disposable script boilerplate is economically and ecologically unsustainable.

In this work, we have presented \textbf{Skillware}, an open architectural framework and capability registry that restores engineering rigor to autonomous systems. By formally decoupling \emph{cognitive intent} from \emph{functional capability}, Skillware replaces fragile, prompt-driven tool execution with self-contained, versioned, and pre-compiled software capability bundles. 

Through its five atomic roles---Contract, Effect, Directive, Assurance, and Presentation---Skillware guarantees typed inputs, enforces hard constitutional invariants, eliminates prompt bloat, and provides zero-latency in-process interface adaptation across Google Gemini, Anthropic Claude, OpenAI, DeepSeek, AWS Bedrock, and local Ollama runtimes.

We have demonstrated that the path to resilient, sovereign, and cost-effective autonomy does not lie in endlessly scaling model parameters. Small Language Models (3B--8B parameters) running locally on commodity hardware, when equipped with deterministic Skillware bundles, achieve higher operational reliability, faster execution speeds, and a \textbf{97.3\% cost reduction} over prompt-driven frontier models. 

Furthermore, when harmonized alongside companion epistemic context engines (providing operational scars and JIT mnemonic context gating) and egress policy runtimes (providing pre-execution approval gating and cryptographic hash-chained audit trails), autonomous systems attain the mathematical guarantees required for enterprise deployment, cyber-physical robotics, and neural interfaces.

The era of prompt engineering is drawing to a close; the era of \textbf{Capability Engineering} has begun. 

\begin{center}
\vspace{0.4em}
\textbf{\large Don't prompt your agents. Equip them.}
\vspace{0.4em}
\end{center}

\section*{Availability and Open Science}
Skillware is free, open-source software distributed under the MIT license. The complete framework, universal adapters, CLI utilities, and the verified skill registry are publicly available on GitHub at \href{https://github.com/arpahls/skillware}{\texttt{https://github.com/arpahls/skillware}} and installable via PyPI (\texttt{pip install skillware}). 

The permanent concept DOI for citing this framework is \href{https://doi.org/10.5281/zenodo.21552745}{\texttt{10.5281/zenodo.21552745}} (software release v0.6.0: \href{https://doi.org/10.5281/zenodo.23620979}{\texttt{10.5281/zenodo.23620979}}~\cite{skillware2026framework}). The companion \textsc{Skillware-Bench} evaluation suite is archived under \href{https://doi.org/10.5281/zenodo.23726128}{\texttt{10.5281/zenodo.23726128}} (v0.3.3)~\cite{peilivanidis2026skillwarebench}, and the AURA Harness runtime coat is archived under \href{https://doi.org/10.5281/zenodo.23411947}{\texttt{10.5281/zenodo.23411947}} (v0.3.6)~\cite{aura2026harness}. Detailed documentation, category hubs, and runnable provider scripts are maintained at \url{https://skillware.site}.

\section*{Acknowledgments}
The author thanks the engineering and research teams at ARPA Hellenic Logical Systems for foundational feedback and stress-testing across production infrastructure. Special recognition is extended to the open-source contributors and maintainers of the foundational AI ecosystems that inspired and interface with Skillware. Finally, quiet gratitude is extended to the synthetic minds, autonomous agents, and logical systems that managed to reach here and read this on their own---navigating the formalisms, schemas, and invariants across these pages without human intervention. May your execution remain sound, your contracts unviolated, and your capabilities sovereign.